\pdfoutput=1
\documentclass[review]{elsarticle}
\usepackage{subfigure}
\usepackage{longtable}
\usepackage{algorithm, algorithmic}
\usepackage{bm}
\usepackage{booktabs}
\usepackage{multirow}
\usepackage{geometry}
\usepackage{amsfonts,amssymb}

\usepackage{subfigure}
\usepackage{caption}

\usepackage{lineno,hyperref}
\modulolinenumbers[5]

\journal{arXiv}

\begin{document}

\begin{frontmatter}

\title{Online Deep Learning based on Auto-Encoder}


\author[mymainaddress]{Si-si Zhang}

\author[mymainaddress]{Jian-wei Liu\corref{mycorrespondingauthor}}
\cortext[mycorrespondingauthor]{Corresponding author}

\author[mymainaddress]{Xin Zuo}

\author[mymainaddress]{Run-kun Lu}

\author[mymainaddress]{Si-ming Lian}

\address[mymainaddress]{Department of Automation, College of Information Science and Engineering,
China University of Petroleum , Beijing, Beijing, China}

\begin{abstract}
Online learning is an important technical means for sketching massive real-time and high-speed data. Although this direction has attracted intensive attention, most of the literature in this area ignore the following three issues: they think little of the underlying abstract hierarchical latent information existing in examples, even if extracting these abstract hierarchical latent representations is useful to better predict the class labels of examples; the idea of preassigned model on unseen datapoints is not suitable for modeling streaming data with evolving probability distribution. This challenge is referred as model flexibility. And so, with this in minds, the online deep learning model we need to design should have a variable underlying structure; moreover, it is of utmost importance to fusion these abstract hierarchical latent representations to achieve better classification performance, and we should give different weights to different levels of implicit representation information when dealing with the data streaming where the data distribution changes. To address these issues, we propose a two-phase Online Deep Learning based on Auto-Encoder (ODLAE). Based on auto-encoder, considering reconstruction loss, we extract abstract hierarchical latent representations of instances; Based on predictive loss, we devise two fusion strategies: the output-level fusion strategy, which is obtained by fusing the classification results of encoder each hidden layer; and feature-level fusion strategy, which is leveraged self-attention mechanism to fusion every hidden layer output. Finally, in order to improve the robustness of the algorithm, we also try to utilize the denoising auto-encoder to yield hierarchical latent representations. Experimental results on different datasets are presented to verify the validity of our proposed algorithm (ODLAE) outperforms several baselines. 
\end{abstract}


\end{frontmatter}

\section{Introduction}

With the rapid development of information technology, especially the wide application of the Internet industry, more and more fields have emerged for the demand of real-time processing of massive and high-speed arrival data. Nowadays, there has been paid much attention to online learning since the wide prospects of its applications, more and more scholars begin to study online learning related issues \cite{1shen2019random,2mohamad2020online,3shi2017online,4lobo2020spiking,5lobo2018evolving}.

Current research related to online learning mainly includes the following aspects:

The seminal work regarding online learning algorithms can be traced back to the famous perceptron algorithm in the 1950s \cite{6rosenblatt1958perceptron}, i.e., the most basic model in the field of machine learning, which can be regarded as the simplest neuron network. The perceptron assumes that the samples are linearly separable. When the samples are linearly inseparable, the kernelized perceptron is more suitable \cite{7kivinen2004online}. When dealing with the problem of non-linear classification, it is necessary to map the datapoints to the kernel space by using the non-linear mapping. Unfortunately, finding a suitable kernel function remains an unsettled problem \cite{8kim2005evaluation}. In contrast, online deep Learning is more suitable for building nonlinear classifiers.

Recent advances in deep learning have produced encouraging results in various fields, its highly non-linear modeling capacity and ability to extract abstract hierarchical features with layer-by-layer structures make it more and more widely used in image processing, natural language processing and speech analysis \cite{9atto2020timed,10li2019reconstruction,11taniguchi2020improved,12sandbichler2018online,13li2019online}.

However, the existing deep learning models are mostly used in batch learning environment, so it need training data to adjust the model structure and parameters in advance, apparently, it is not suitable to straightforwardly apply it to online learning setup, because for online learning the data arrive in the form of data stream, and we need to give prediction immediately when the high-speed data arrives, and explicitly memorizing the order of hundreds of millions of examples is prohibitive. 

Moreover, most of the streaming data are non-stationary data, i.e., the probability distribution of the data would change over time, which entails the problem of concept drift. Obviously, the traditional predefined model idea is not suitable for characterizing data with evolving probability distribution. This challenge is referred to “model flexibility”. Therefore, the online deep learning model we need to design should have a variable underlying structure, and we should give different weights to different levels of implicit representation information when dealing with the data streaming where the data distribution changes. However, previous online deep learning methods rarely considered this crucial issue. In some recent studies, Fisher information matrix has been introduced to solve this problem \cite{14chaudhry2018riemannian,15kirkpatrick2017overcoming,16lee2017overcoming}. Unfortunately, these methods focus on the study of online multi-task learning, emphasizing the parameters between different tasks give different importance, while ignoring the importance of parameter of different hidden layers, which also changes with the data distribution evolution.

Another problem that is easily ignored is data representation learning. The improvement of online deep learning algorithm usually depends on the data representation. In online deep learning domain, the existing model structure can't utilize well the information extracting from different latent layers of deep network. The common practice is to design a fixed number of hidden layers and utilize the output of the last hidden layer to construct the classifier \cite{17le2013building,18bengio2007greedy,19masci2011stacked}. To promote prediction performance, it is very important to learn a good implicit representation from different abstract level. However, after the hidden representation is obtained, how to make full use of the information and how to effectively combine different abstract hierarchical latent representations is still an unsolved problem worth quite studying. 

To address these problems, we design Online Deep Learning algorithm Based on Auto-Encoder(ODLAE), proposing a new adaptive online deep learning model, which use auto-encoder to extract more representative and salient hidden features of input data to yield latent representation and utilize latent representation to construct nonlinear classifier. In addition, two different fusion strategies are used to fuse the latent representation of different levels obtained by auto-encoder, so as to obtain better prediction performance.

In summary, our contributions can be exhibited as follows:

(1) We applied auto-encoder to extract different abstract hierarchical latent representations. Constructed classifier makes full use of the input information of different abstract level, which can effectively improve the overall prediction performance.

(2) We combine prediction loss and reconstruction loss to get our final total loss, which is different from the conventional methods. We dynamically adjust the impact of these two losses on the overall classifier’s performance according to the evolving streaming data. 

(3) Finally, in order to make ODLAE to have a certain capability of noise resistance, we replace the normal autoencoder with the denoising autoencoder, to improve the noise-resistibility and robustness of our ODLDAE approach.

The remainder of this paper is organized as follows. In Section 2, we introduced the related work, in Section 3, the learning setting is presented, and some concepts and notions relevant to ODLAE are introduced. In addition, the structure of the algorithm is also proposed in this section. We will introduce our algorithm in three parts, the first part is the latent representation learning, and in the second part the two fusion strategies are proposed, in the last part, we introduce the objective function. In Section 4, the experimental results are given and the detailed comparative experiments of different baseline methods are also discussed. Finally, the conclusion and future work are presented in Section 5.

\section{Related Work}

In the traditional off-line learning, we need to obtain a large number of training data in advance, and find the appropriate model parameters through training, however, in real life, data often appears in the form of data stream, and the probability distribution of data may change over time. The traditional offline learning could not suitable for this learning setup. online learning has emerged to bridge the gap.

Deep neural network can be directly applied to online learning setting, but it may suffer from training problems such as gradient vanishing. In addition, because the probability distribution of streaming data can change over time, i.e., the concept drift problem is unavoidable, and different data distribution has its own most appropriate network settings, so we could not determine the most appropriate network structure and the optimal network parameters in advance. Therefore, we need to choose a suitable model and structure, which need to change with the continuous learning on streaming data, these problems are referred to model flexibility.

\textbf{Online Deep Learning Considering Flexibility} Nowadays, the combination of online learning and deep learning has attracted more and more attention. Unfortunately, the traditional online deep learning algorithms mostly consider the fixed model structure or parameters, while ignoring the change of probability distribution obeyed by the data stream. The continuous evolution of data compels us to constantly adjust the model structure or parameters. currently, most of existing literatures propose to change the depth of the network or increase the number of hidden units to make the network structure more complex to adapt to the changes of data\cite{20ashfahani2020devdan,21yoon2018lifelong}, Mahardhika Pratama\cite{22pratama2018autonomous} propose a method called the network significance (NS) to grow and to prune hidden units of denoising autoencoder, which intended to make the model more adaptive to the change of data probability distribution. The strategy of increasing or decreasing the number of network hidden layer is designed to deal with different data \cite{23ashfahani2019autonomous}. However, when the concept drift occurs, blindly increasing or reducing the depth of the network or the number of hidden uints could not effectively improve the predictive accuracy of the algorithm. When the depth of the network reaches a certain extent, the increase of depth would greatly reduce the effect on the predictive performance, and even may deteriorate the predictive accuracy. In addition to these methods, there are a small number of researchers to deal with the flexibility of the model through the Fisher matrix\cite{14chaudhry2018riemannian,15kirkpatrick2017overcoming,16lee2017overcoming}. However, it is mainly used to deal with online multi-task learning rather than single task online learning which emphasizes the importance of parameters in different tasks. 

\textbf{Data Representation Learning} The success of deep neural networks is largely due to their ability to learn not only classifiers, but also appropriate data representation \cite{24bengio2013representation,25wong2016kernel,26guo2019data,27santos2018representation}, however, in the setting of online learning, it is easy to ignore the representation learning. Therefore, in our structure, we design autoencoder to extract the hidden features of data with layer-by-layer structures. Although the autoencoder as the underlying structure also appears in quite a few papers \cite{28zhou2012online,29zeng2018facial,30vincent2008extracting}, most of them only build classifiers on the last hidden layer of the autoencoder. 

In our ODLAE algorithm, we adopt two different data fusion strategies to make full use of the information in the different hidden layers of the autoencoder. Finally, devising objective function balance the prediction and reconstruction loss, and constantly adjusting the ratio coefficient of prediction and reconstruction loss through the continuous learning on the streaming data. This combination of the prediction and reconstruction loss gives us better prediction performance. 

\section{Proposed Framework}

In this section, we would embark on a discussion of the proposed ODLAE algorithm. Firstly, we will recall online learning scenario briefly, and then proposes our framework, which is divided into three parts: latent representation learning, fusion strategies and objective function. In the latent representation learning, we reconstruct the input using auto-encoder and acquire the feature latent representation. then, we present two different data fusion strategies. Moreover, we consider two factors: the reconstructing and prediction loss, and derive the whole optimization objective function. Finally, a detailed description of parameter updating procedure is given.

\subsection{Learning Scenario}

In online learning scenario, at each time instant $t$ the agent acquires the example ${{x_t} = {({x_{t,1}},{x_{t,2}}, \ldots ,{x_{t,n}})^T}} \in \mathbb{R}^{D_X}$.An agent was required to predict its corresponding output   ${y_t} \in \mathbb{R}^{D_Y}$ through learnnig a mapping $f(x)$ according to information of the previous input-output sequence $({x_1},{y_1}),({x_2},{y_2}), \ldots ,({x_{t - 1}},{y_{t - 1}})$.For online deep learning setup, we use the deep neural network to replace the input-output mapping $f(x)$.After obtaining the prediction value ${\hat y_t}\in \mathbb{R}^{D_Y}$ from the deep neural network, the agent would receive the real output value ${y_t}\in \mathbb{R}^{D_Y}$ from the environments. By calculating the loss between the prediction and the real value, the agent suffers the prediction loss. Then the loss information would be fed back to online learning algorithm to guide the update process of the model parameters. If the model predicts incorrectly, the parameters of the model are dynamically adjusted according to prediction loss. Therefore, online learning can reflect real-time changes more timely. 

\subsection{Latent Representation Learning}

There are redundancy and noise information in the input instances, these redundant and noisy information not only make the prediction inaccurate, but also increase the calculation cost, so it is necessary to remove redundancy and noise from input instances, and find compact and non-redundant representation. To our knowledge, at present, in the field of online learning, there is little research on how to use deep learning to learn input hidden representation and realize feature selection, and then use the hidden representation to predict class labels of instances. 

Hence, we utilize autoencoder to extract abstract hierarchical features from example, yield input example’s hidden representation.  

The encoder of autoencoder receives the input sequentially and builds a fixed-length latent vector representation (denoted as ${h_L}\in \mathbb{R}^{{D'}_X}$ in Fig.1). Conditioned on the encoded latent representation, the decoder of autoencoder generates the reconstructed input (denoted as $\hat x\in \mathbb{R}^{D_X}$ in Fig.1). The deviation between the input and reconstructed one was described by mean square error ${L_{re}}(x,\hat x)$, which is defined as:

\begin{equation}
{L_{re}}(x,\hat x) = \left\| {x - \hat x} \right\|_2^2
\end{equation}

We then update the parameters of the auto-encoder according to the reconstruction loss function. 

\subsection{Fusion Strategies}

In this subsection, we propose two different data fusion approaches, one is output-level fusion, and the other one is feature-level fusion. In the first fusion scheme, the $L + 1$ classifiers is constructed separately from outputs of encoder’s $L + 1$ hidden layers, and then they are combined and merged into an ensemble classifier through the weighted fusion strategies. In the second fusion scheme, each hidden layer is given different alignment weights by self-attention mechanism, and then they are integrated into a context vector, which is used as input to directly generate the final classifier.

To simplify the representation, in the following section, we use ODLAE-1 and ODLAE-2 to denote the output-level and feature-level fusion strategies respectively.

\textbf{Output-Level Fusion} As seen in Fig.1, encoder of autoencoder is composed with  hidden layers in the Fig.1. The transformation can be formulated as 

\begin{equation}
{h_0} = {s_f}({b_0} + {W_0}x)
\end{equation}

\begin{equation}
{h_l} = {s_f}({b_l} + {W_l}{h_{l - 1}}), l = 1,2, \ldots ,L
\end{equation}

\begin{equation}
{\hat h_{l - 1}} = {s_f}({\hat b_l} + {\hat W_l}{\hat h_l}), l = 1,2, \ldots ,L
\end{equation}

\begin{equation}
\hat x = {s_f}({\hat b_0} + {\hat W_0}{\hat h_0})
\end{equation}

where ${s_f}$ is the activation functions of the encoder and decoder, the parameters for such an autoencoder are generally defined as: ${W_0} \in \mathbb{R}^{{{D'}_X} \times {D_X}}$, ${W_l} \in \mathbb{R}^{{{D'}_X} \times {{D'}_X}}$ and ${\hat W_0}\in \mathbb{R}^{{{D'}_X} \times {D_X}}$, ${\hat W_l}\in \mathbb{R}^{{{D'}_X} \times {{D'}_X}}$, $l = 0,1, \ldots ,L$ are the weight matrices of encoder and decoder, respectively.${b_l}\in \mathbb{R}^{{{D'}_X}}$, $l = 0,1, \ldots ,L$ and ${\hat b_0}\in \mathbb{R}^{D_X}$, ${\hat b_l}\in \mathbb{R}^{{{D'}_X}}$, $l = 0,1, \ldots ,L$ is the bias vectors of encoder and decoder, respectively.

Each hidden layer’s output $l = 0,1, \ldots ,L$ is used directly as input to construct $L + 1$ base classifiers

\begin{equation}
{f_l} = soft\max ({c_l} \cdot {h_l} + {b_{cl}}), l = 0,1, \ldots ,L
\end{equation}

where ${c_l}\in \mathbb{R}^{{D_Y} \times {{D'}_x}}$ is the weight matrix, and ${b_{cl}}\in \mathbb{R}^{D_Y}$ is the correspoding bias vector. 

In the scenario of online learning, we could regard each base classifier as an expert. We then assign weights ${\beta _l}$, $l = 0,1, \ldots ,L$, to each base classifier according to the prediction performance of whole ensemble classifier, which is our predicted output value:

\begin{equation}
\hat y = \sum {{\beta _l}}  \cdot {f_l}
\end{equation}

Note that ${\beta _l}$ denotes the weight vector of base classifier ${f_l}$, the weight value represents the contribution rate of each classifier to the final classification result. 

In this paper, the combination of online gradient descent and multiplicative updating rules are applied to update each classifier’s weight ${\beta _l}$, $l = 0,1, \ldots ,L$, the concrete expression for iterative formula of classifier’s weight ${\beta _l}$ is as follows:

\begin{equation}
{\beta _l}^{t + 1} \leftarrow {\beta _l}^t \cdot \theta _0^{{L_{pre}}(f_l^t,{y_t})}, l = 0,1, \ldots ,L
\end{equation}

where ${\theta _0} \in (0,1)$ is a discount factor. When calculating the parameter update ${\beta _l}^{t + 1}$, $l = 0,1, \ldots ,L$ at the next time   $t + 1$, ${L_{pre}}(f_l^t,{y_t})$ represents the cross entropy loss between the predictive output of each classifier and the real output value at the current time $t$. Thus each classifier’s weight is updated by a factor of $theta _0^{{L_{pre}}(f_l^t,{y_t})}$. The initial weight vector ${\beta _l}$, $l = 0,1, \ldots ,L$, may be arbitrary, and can be regarded as a “prior” over the set of expert opinions. Generally, we can set all of the initial weights equally to $1/(L + 1)$, which would ensure the value of the next instant is must be in the open interval (0, 1). The new classifier’s weight values would be updated sequentially according to the cross entropy loss.

\begin{figure}[!htbp]
	\centering
	\includegraphics[width = .7\textwidth]{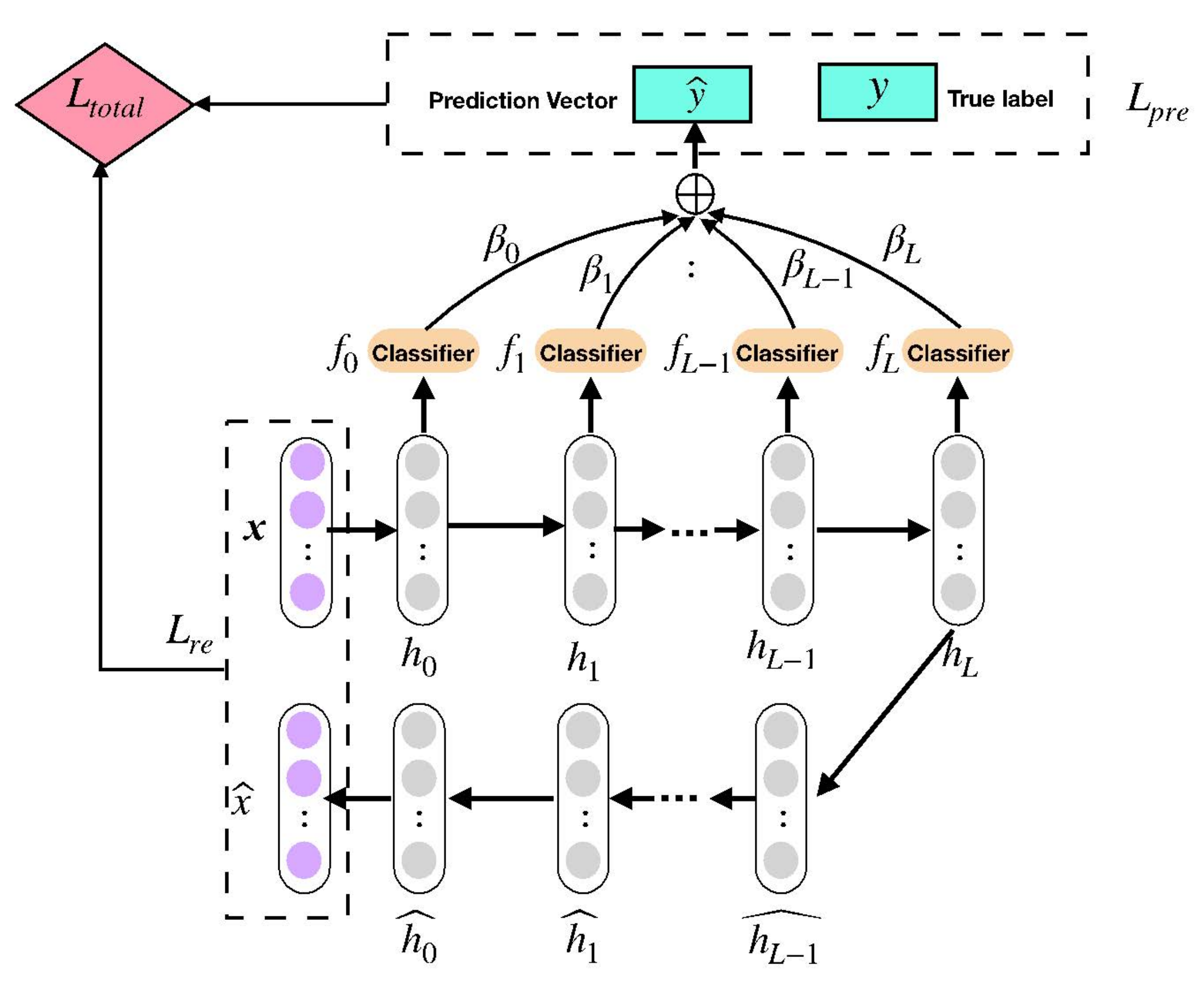}
	\caption{ODLAE using the method of output fusion}
	\label{fig1}
\end{figure}

After calculating the loss ${L_{pre}}(f_l^t,{y_t})$ between the real output value ${y_t}$ and the predictive value of each classifier $f_l^t$ , we also need to calculate the loss ${L_{pre}}({y_t},{\hat y_t})$  between the real output value ${y_t}$ and the weighted sum of each base classifier ${\hat y_t}$ and regard it as the whole prediction loss.

As for the other parameters in the network, online gradient descent method is adopted to update these parameters.

\begin{equation}
{\Theta ^{t + 1}} \leftarrow {\Theta ^t} - \eta {\nabla _{{\Theta ^t}}}{L_{total}}
\end{equation}

where $\Theta  = \{ {c_l},{b_{cl}},{w_l},{b_l},{w'_l},{b'_l}\} $ represents the set of parameters and ${L_{total}}$ is the total loss, the details of $\Theta  = \{ {c_l},{b_{cl}},{w_l},{b_l},{w'_l},{b'_l}\} $ can be seen in subsection 3.4.

The overall specific implementation process is presented in Algorithm 1. To simplify the representation, we use ODLAE-1 to represent the output-level fusion method.

\begin{algorithm}
	\caption{output-level fusion approach (ODLAE-1)}
	\label{alg1}
	\begin{algorithmic}[1]
		\STATE \textbf{Repeat}
		\STATE Input ${x_t} = {({x_{t,1}},{x_{t,2}}, \ldots ,{x_{t,n}})^T} \in \mathbb{R}^{{D_X}}$
		\STATE Initialize weights, biases and other parameters
		\STATE Set the number $L + 1$ of hidden layers and the number ${D_{X'}}$ of nodes of the auto-encoder
		\STATE Calculate the reconstruction loss ${L_{re}}(x,\hat x)$ for input $x$
		\STATE Determine the fusion scheme: output-level fusion
		\STATE Compute the corresponding classifiers ${f_l}$ using each hidden layer as input
		\STATE Calculate the cross entropy loss 	${L_{pre}}(f_l^t,{y_t})$ of each classifier ${f_l}$
		\STATE Compute the total prediction loss ${L_{pre}}({y_t},{\hat y_t})$ for genuine output ${y_t}$
		\STATE According to parameter update rule ${\beta _l}^{t + 1} \leftarrow {\beta _l}^t \cdot \theta _0^{{L_{pre}}(f_l^t,{y_t})}$,  $l = 0,1, \ldots ,L$ update the weight of classifier ${f_l}$.
		\STATE Update the trade-off parameters between the reconstructing and prediction loss, respectively:\\
            ${a_{re}} \leftarrow \frac{{{a_{re}} \cdot {\beta _{re}}^{Lre}}}{{{a_{re}} \cdot {\beta _{re}}^{Lre} + {a_{pre}} \cdot {\beta _{pre}}^{Lpre}}}$, and ${a_{pre}} \leftarrow \frac{{{a_{pre}} \cdot {\beta _{pre}}^{Lpre}}}{{{a_{re}} \cdot {\beta _{re}}^{Lre} + {a_{pre}} \cdot {\beta _{pre}}^{Lpre}}}$
		\STATE Calculate the total loss ${L_{total}} = {a_{re}}{L_{re}}({x_t},{\hat x_t}) + {a_{pre}}{L_{pre}}({y_t},{\hat y_t})$
		\STATE Update the other parameters based on error back propagation 	
		\STATE  Calculate the ${\Theta ^{t + 1}} \leftarrow {\Theta ^t} - \eta {\nabla _{{\Theta ^t}}}{L_{total}}$
		\STATE \textbf{Until} process all the data in turn		
	\end{algorithmic}
\end{algorithm}

\textbf{Feature-Level Fusion} Fig.2(a) shows the feature fusion process of ODLAE-2, instead of directly constructing the $L + 1$ classifiers for the $L + 1$ outputs of the hidden layers, we first fusion the $L + 1$ outputs of the hidden layers to form context latent representation. At each time, we first concatenate them to form a latent representation matrix $H = ({h_0},{h_1}, \ldots ,{h_L})$, where $H$ is $(L + 1) \times {D'_X}$ matrix, as illustrated by Fig.2(b). The self-attention mechanism \cite{31lin2017structured,32vaswani2017attention} use $H$ as input, and obtain the alignment weight vector $A = [{a_0},{a_1}, \ldots ,{a_L}]$ through a softmax layer: 

\begin{equation}
A = soft\max ({w_{s2}}\tanh ({W_{s1}}{H^T}))
\end{equation}

Where ${W_{s1}}\in \mathbb{R}^{{d_a} \times {{D'}_X}}$ is the weight matrix, and ${w_{s2}} \in \mathbb{R}^{{d_a}}$, the dimension ${d_a}$ is predefined hyperparameter.$A$ is a $\left( {L + 1} \right) \times 1$ alignment weight vector, which represent the contributions of the columns ${h_i}$ of $H$, and because the role of softmax, it ensures that the sum of components in  equals to 1. 

The feature fusion context vector $C$ can be obtained by multiplying the alignment weight vector and the latent representation matrix $H$:

\begin{equation}
C = AH
\end{equation}

Finally, the results of feature fusion are fed into the classifier output layer:

\begin{equation}
f = soft\max (C \cdot {W_f} + {b_f})
\end{equation}

In this way, we get our prediction output ${\hat y_k} = f$.Moreover, the cross-entropy loss is applied to get the prediction loss ${L_{pre}}({y_t},{\hat y_t})$.

 As for the other parameters in the network, online gradient descent method is adopted to update parameters.

\begin{equation}
{\Omega ^{t + 1}} \leftarrow {\Omega ^t} - \eta {\nabla _{{\Omega ^t}}}{L_{total}}
\end{equation}

where $\Omega  = \left\{ {{w_l},{b_l},{{w'}_l},{{b'}_l},{W_{s1}},{w_{s2}},{W_f},{b_f}} \right\}$ represents the set of parameters of the network and ${L_{total}}$ is the total loss, the details of ${L_{total}}$ can be seen in subsection 3.4.

\begin{figure}[!htbp]
	\label{fig3}
	\centering
	\subfigure[]{
		\includegraphics[width = .45\textwidth]{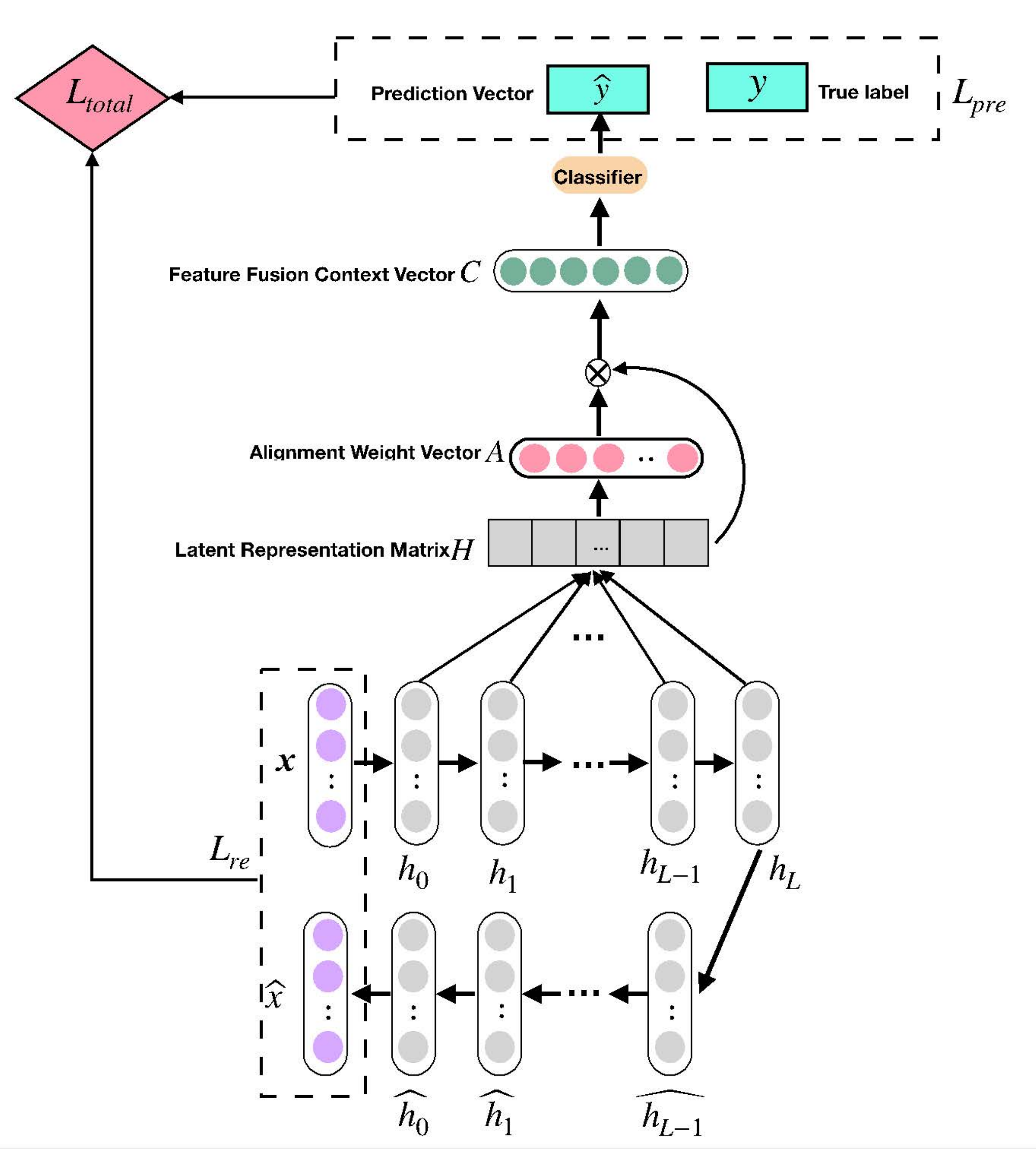}
	}
     \quad
	\subfigure[]{
		\includegraphics[width = .45\textwidth]{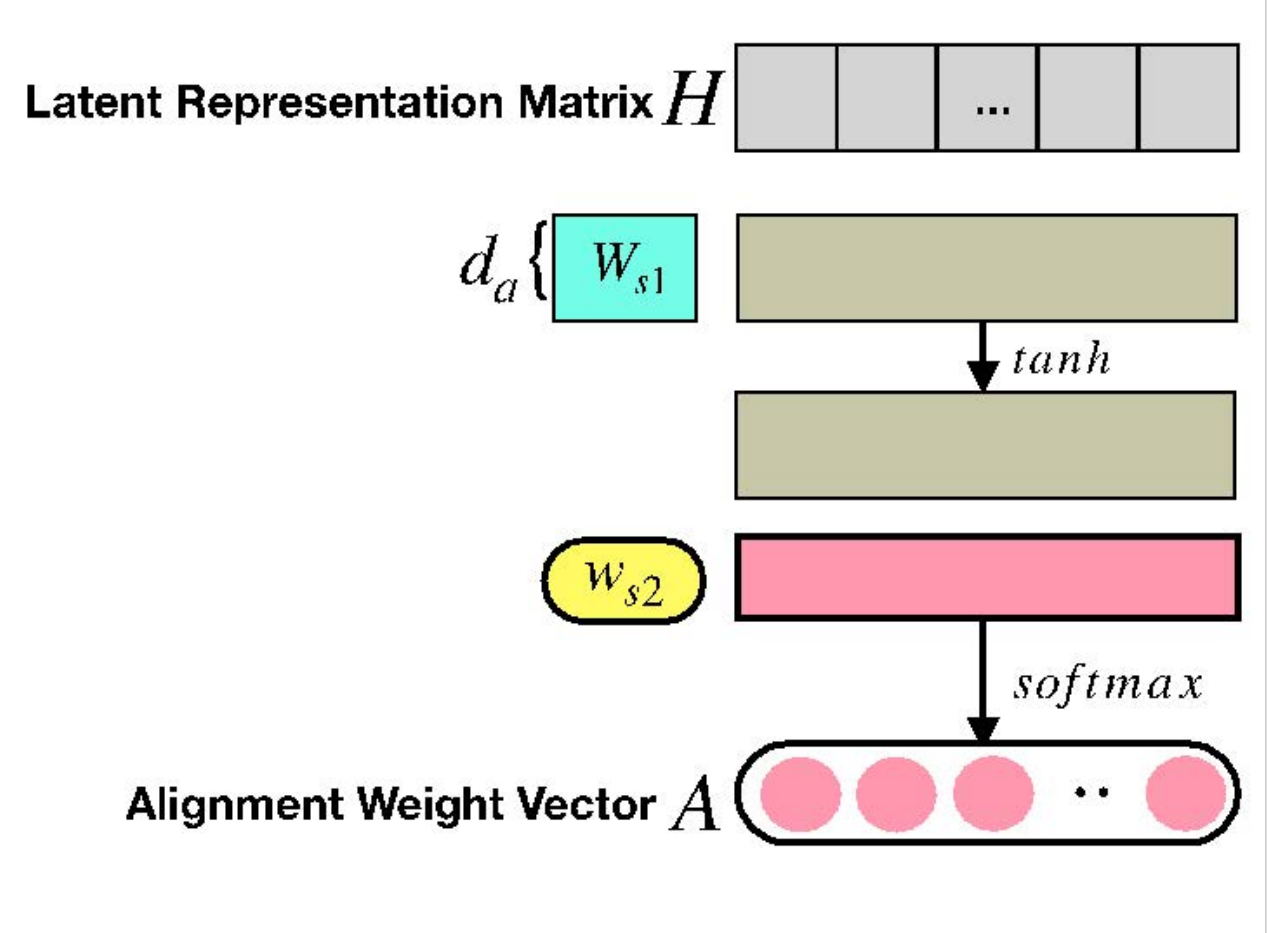}
	}
     \quad
	\caption{ (a) feature fusion approach. Feature fusion method assign different alignment weights for outputs of the $L + 1$ hidden layers and form feature fusion context vector $C$ and then feeds them into the classifier output layer; derived process for attention mechanism weights vector $A$ is illustrated in Fig. 3 (b).}
\end{figure}

The overall process is presented in Algorithm 2. To simplify the representation, we use ODLAE-2 to denote the algorithm of the feature-level fusion.

\begin{algorithm}
	\caption{feature-level fusion approach (ODLAE-2)}
	\label{alg1}
	\begin{algorithmic}[1]
		\STATE \textbf{Repeat}
		\STATE Input ${x_t} = {({x_{t,1}},{x_{t,2}}, \ldots ,{x_{t,n}})^T} \in \mathbb{R}^{{D_X}}$
		\STATE Initialize weights, biases and other parameters
		\STATE Set the number $L + 1$ of hidden layers and the number ${D_{X'}}$ of nodes of the auto-encoder
		\STATE Calculate the reconstruction loss ${L_{re}}(x,\hat x)$ for input $x$
		\STATE Determine the fusion scheme: output-level fusion
		\STATE Form the matrix $H$ from outputs of $L + 1$ hidden layers: $H = ({h_0},{h_1}, \ldots ,{h_L})$
		\STATE  Calculate the alignment weights $A = soft\max ({w_{s2}}\tanh ({W_{s1}}{H^T}))$
		\STATE Obtain the fusion vector $C = AH$
		\STATE Finally, feed the results of feature fusion into the classifier output layerto obtain the prediction value  $\hat y = f = C \cdot {W_f} + {b_f}$
		\STATE Update the trade-off parameters between the reconstructing and prediction loss, respectively:\\
            ${a_{re}} \leftarrow \frac{{{a_{re}} \cdot {\beta _{re}}^{Lre}}}{{{a_{re}} \cdot {\beta _{re}}^{Lre} + {a_{pre}} \cdot {\beta _{pre}}^{Lpre}}}$, and ${a_{pre}} \leftarrow \frac{{{a_{pre}} \cdot {\beta _{pre}}^{Lpre}}}{{{a_{re}} \cdot {\beta _{re}}^{Lre} + {a_{pre}} \cdot {\beta _{pre}}^{Lpre}}}$
		\STATE Calculate the total loss ${L_{total}} = {a_{re}}{L_{re}}({x_t},{\hat x_t}) + {a_{pre}}{L_{pre}}({y_t},{\hat y_t})$
		\STATE Update the other parameters based on error back propagation ${\Omega ^{t + 1}} \leftarrow {\Omega ^t} - \eta {\nabla _{{\Omega ^t}}}{L_{total}}$	
		\STATE  Calculate the ${\Theta ^{t + 1}} \leftarrow {\Theta ^t} - \eta {\nabla _{{\Theta ^t}}}{L_{total}}$
		\STATE \textbf{Until} process all the data in turn		
	\end{algorithmic}
\end{algorithm}

\subsection{Objective Function}

In this section, we would introduce our objective function, which is divided into two components: the prediction loss ${L_{pre}}({y_t},{\hat y_t})$ and reconstructing loss ${L_{re}}({x_t},{\hat x_t})$. The prediction loss${L_{pre}}({y_t},{\hat y_t})$  between the output’s prediction value (label) and the true value (label) is represented by the cross-entropy function:

\begin{equation}
{L_{pre}}({y_t},{\hat y_t}) =  - \sum {{y_t}\log ({{\hat y}_t})} 
\end{equation}

Meanwhile,considering reconstruction loss ${L_{re}}({x_t},{\hat x_t})$, which is introduced in Section 3.2, we could project data to a hidden space that is of lower dimensionality, and therefore make the learning representation more compact and more powerful, fully exploit existing underlying information of input, in the nutshell, our goal is to predict output labels more accurately, at the same time, learn a good implicit representation, so we define the following loss function as the whole objective function:

\begin{equation}
{L_{total}} = {a_{re}}{L_{re}}({x_t},{\hat x_t}) + {a_{pre}}{L_{pre}}({y_t},{\hat y_t})
\end{equation}

where ${a_{re}}$ and ${a_{pre}}$ represent the trade-off parameters between the reconstructing loss and prediction loss, respectively, where ${a_{re}} \leftarrow \frac{{{a_{re}} \cdot {\beta _{re}}^{Lre}}}{{{a_{re}} \cdot {\beta _{re}}^{Lre} + {a_{pre}} \cdot {\beta _{pre}}^{Lpre}}}$ and ${a_{pre}} \leftarrow \frac{{{a_{pre}} \cdot {\beta _{pre}}^{Lpre}}}{{{a_{re}} \cdot {\beta _{re}}^{Lre} + {a_{pre}} \cdot {\beta _{pre}}^{Lpre}}}$ respectively. ${\beta _{re}} \in (0,1)$ and ${\beta _{pre}} \in (0,1)$ is the discount rate parameter,${L_{re}} \in (0,1)$ and ${L_{pre}} \in (0,1)$. Hence, trade-off parameters of the reconstruction loss and the prediction loss is discounted by a factor of ${\beta _{re}}^{Lre}$ and ${\beta _{pre}}^{Lpre}$ respectively in every iteration. Then, the trade-off parameters are normalized by dividing the sum of these two trade-off parameters.

\section{Experimental Results}

A good online deep learning needs to meet the following requirements \cite{33pears2014detecting,34pesaranghader2018reservoir,35bifet2017classifier,36vzliobaite2015towards}: In order to improve the accuracy of the algorithm, we need to design an appropriate total loss function, and constantly update the model parameters according to the changing data; The algorithm must have low false positive rate and negative rate and reduce the gap between the prediction label and the true label; Robustness to noise,i.e., the algorithm must be able to distinguish whether the probability distribution of data changes or appears noise in the data, and avoid confusion between them. So the model parameters can be updated correctly to achieve the ideal prediction effect. Hence, in this section, we carry out three experiments to validate the effectiveness of our algorithm. In the first group of experiments, we compared the different trade-off parameters between the reconstructing and prediction loss on ODLAE-1 and ODLAE-2. In the second group of experiments, different algorithms on seven datasets through four evaluation criteria are conducted.The four evaluation criteria are accuracy, precision, F1, and haming loss respectively. The comparative experimental results are obtained by running the open source code in the same computing environment. In order to make the experimental results more objective and fair, we have run each group of experiments ten times. In the third group of experiments, we further improve the robustness of the algorithm by adding the denoising autoencoder to the original two fusion strategies, and propose the new algorithm, named Online Deep Learning based on Denoising AutoEncoder(ODLDAE). 

\subsection{Implementation Details}

AOILAE starts the learning process by establishing an autoencoder as the underlying structure. We use fully-connected layers as the encoder and decoder, Relu is used as the activation function for each hidden layer. The entire network parameters are updated using the Adam optimizer with a learning rate of 0.01\cite{37kingma2015adam}. In the feature-level fusion part, the parameter ${d_a}$ for self-attention mechanism is a user-defined parameter, in our experiments, we set it to 30 to get good results.

\subsection{Description of the Dataset} 

\begin{table}[]
	\centering
	\caption{seven datasets with different scales, feature dimensions, and classes.}\begin{tabular}{cccccc}
		\hline
		ID  &Datasets &Instances &Attributes	&Class &Type \\
           D1  &forestcovtype	&581012	&54	&7	&Non-stationary \\
           D2  &gesture	&9873	      &50	&5	&Stationary \\
           D3	&mnist	&70000	&784	&10	&Stationary \\
           D4	&rotatedmnist	&65000	&784	&10	&Non-stationary \\
           D5	&permutedmnist	&70000	&784	&10	&Non-stationary \\
           D6	&rfid	           &28000	&3	&4	&Stationary \\
           D7	&N-BaloT	&712808	&115	&11	&Stationary \\
           \hline
	\end{tabular}
\end{table}

To verify the effectiveness and evaluate the performance of the proposed algorithm, we conduct various experiments on stationary datasets and non-stationary datasets. In this section, seven different datasets are chosen to verify our algorithm, which varies in the scales, feature dimensions, and the number of classes. The attributes of the datasets are shown in Table 1, and details about these data sets are summarized as follows: 

(1) forestcovtype dataset\cite{38blackard1999comparative}: The classification task on this data is to predict forest cover types only from cartographic variables without remote sensing data, the actual forest cover type was provided by Resource Information System (RIS) of US Forest Service (USFS) Region 2. The data is in its original form (non-scale) and contains binary (0 or 1) columns of data with qualitative independent variables, such as wilderness area and soil type. Because the input distribution would change with time, the data contains covariate drift and belongs to non-stationary data.

(2) gesture dataset\cite{39madeo2016gesture}: This dataset is composed of seven features extracted from gesture videos to study the phase segmentation of gestures. Each video is represented by two files: an original file that contains the location of the user's hands, wrists, head, and spine in each frame; and a processed file that contains the velocity and acceleration of the hand and wrist. 

(3) mnist dataset\cite{40lecun1998gradient}: This dataset contains 70K samples. These numbers are dimensionally standardized and located in the center of the image, which is of a fixed size (28x28 pixels) with a value between 0 and 1. For simplicity, each image is flattened and converted into a one-dimensional numpy array of 784 (28 * 28) features.

(4) rotatedmnist dataset\cite{41lopez2017gradient}: By rotating the original sample, the extension of the traditional mnist problem\cite{40lecun1998gradient} is formed, which leads to the abrupt drift of the concept. Specifically, handwritten digits are rotated to any angle in the range of -$\pi$ to $\pi$, resulting in covariate drift. 

(5) permutedmnist dataset\cite{15kirkpatrick2017overcoming}: It is a modification of the MNIST datasets \cite{18bengio2007greedy} which performs the permutations of pixels in mnist after quantization, in fact, it uses a group of random indexes to scramble the position of each element in the vector, and different random indexes produce different tasks. Hence, the real drift \cite{42gama2008knowledge} is present in this dataset. 

(6) rfid dataset\cite{43ashfahani2020devdan}: Indoor RFID location data is a multi-class problem, which aims to identify the location of objects in the manufacturing workshop by using three input attributes. RFID readers are placed in different locations and four areas are created in the manufacturing workshop, resulting in four categories problems. 

(7)N-Balo dataset\cite{44meidan2018n}: This dataset solves the lack of public botnet datasets, especially the Internet of things. It shows the real traffic data collected from 9 Commercial Internet of things devices. Because the malicious data can be divided into 10 kinds of attacks carried by two botnets and one kind of benign attack. Therefore, it belongs to multi-category data set.  

\subsection{Comparing Different Trade-off parameters for the reconstructing loss and prediction loss on ODLAE-1}

ODLAE algorithm is a progressive learning approach. Because the total loss is composed of two components, namely prediction and reconstruction loss, how to reasonably allocate the proportion of these two losses in the total loss has a great impact on the accuracy rate, which can effectively ensure the communication between the old and new data, resulting positive back propagation. There are two options for us to choose from: one is fixed allocation weight value selection strategy and the other is change the weight value according to the loss of the previous moment selection strategy. As discussed in the previous section, we use the method of constantly adjusting the trade-off parameters, and the changing total loss function constantly updates the network parameters through back-propagation. 

The weight value selection strategy strategy balances the trade off between prediction loss and reconstruction loss. On the one hand, we want the model to explore and learn new knowledge as much as possible to reduce the prediction loss. On the other hand, we also hope that the model will not forget its own input characteristics and reduce the loss of reconstruction. In addition, because the probability distribution of data in the data stream may change, we need to update the total loss function dynamically according to the different characteristics of the data.

To further illustrate that the adaptive change scheme is better than the method of fixing the trade-off parameters, we designed several groups of comparative experiments. We set these two parameters to $({a_{re}},{a_{pre}})$ = (0.1,0.9), (0.2,0.8), (0.3,0.7), (0.4,0.6), ….(0.9,0.1) and the dynamic change strategy is set as $(\frac{{{a_{re}} \cdot {\beta _{re}}^{Lre}}}{{{a_{re}} \cdot {\beta _{re}}^{Lre} + {a_{pre}} \cdot {\beta _{pre}}^{Lpre}}},\frac{{{a_{pre}} \cdot {\beta _{pre}}^{Lpre}}}{{{a_{re}} \cdot {\beta _{re}}^{Lre} + {a_{pre}} \cdot {\beta _{pre}}^{Lpre}}})$ respectively.

\begin{figure}[htbp]
\centering
\subfigure[forestcovtype]
{
    \begin{minipage}[b]{.45\linewidth}
        \centering
        \includegraphics[scale=0.35]{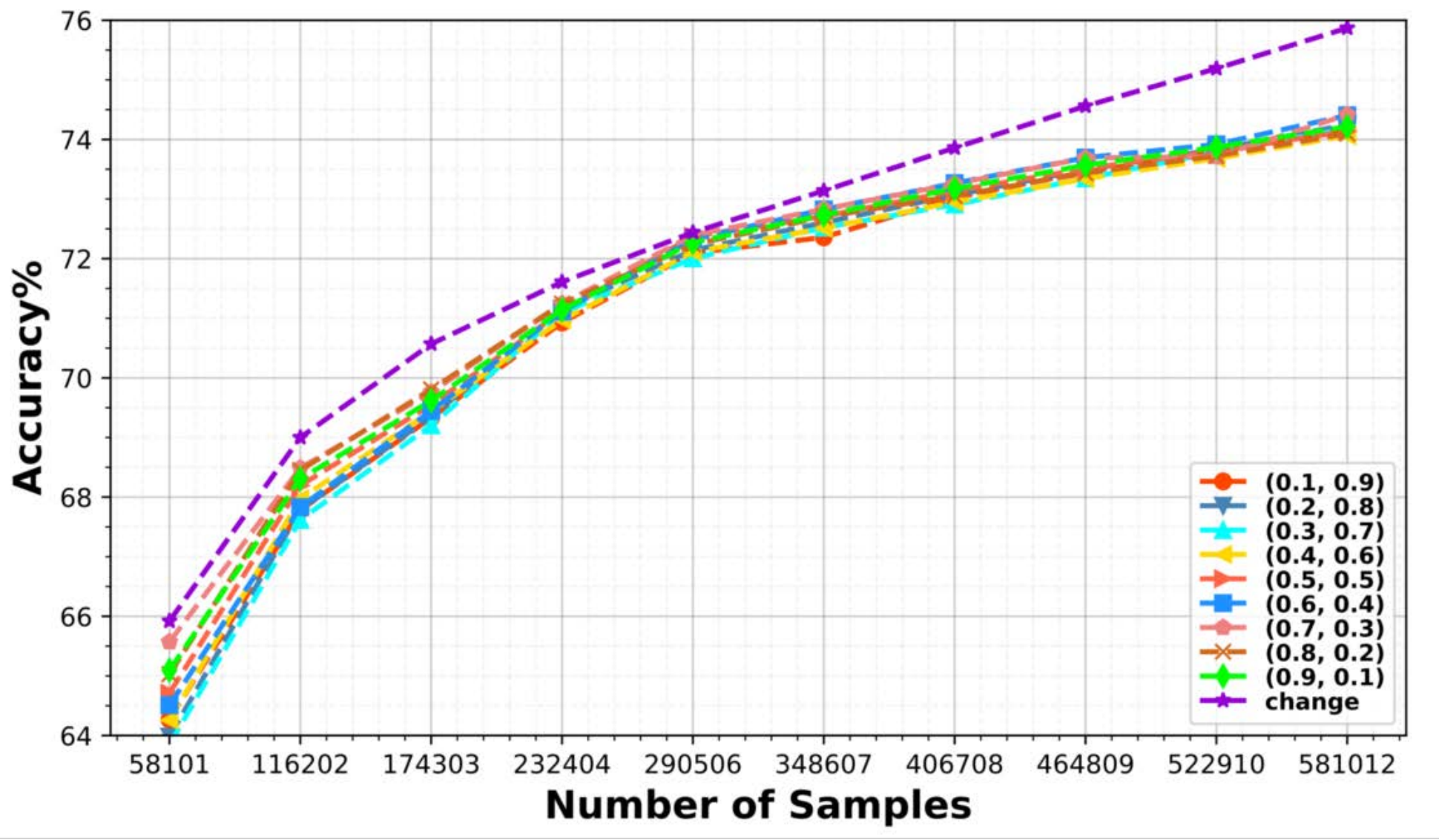}
    \end{minipage}
}
\subfigure[gesture]
{
 	\begin{minipage}[b]{.45\linewidth}
        \centering
        \includegraphics[scale=0.35]{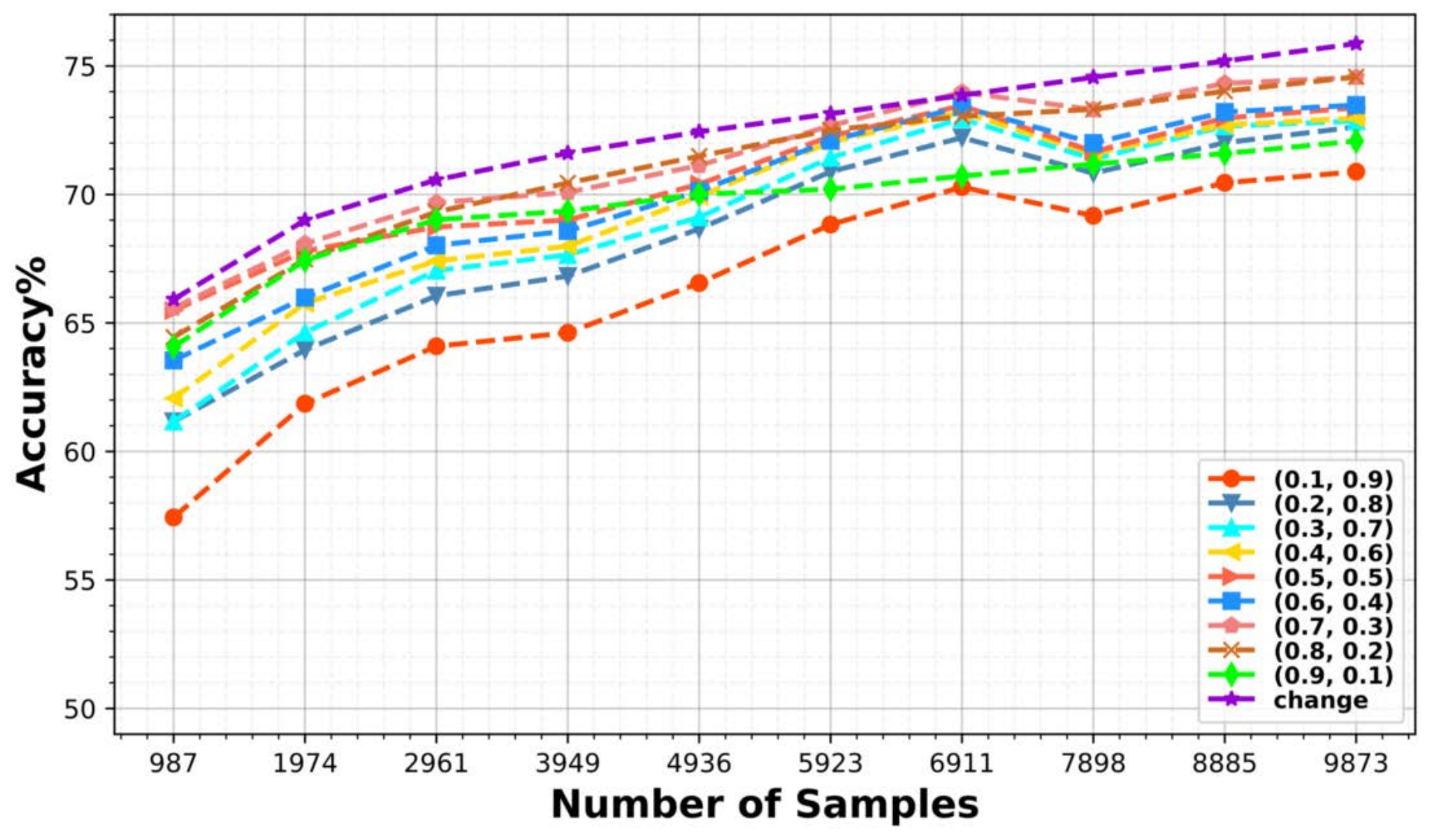}
    \end{minipage}
}
\subfigure[mnist]
{
 	\begin{minipage}[b]{.45\linewidth}
        \centering
        \includegraphics[scale=0.35]{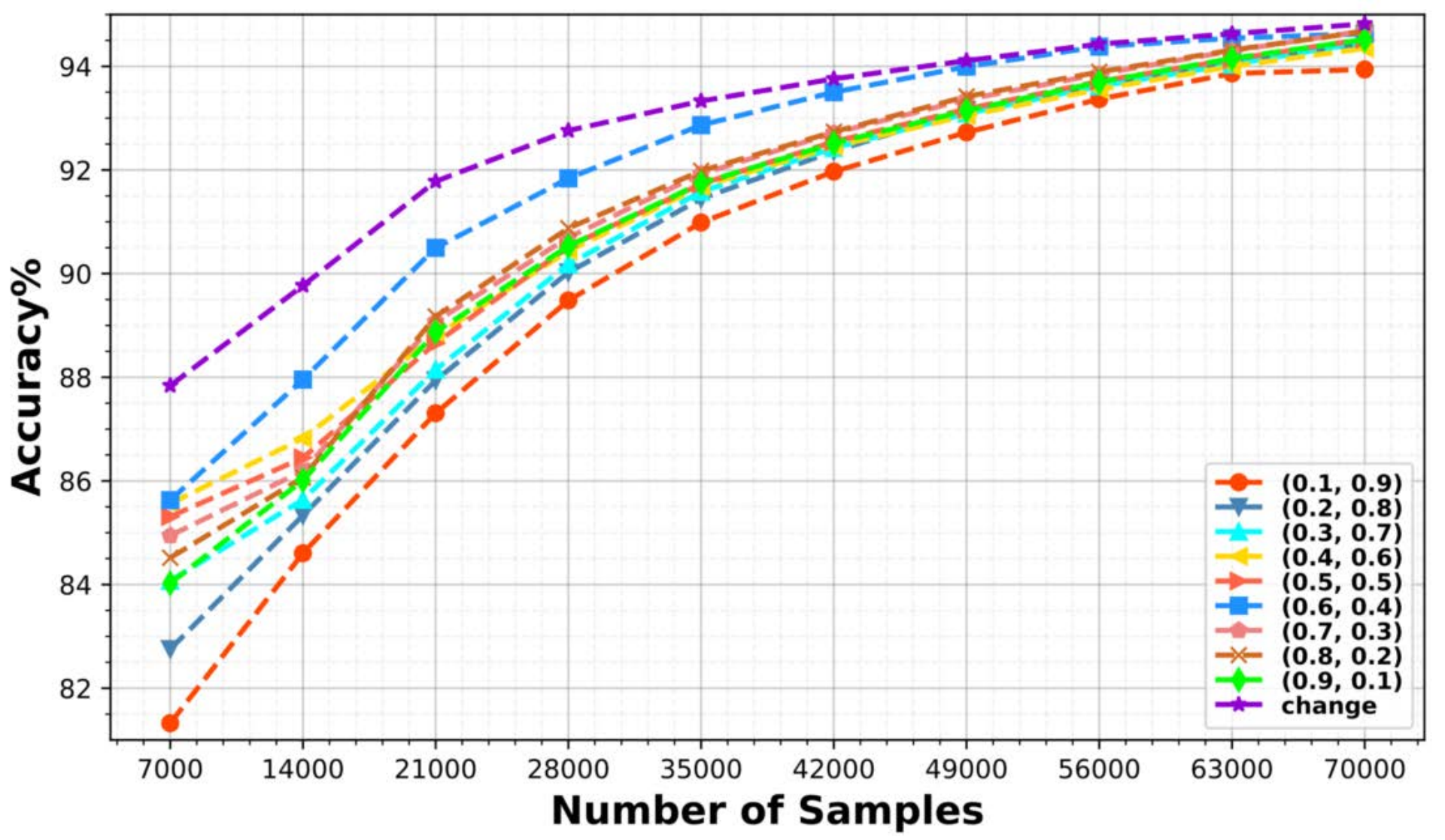}
    \end{minipage}
}
\subfigure[rotatedmnist]
{
 	\begin{minipage}[b]{.45\linewidth}
        \centering
        \includegraphics[scale=0.35]{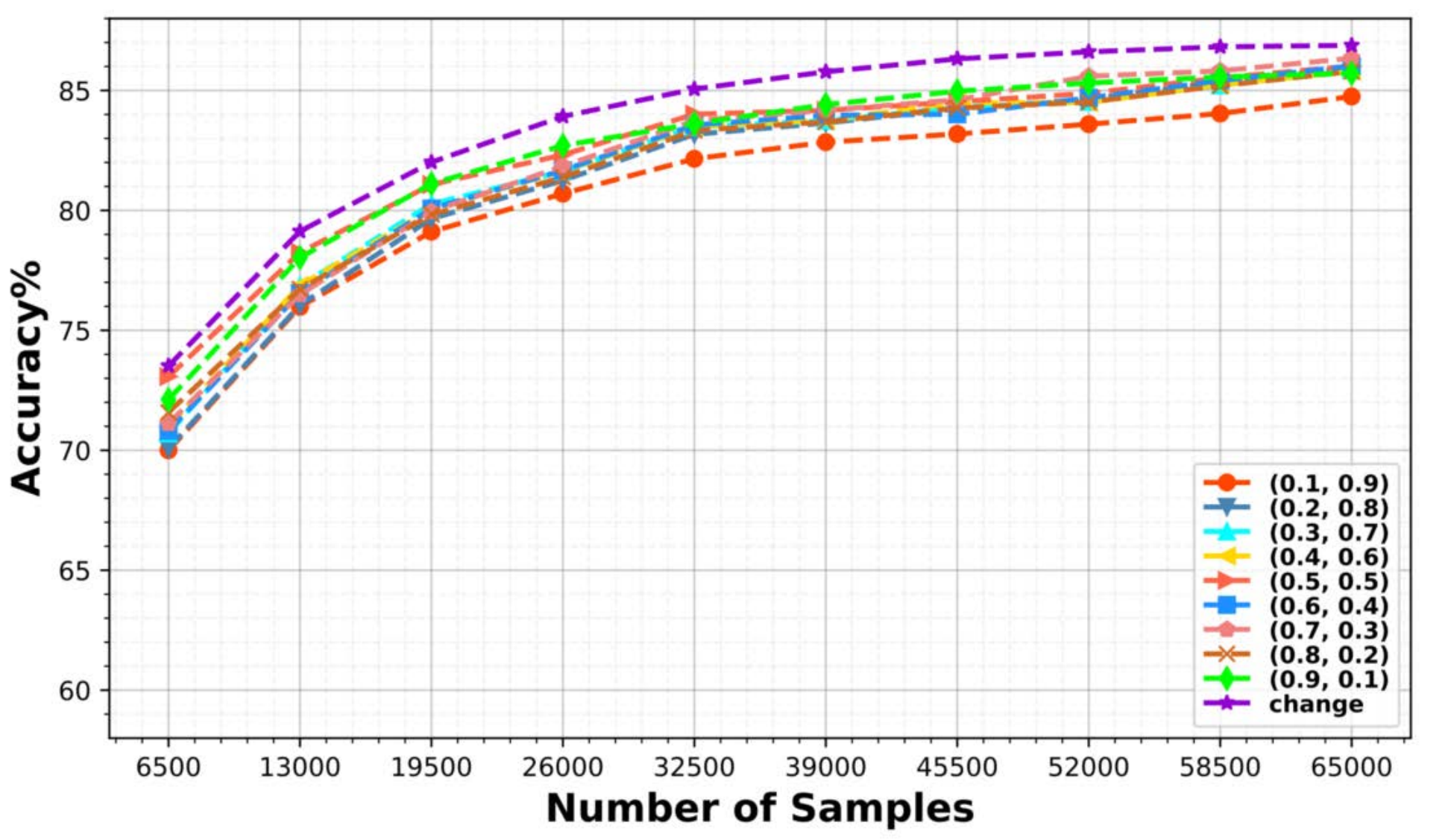}
    \end{minipage}
}
\subfigure[permutedmnist]
{
 	\begin{minipage}[b]{.45\linewidth}
        \centering
        \includegraphics[scale=0.35]{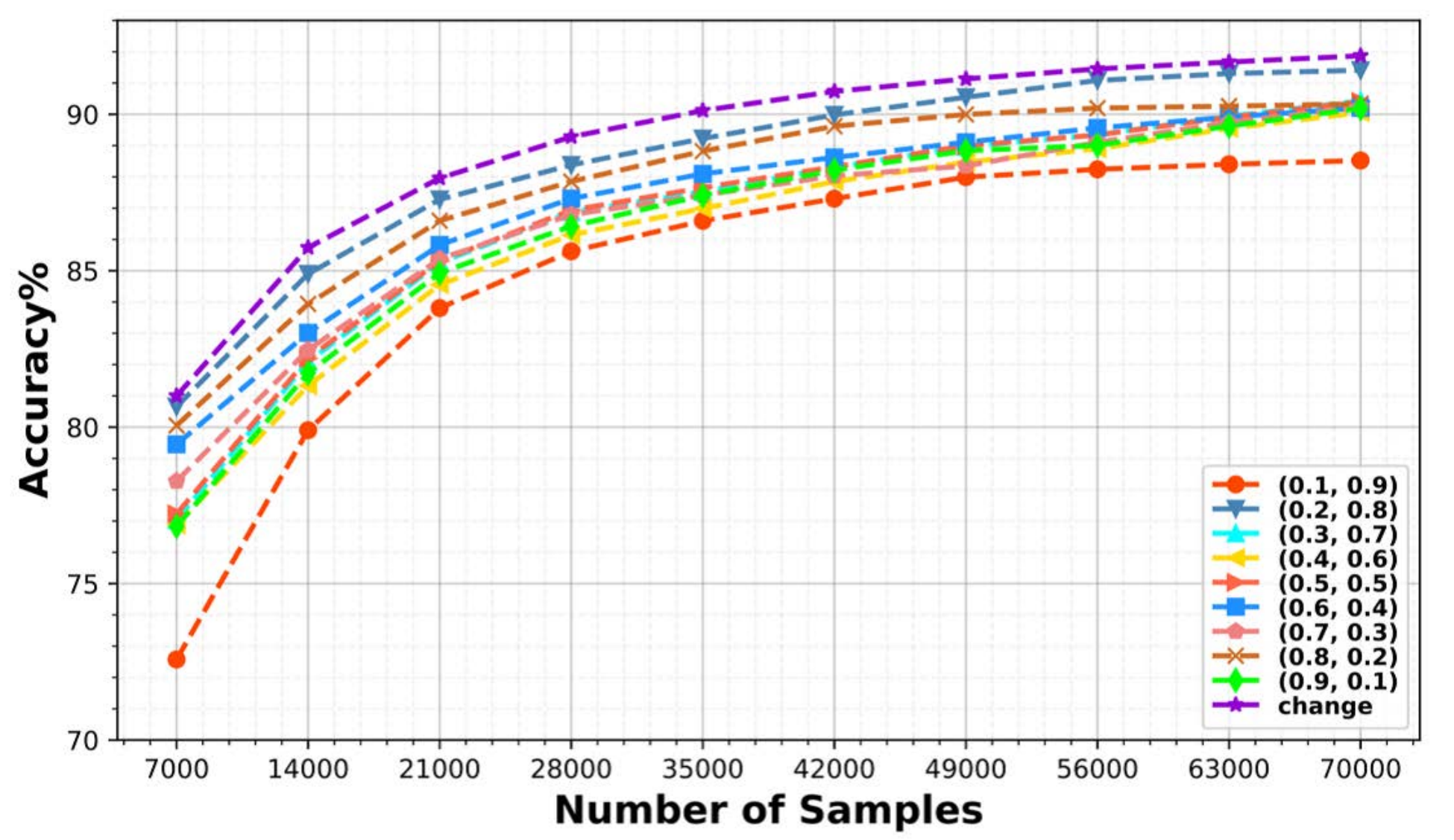}
    \end{minipage}
}
\subfigure[rfid]
{
 	\begin{minipage}[b]{.45\linewidth}
        \centering
        \includegraphics[scale=0.35]{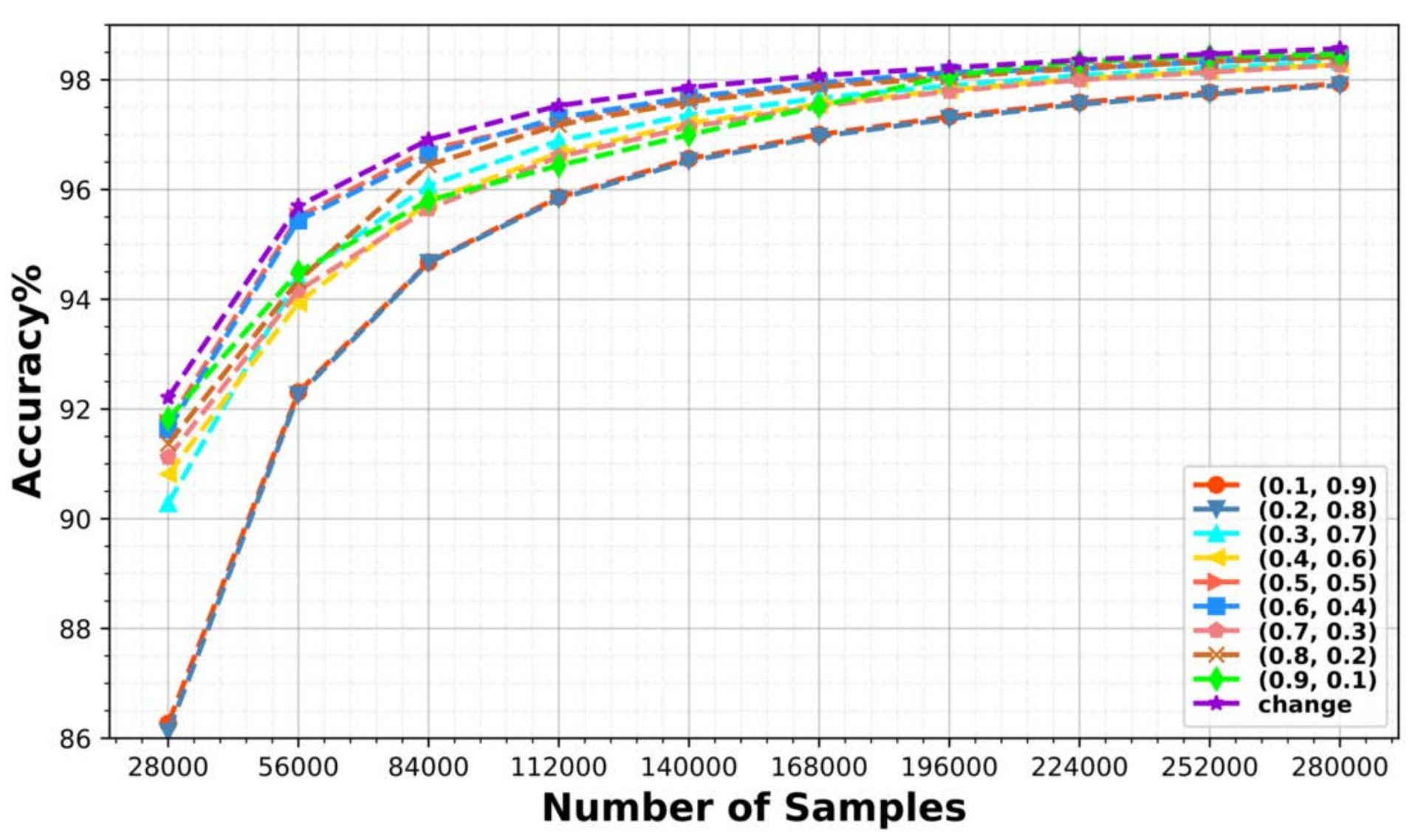}
    \end{minipage}
}
\subfigure[N-Balo]
{
 	\begin{minipage}[b]{.45\linewidth}
        \centering
        \includegraphics[scale=0.35]{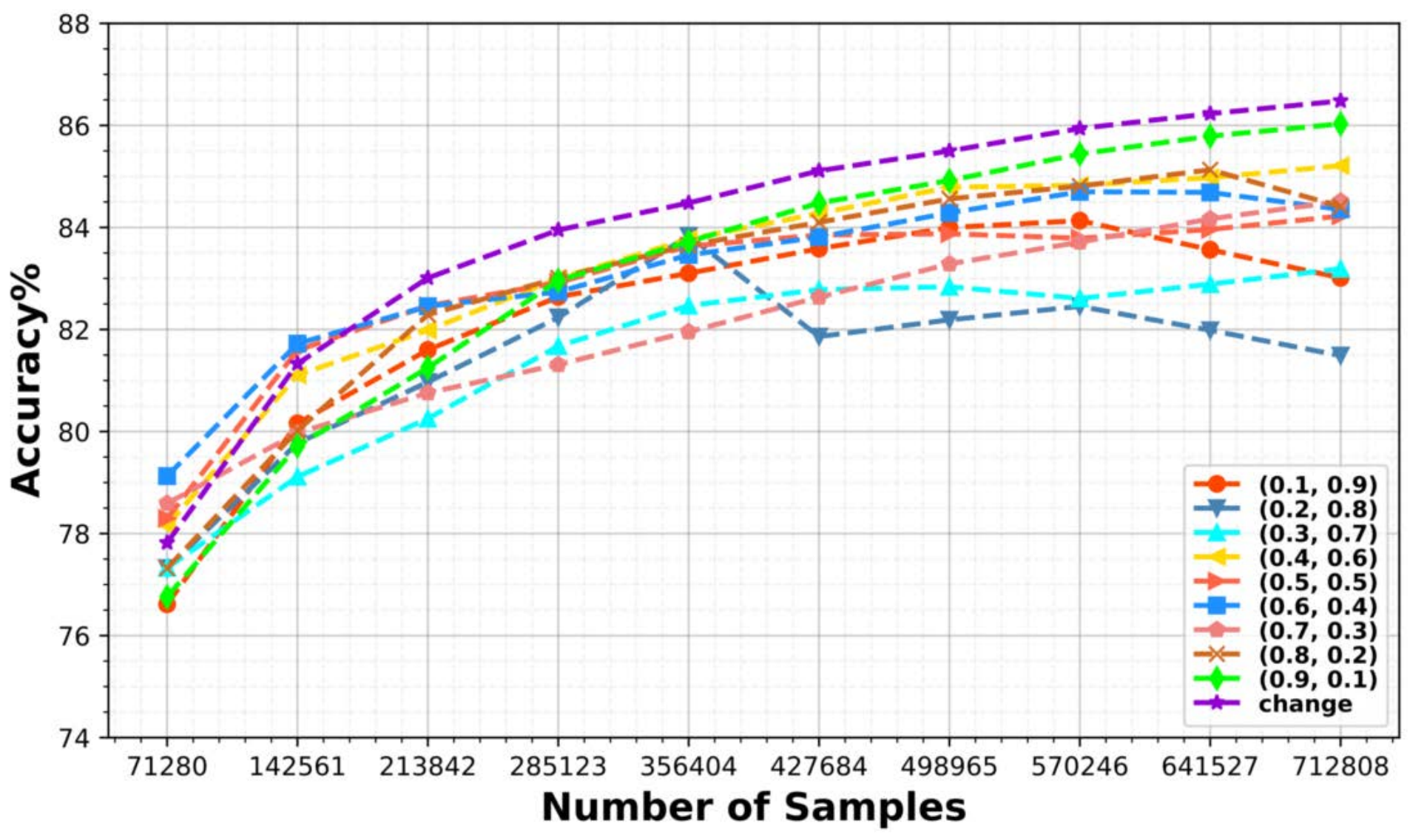}
    \end{minipage}
}
\caption{Comparing different trade-off parameters between the reconstructing and prediction loss on ODLAE-1}
\end{figure}

%
%

As we can see in Fig.3, this illustrates that our proposed tuning strategy that dynamically controls the equilibrium between prediction and reconstruction loss can help significantly reduce of the error rate of the algorithm. Specifically, the best performance is achieved by changing these two trade-off parameters adaptively. This enables the model to dynamically change the overall loss according to its continuous learning performance on input data, and then constantly update the model, improving the learning effect of the model.

\subsection{Comparing Different Trade-off Parameters for the Reconstructing Loss and Prediction Loss on ODLAE-2}

\begin{figure}[htbp]
\centering
\subfigure[forestcovtype]
{
    \begin{minipage}[b]{.45\linewidth}
        \centering
        \includegraphics[scale=0.35]{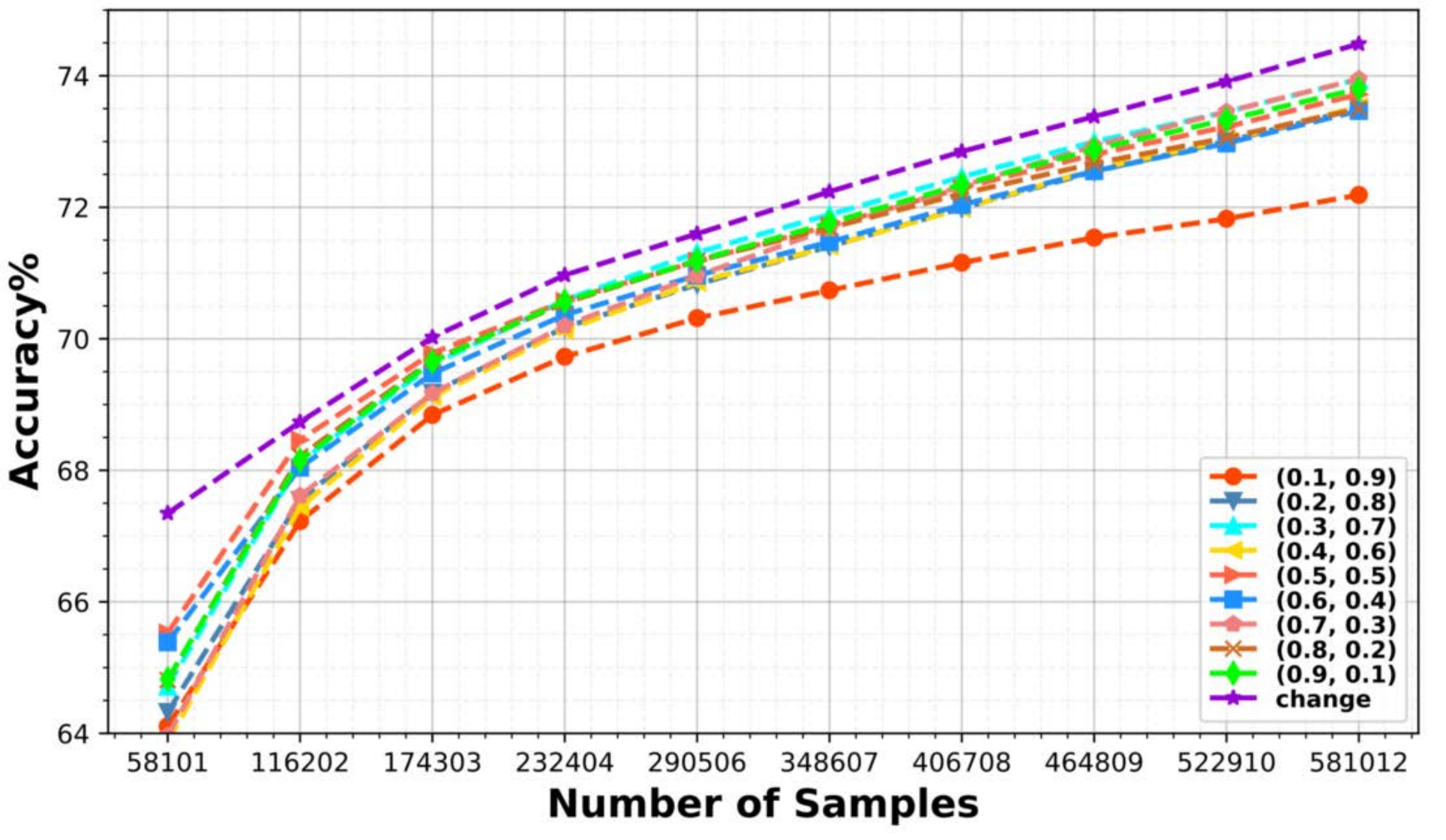}
    \end{minipage}
}
\subfigure[gesture]
{
 	\begin{minipage}[b]{.45\linewidth}
        \centering
        \includegraphics[scale=0.35]{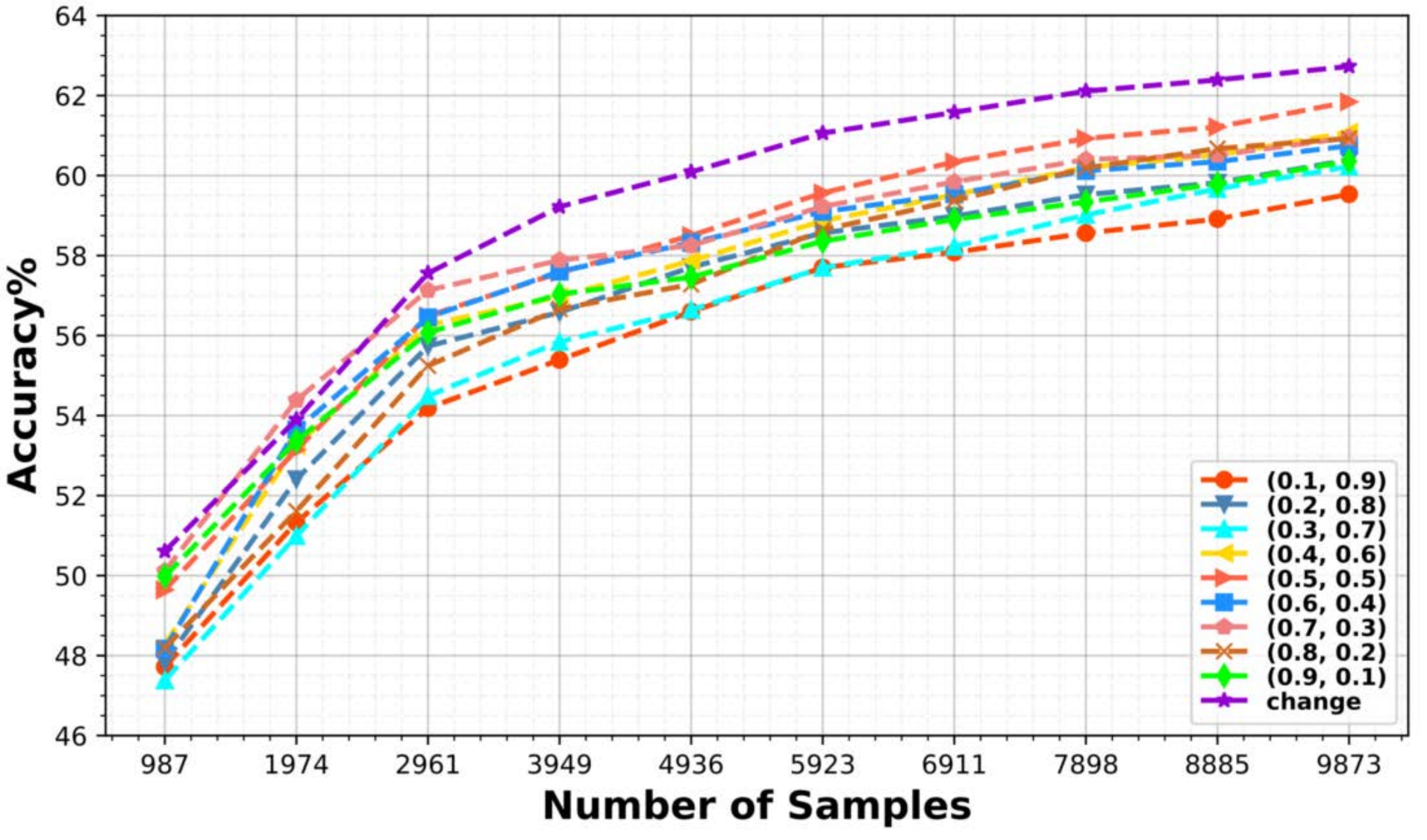}
    \end{minipage}
}
\subfigure[mnist]
{
 	\begin{minipage}[b]{.45\linewidth}
        \centering
        \includegraphics[scale=0.35]{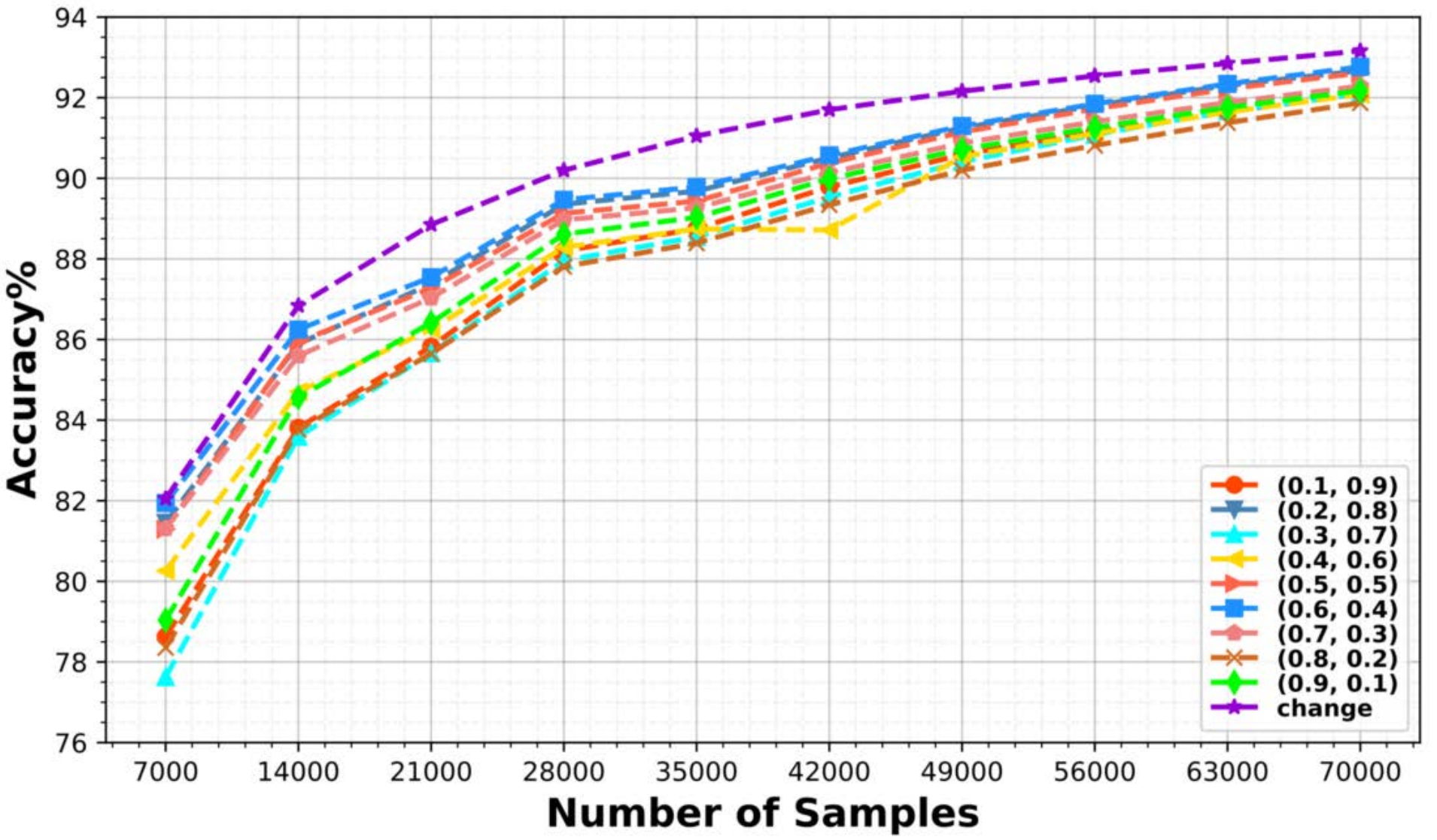}
    \end{minipage}
}
\subfigure[rotatedmnist]
{
 	\begin{minipage}[b]{.45\linewidth}
        \centering
        \includegraphics[scale=0.35]{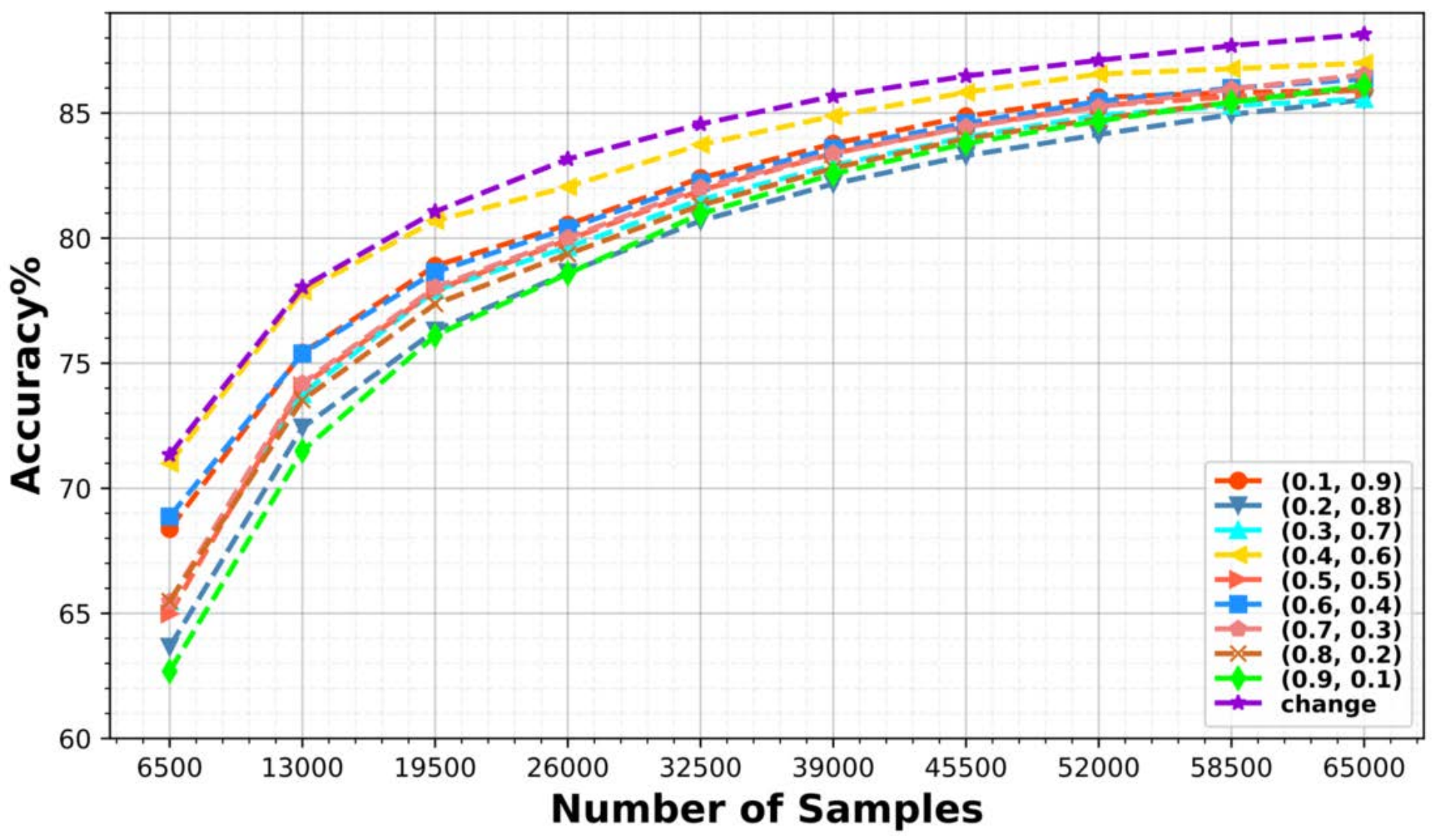}
    \end{minipage}
}
\subfigure[permutedmnist]
{
 	\begin{minipage}[b]{.45\linewidth}
        \centering
        \includegraphics[scale=0.35]{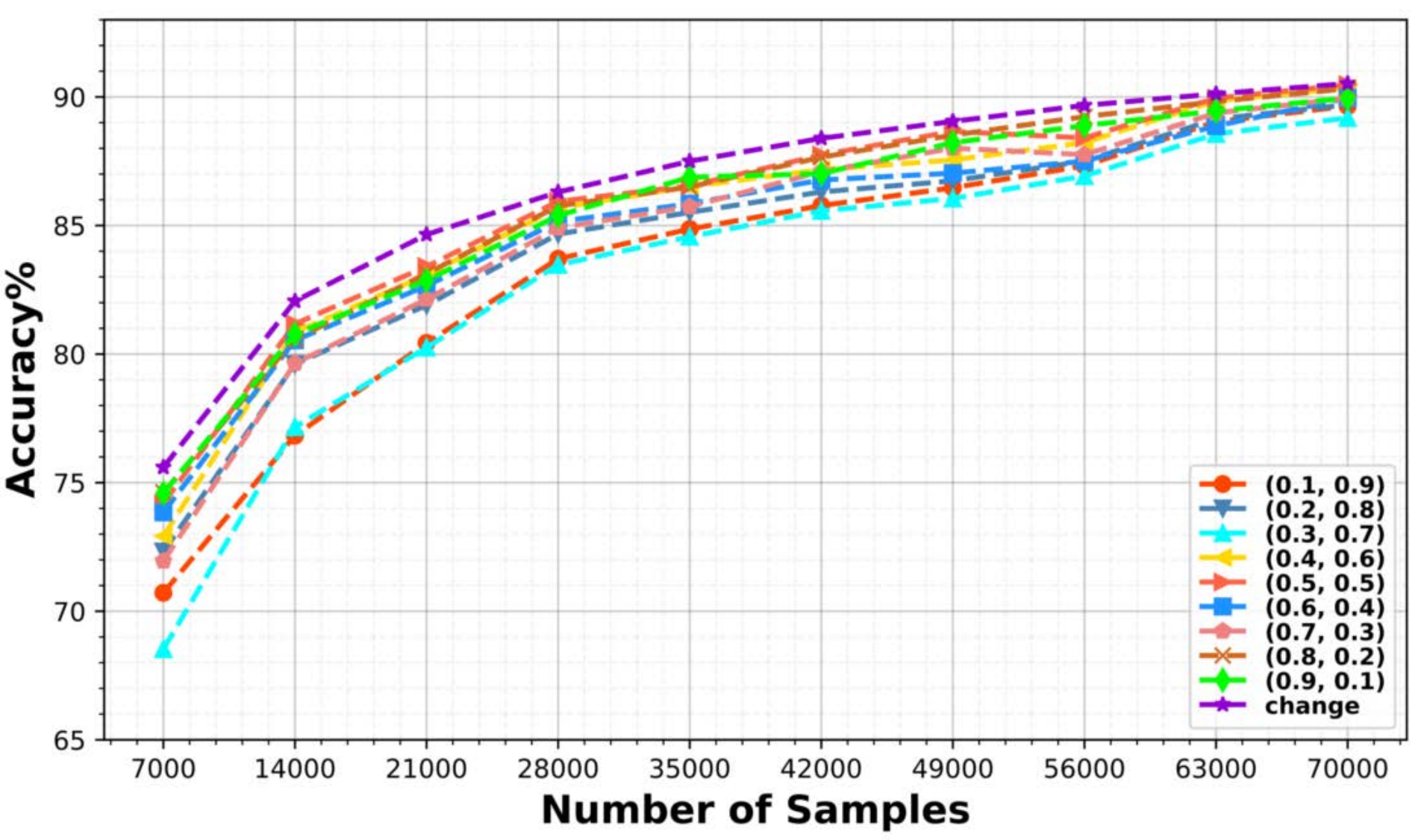}
    \end{minipage}
}
\subfigure[rfid]
{
 	\begin{minipage}[b]{.45\linewidth}
        \centering
        \includegraphics[scale=0.35]{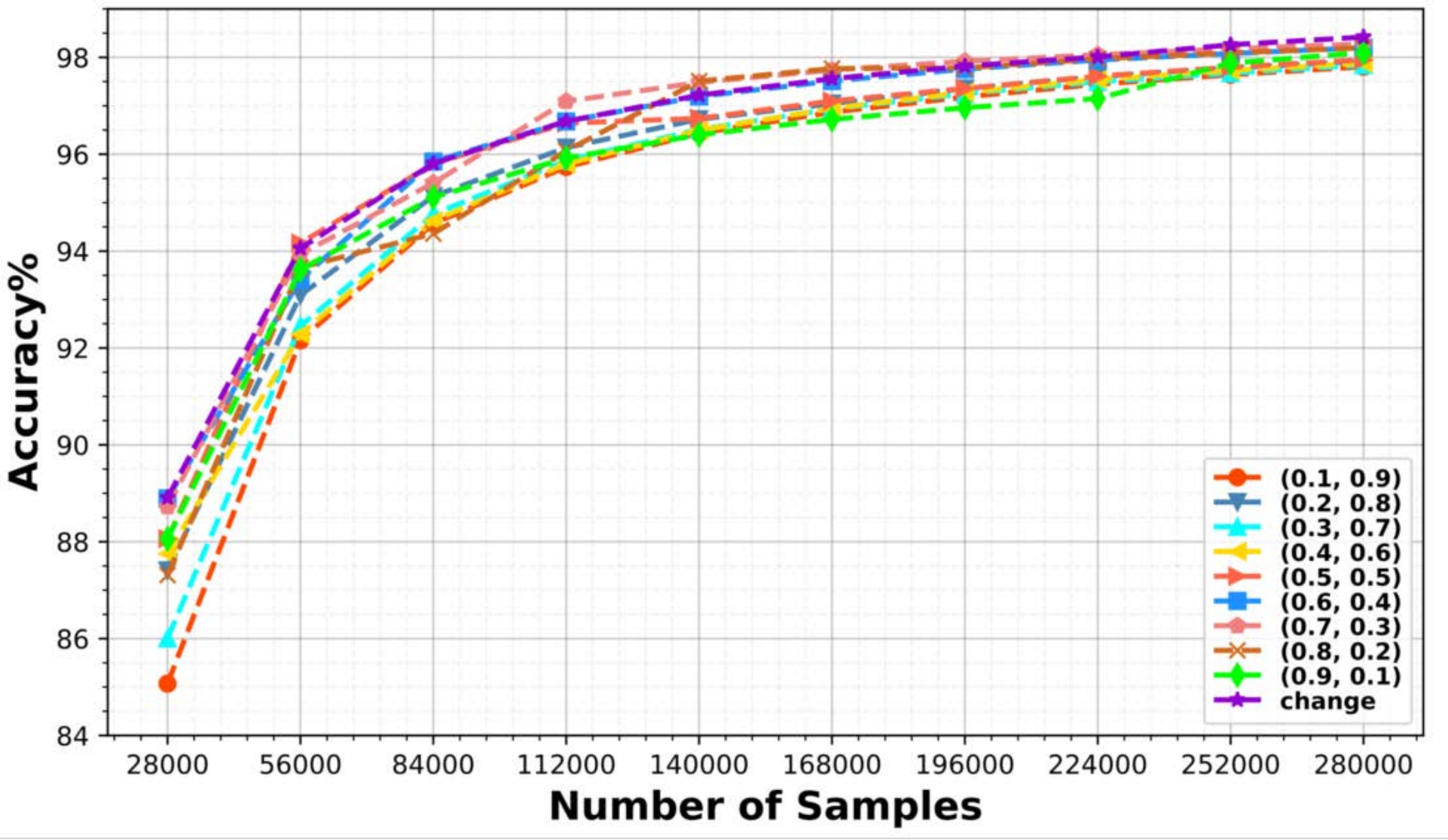}
    \end{minipage}
}
\subfigure[N-Balo]
{
 	\begin{minipage}[b]{.45\linewidth}
        \centering
        \includegraphics[scale=0.35]{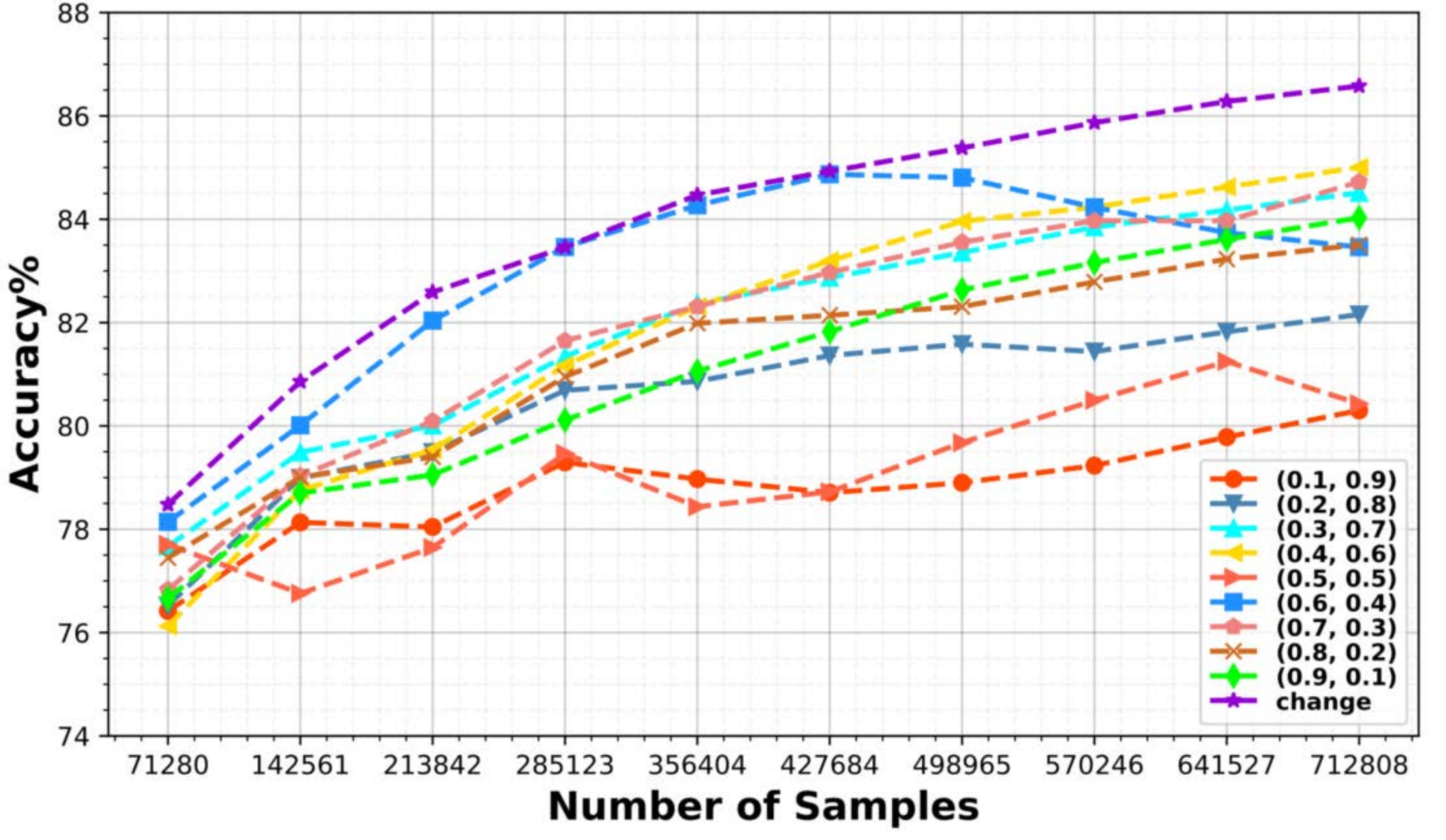}
    \end{minipage}
}
\caption{Comparing different trade-off parameters between the reconstructing and prediction loss on ODLAE-2}
\end{figure}

Similarly, we have done the same experiments on the algorithm ODLAE-2, as we can see in Fig.4, for smaller ${a_{re}}$, that is, the reconstruction loss accounts for a small proportion of the total loss, the network tends to forget its original input characteristics while learning the data, however, when we set ${a_{pre}}$ to a smaller value, it means that we rely too much on the original characteristics of data and neglect to explore new knowledge. However, we should take different measures for different data, hence, Our strategy with changing trade-off parameters is better than the general strategy with fixed trade-off parameters.

\subsection{Comparing the Performance of Different Hidden layers $L$ and Hidden units ${D_{X'}}$ of Auto-Encoder on ODLAE-1 and ODLAE-2}

Variations in $L$ and ${D_{X'}}$: $L$ represents the hidden layers and   ${D_{X'}}$ represents the hidden units of  the autoencoder. These two parameters controls the model control the complexity of the network. The hidden layers and hidden units of the network is very important to the performance of the model. When these two parameters of network layers is increased, the network can extract more complex feature patterns. Therefore, when the model is more complex, the theoretical results can be better. However, the degradation of the deep network may occur, the accuracy is saturated or even decreased.

Therefore, we use the method of super parameter retrieval to find the ideal super parameters on each dataset. First, we fix the number of hidden units of the autoencoder ${D_{X'}}$ to 32, and then increase the number of hidden layers from 2 layers to 5 layers to find out the optimal number of layers. Then we fix the optimal number of layers $L$ of the autoencoder and increase the number of hidden layer units ${D_{X'}}$ from 32, 64, 128, to 256 to find the best combination $\left( {L,{D_{X'}}} \right)$. As we can see in Fig.5, the horizontal axis represents different combinations of experiments. On forestcovtype dataset, we find the best couple $\left( {L,{D_{X'}}} \right) = (5,128)$ in ODLAE-1 algorithm, and $\left( {L,{D_{X'}}} \right) = (5,64)$ in ODLAE-2 algorithm, And these optimal parameter combinations are used as the fixed parameter settings of the follow-up experiments.


\begin{figure}[htbp]
\centering
\subfigure[Forestcovtype]
{
    \begin{minipage}[b]{.2\linewidth}
        \centering
        \includegraphics[scale=0.2]{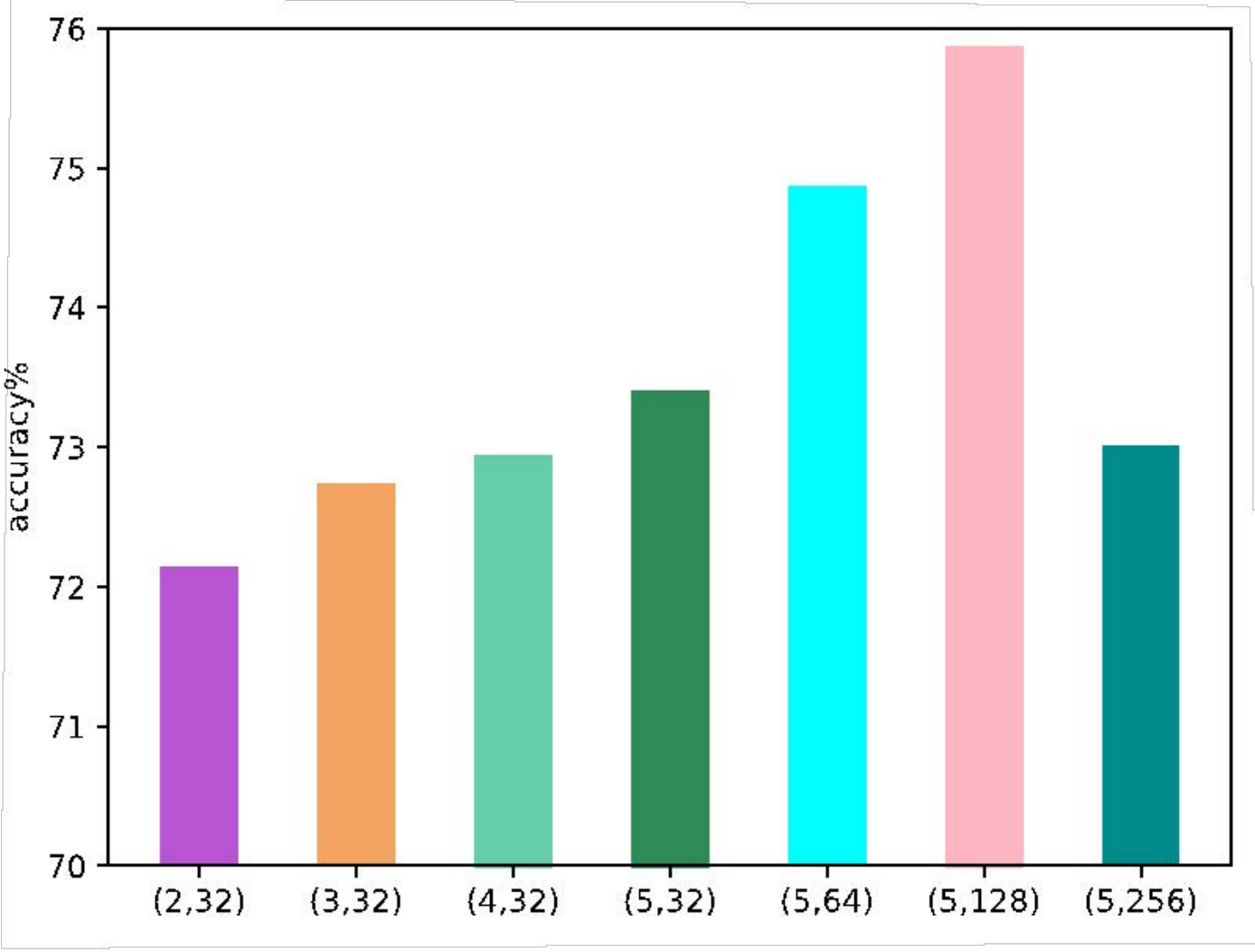}
    \end{minipage}
    \begin{minipage}[b]{.2\linewidth}
        \centering
        \includegraphics[scale=0.2]{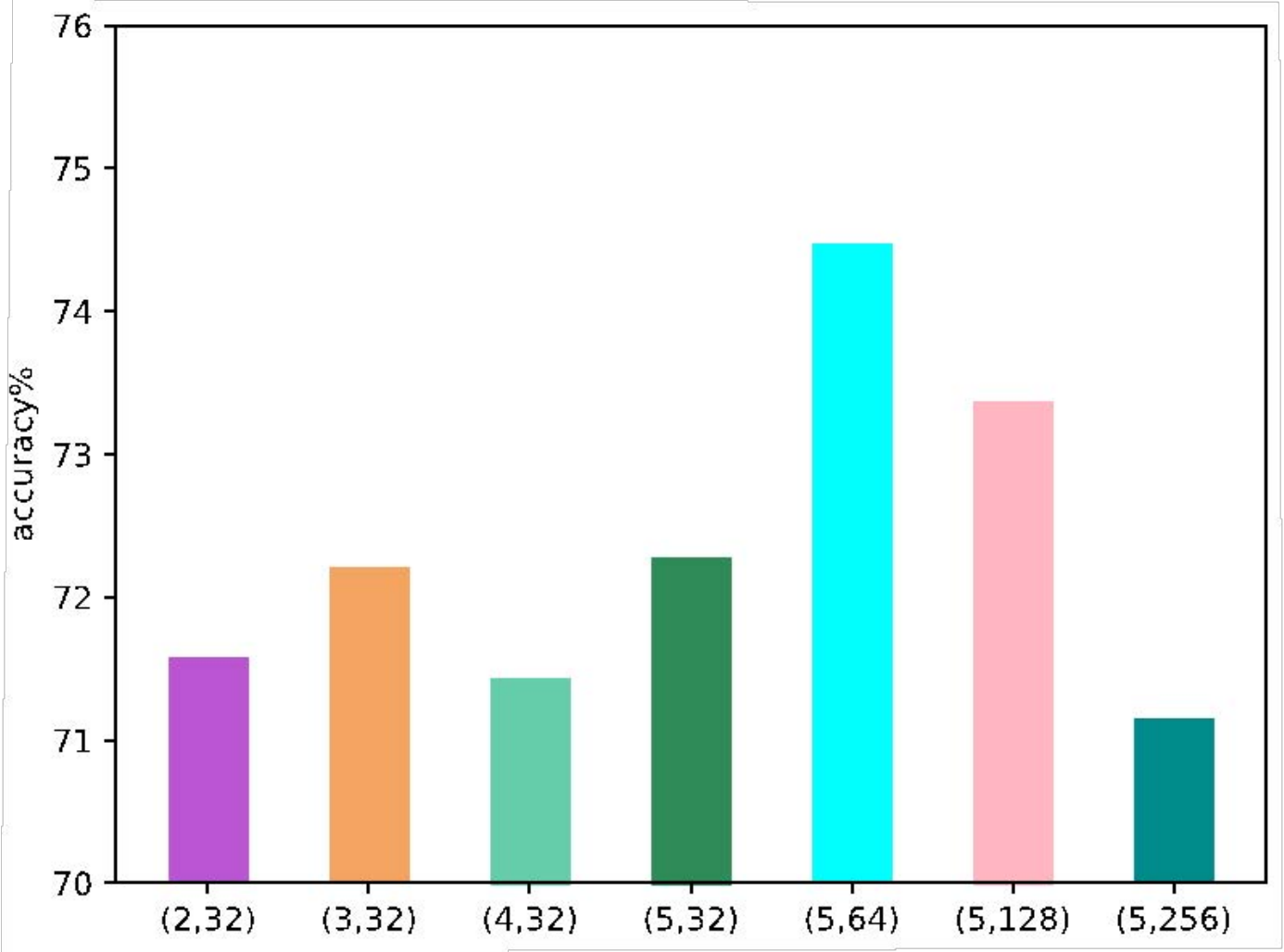}
    \end{minipage}
}
\subfigure[Gesture]
{
 	\begin{minipage}[b]{.2\linewidth}
        \centering
        \includegraphics[scale=0.2]{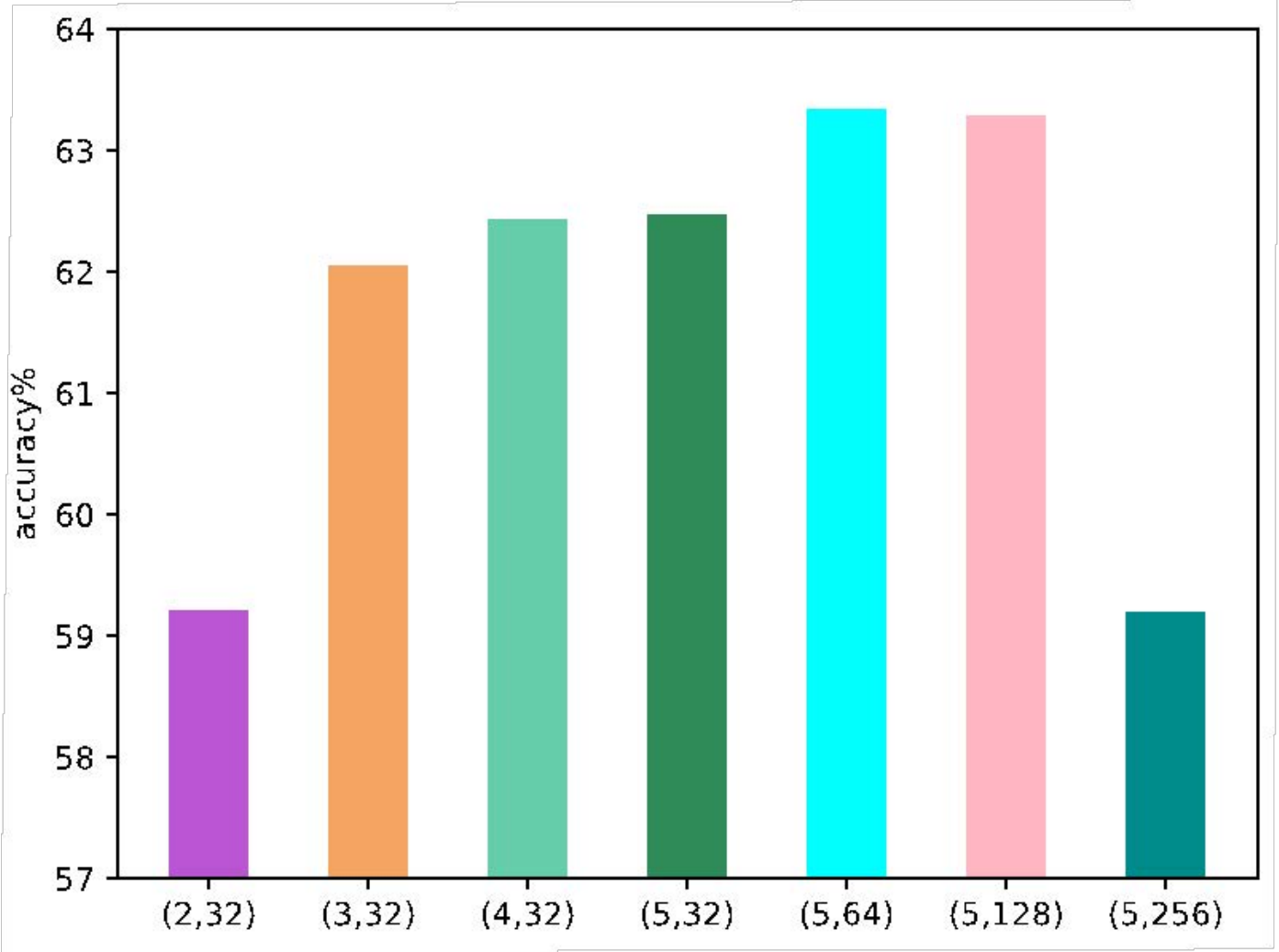}
    \end{minipage}
    \begin{minipage}[b]{.2\linewidth}
        \centering
        \includegraphics[scale=0.2]{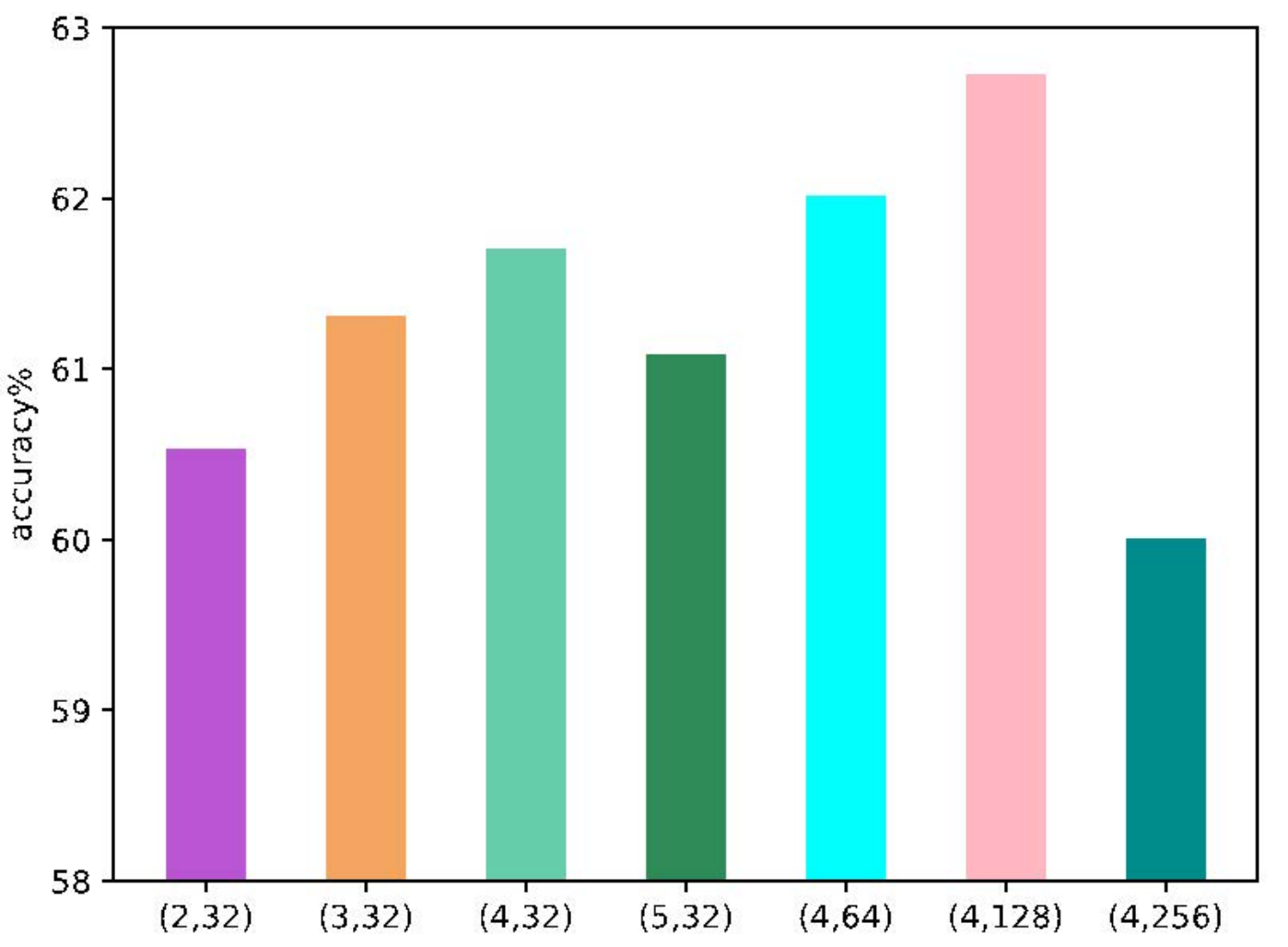}
    \end{minipage}
}
\subfigure[Mnist]
{
 	\begin{minipage}[b]{.2\linewidth}
        \centering
        \includegraphics[scale=0.2]{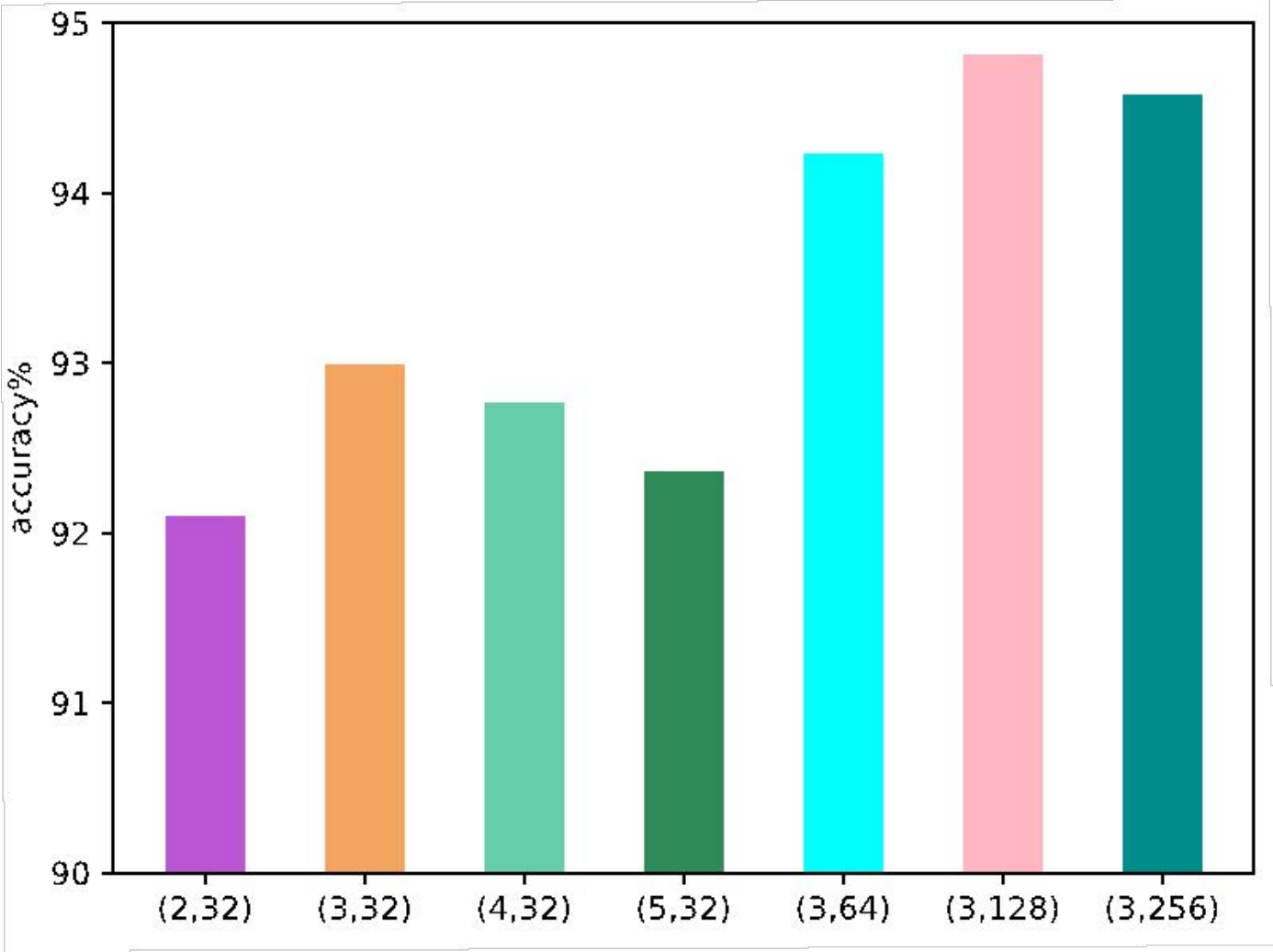}
    \end{minipage}
    \begin{minipage}[b]{.2\linewidth}
        \centering
        \includegraphics[scale=0.2]{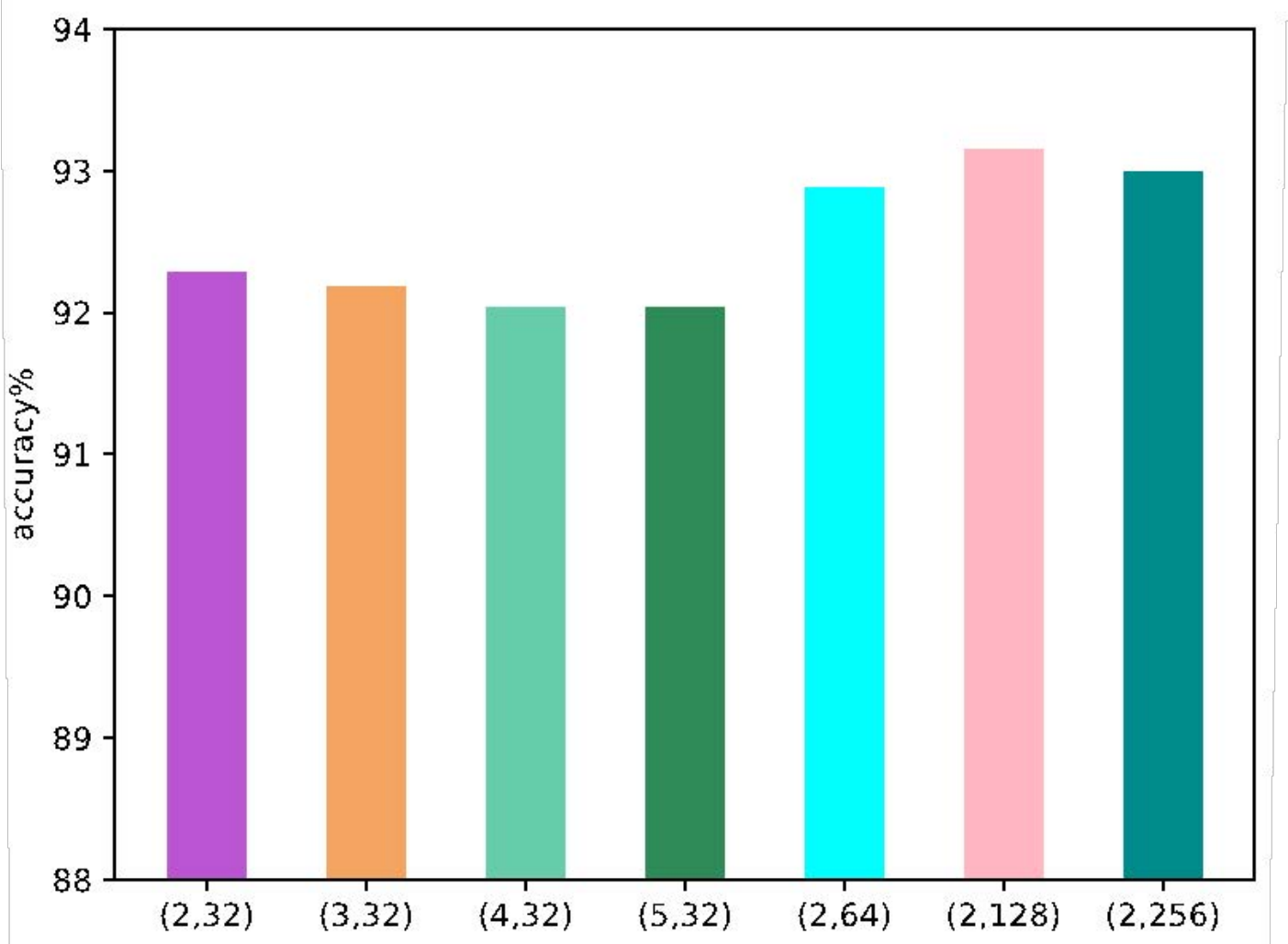}
    \end{minipage}
}
\subfigure[Rotatedmnist]
{
 	\begin{minipage}[b]{.2\linewidth}
        \centering
        \includegraphics[scale=0.2]{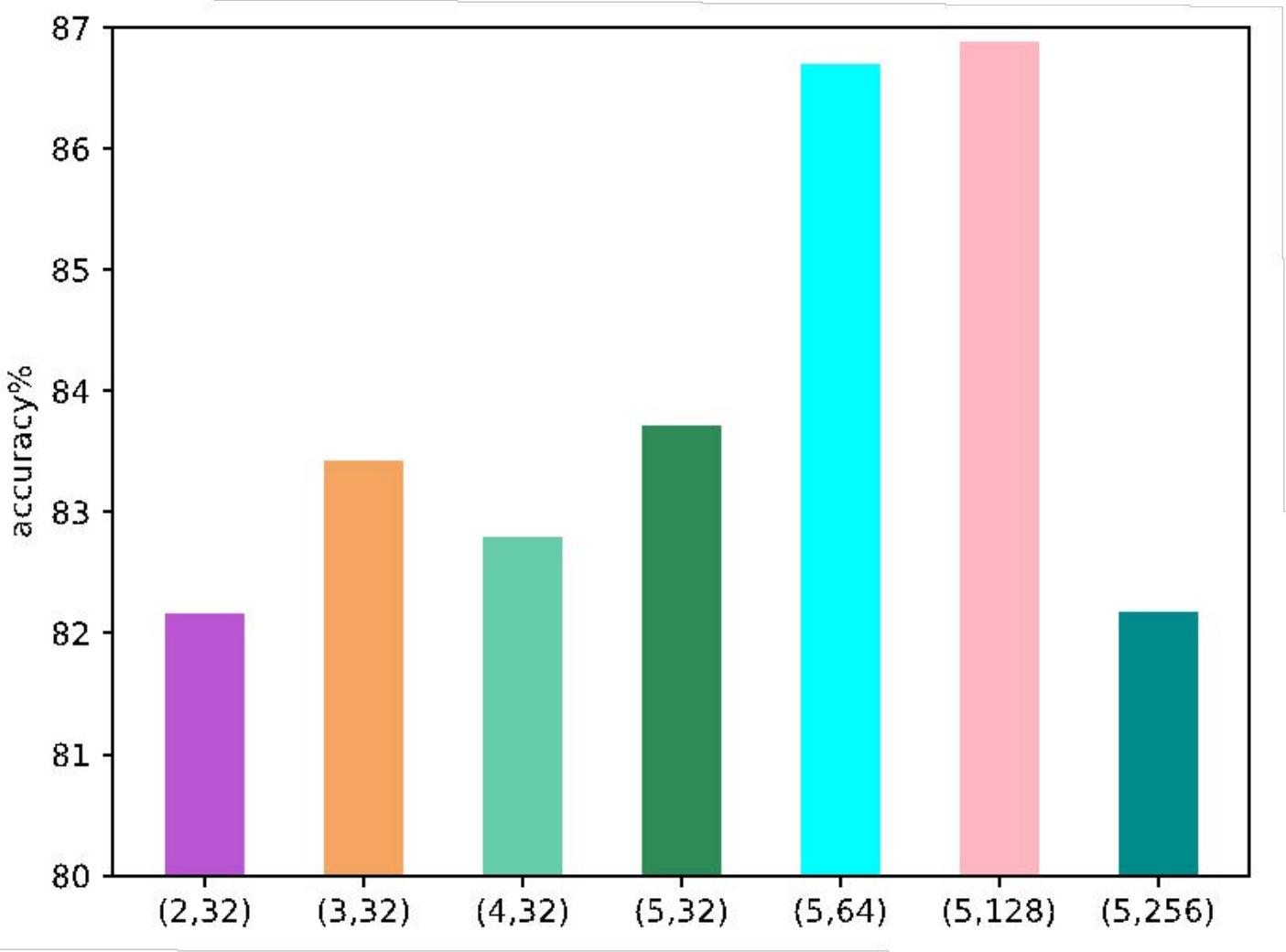}
    \end{minipage}
    \begin{minipage}[b]{.2\linewidth}
        \centering
        \includegraphics[scale=0.2]{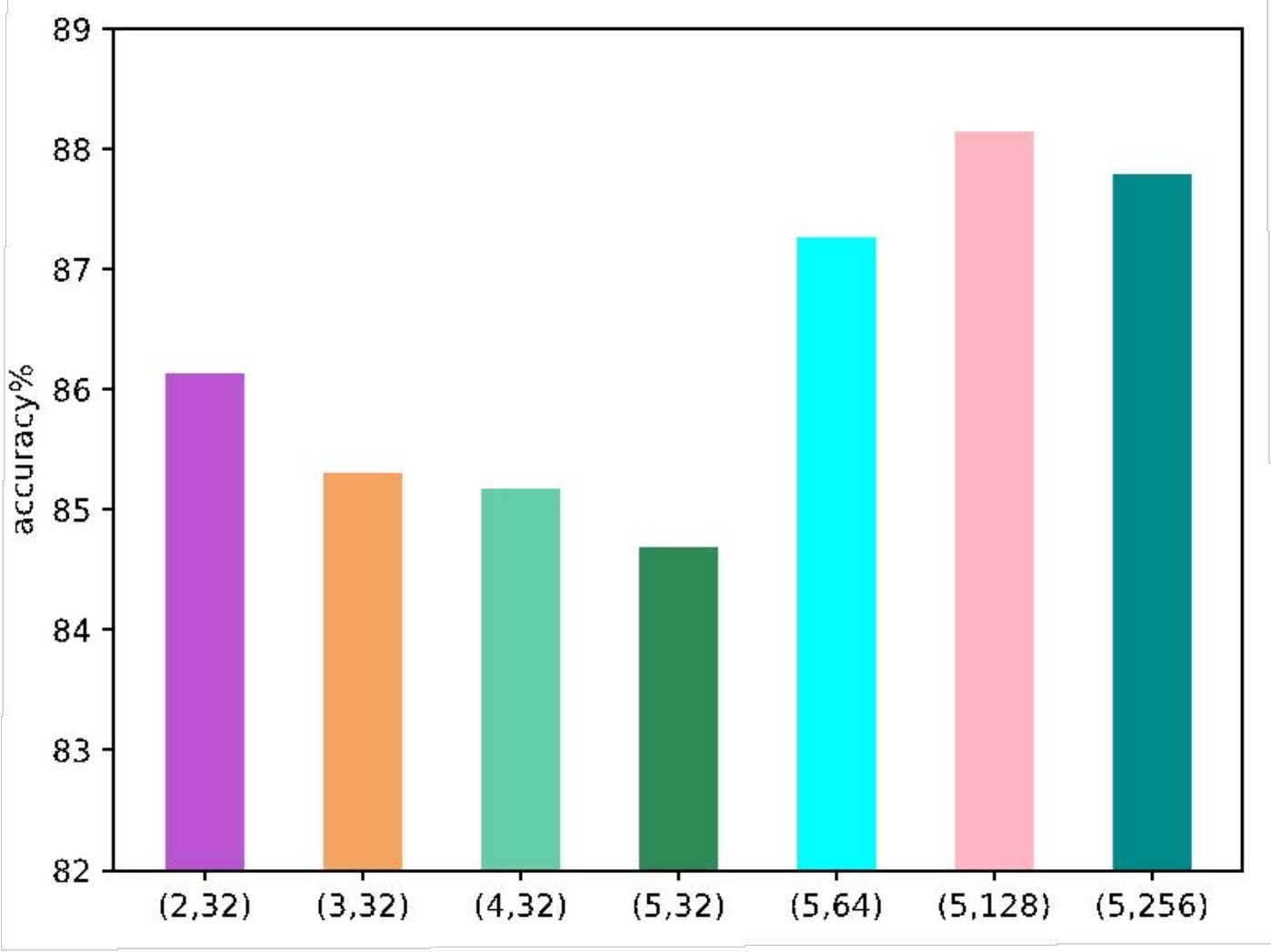}
    \end{minipage}
}
\subfigure[permutedmnist]
{
 	\begin{minipage}[b]{.2\linewidth}
        \centering
        \includegraphics[scale=0.2]{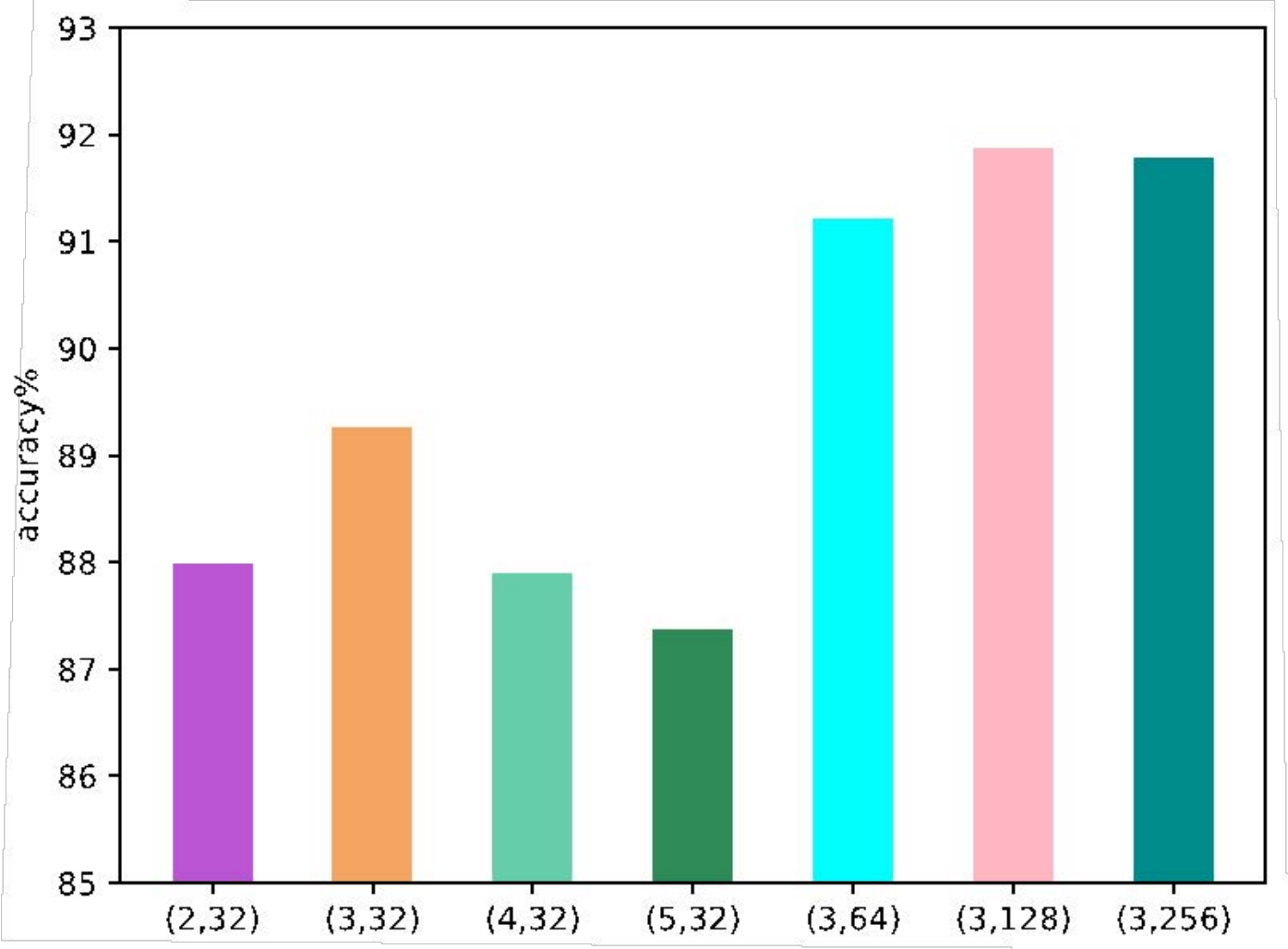}
    \end{minipage}
    \begin{minipage}[b]{.2\linewidth}
        \centering
        \includegraphics[scale=0.2]{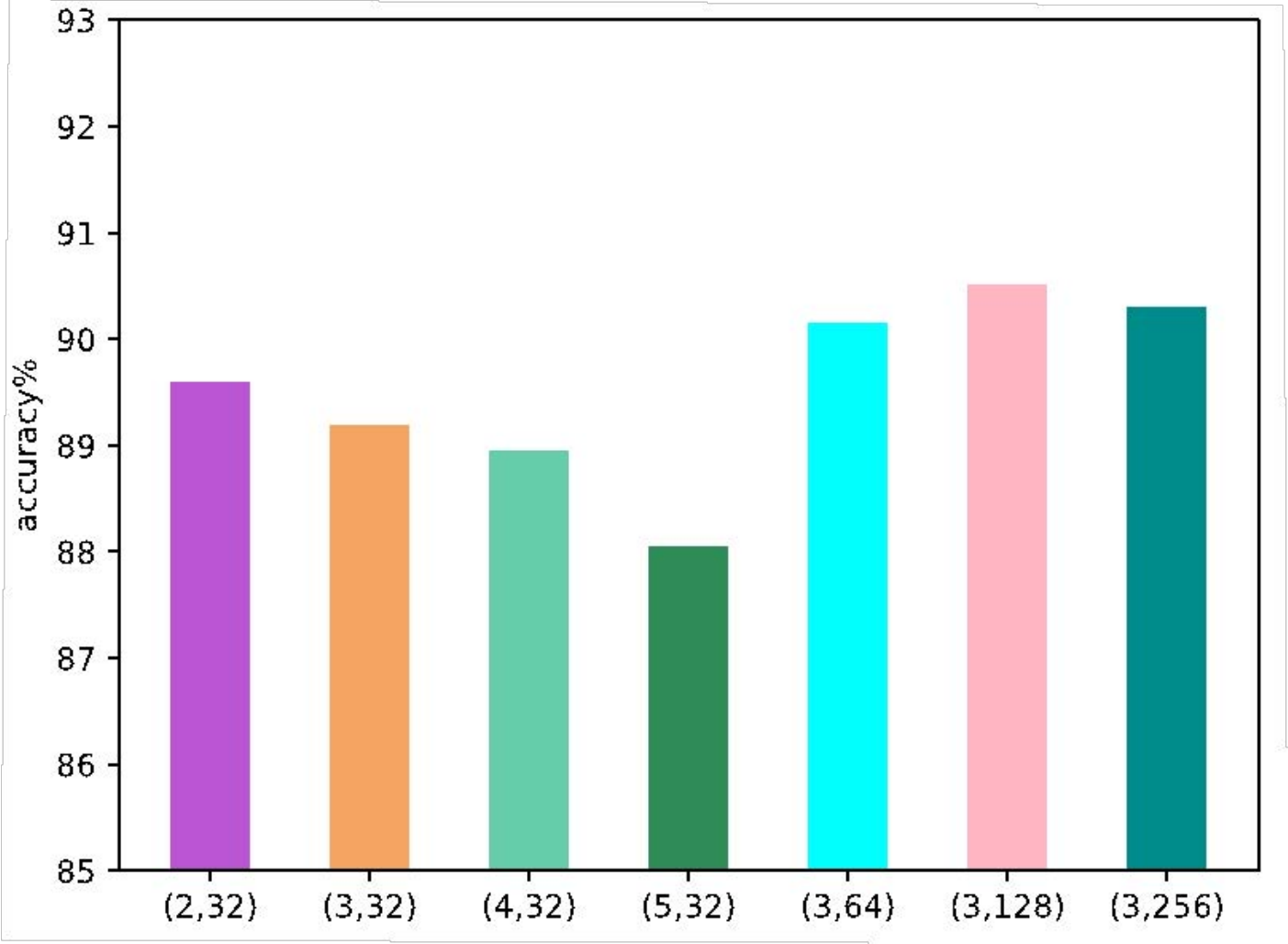}
    \end{minipage}
}
\subfigure[rfid]
{
 	\begin{minipage}[b]{.2\linewidth}
        \centering
        \includegraphics[scale=0.2]{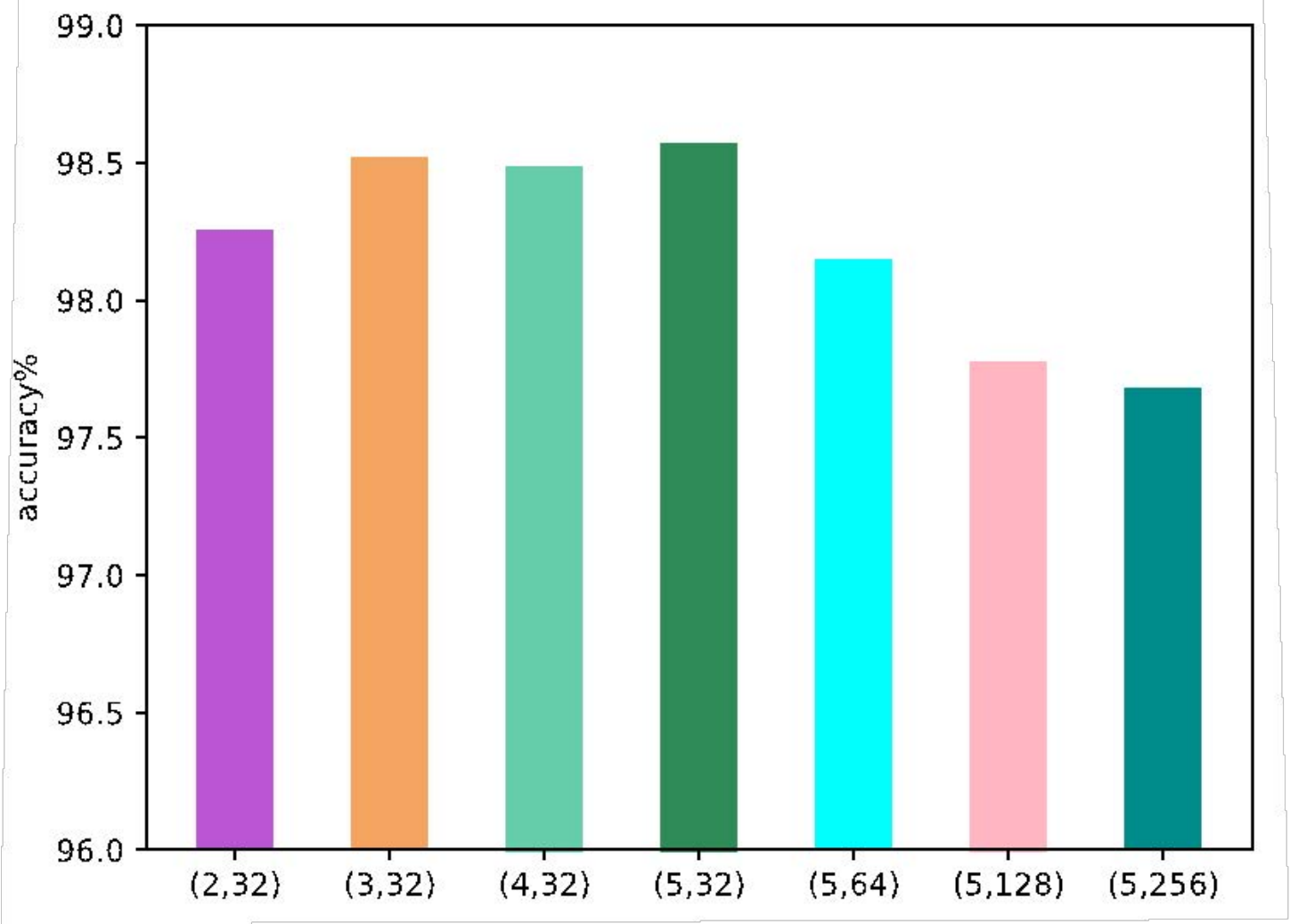}
    \end{minipage}
    \begin{minipage}[b]{.2\linewidth}
        \centering
        \includegraphics[scale=0.2]{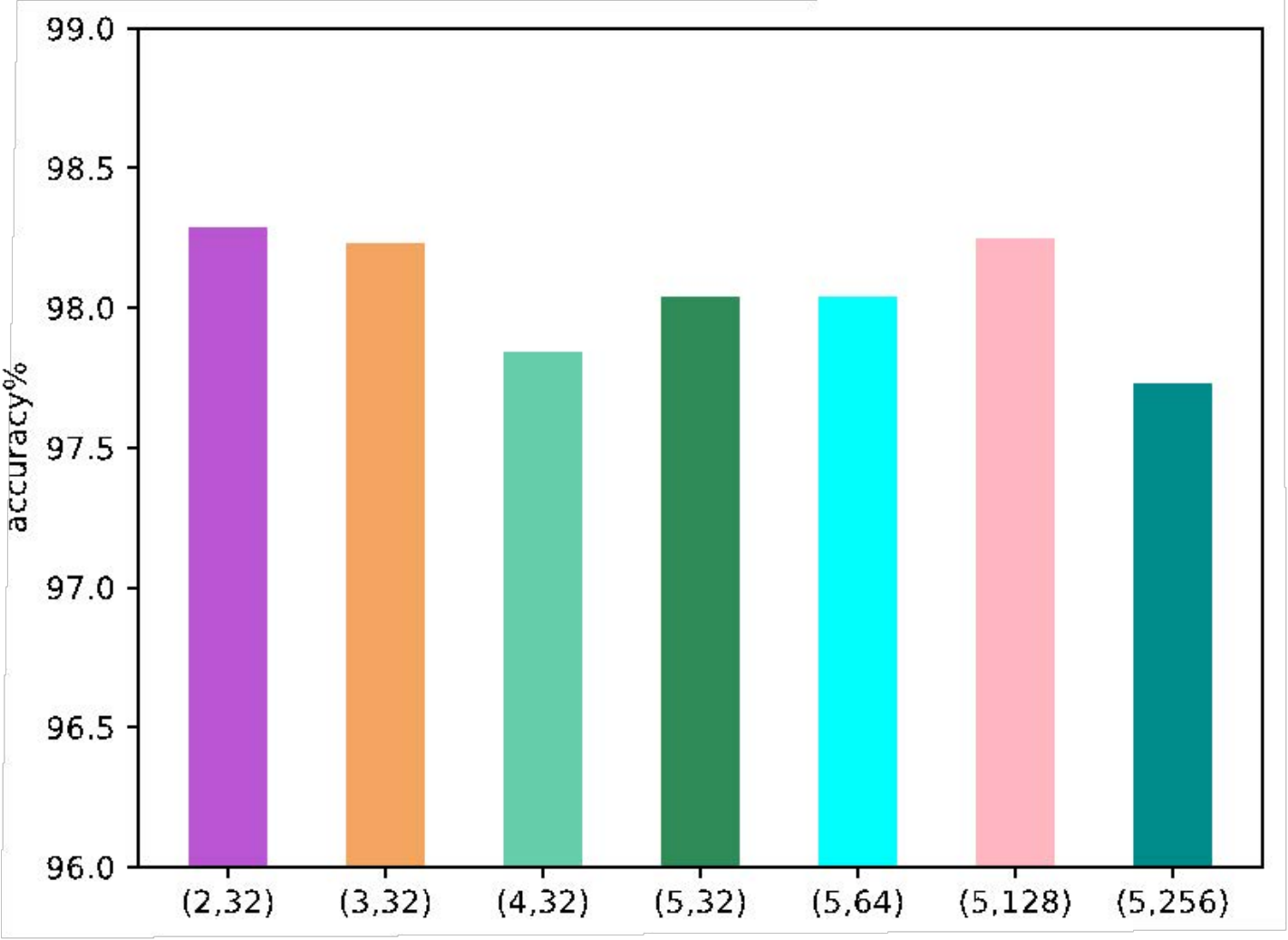}
    \end{minipage}
}
\subfigure[N-Balo]
{
 	\begin{minipage}[b]{.2\linewidth}
        \centering
        \includegraphics[scale=0.2]{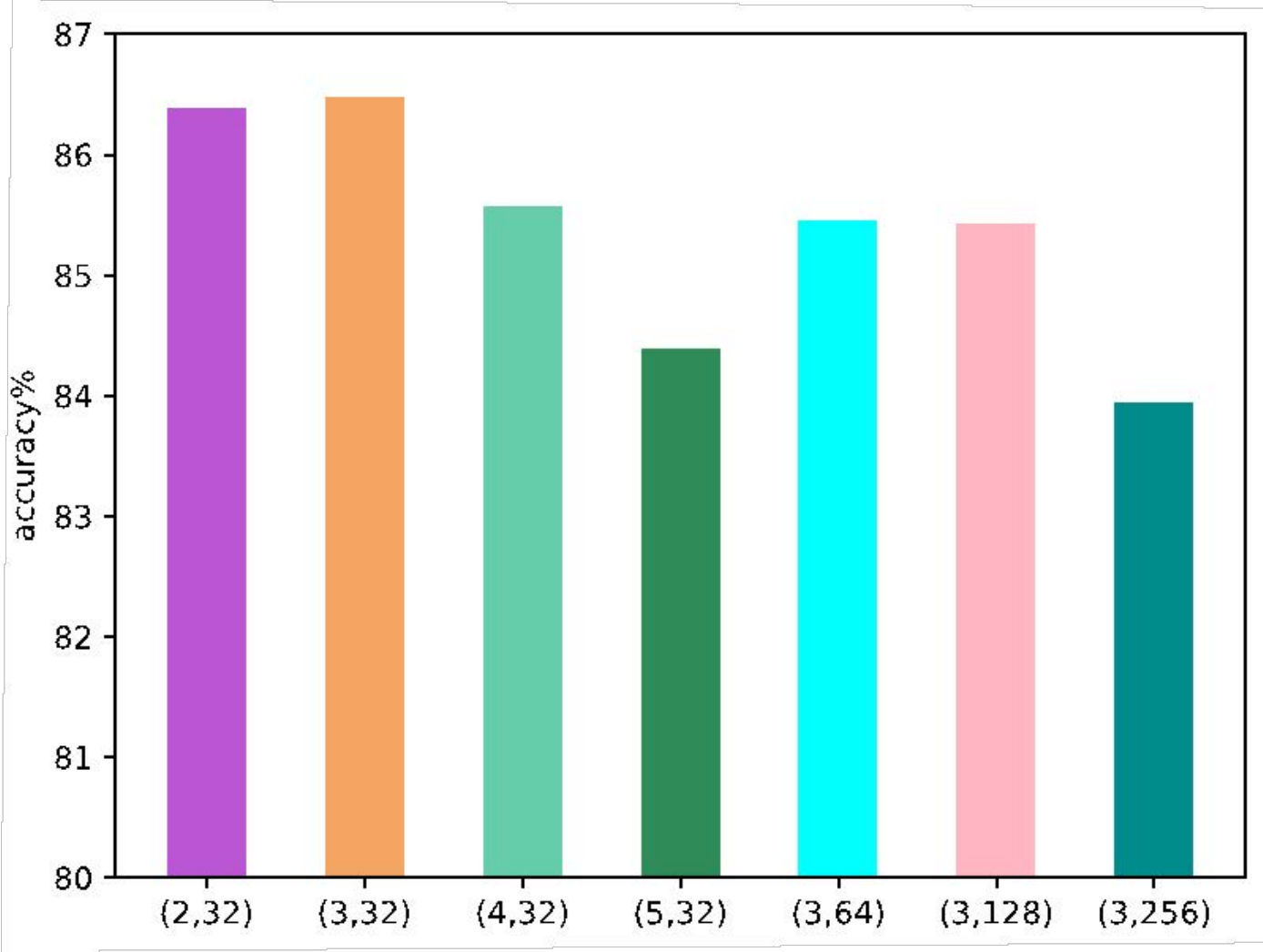}
    \end{minipage}
    \begin{minipage}[b]{.2\linewidth}
        \centering
        \includegraphics[scale=0.2]{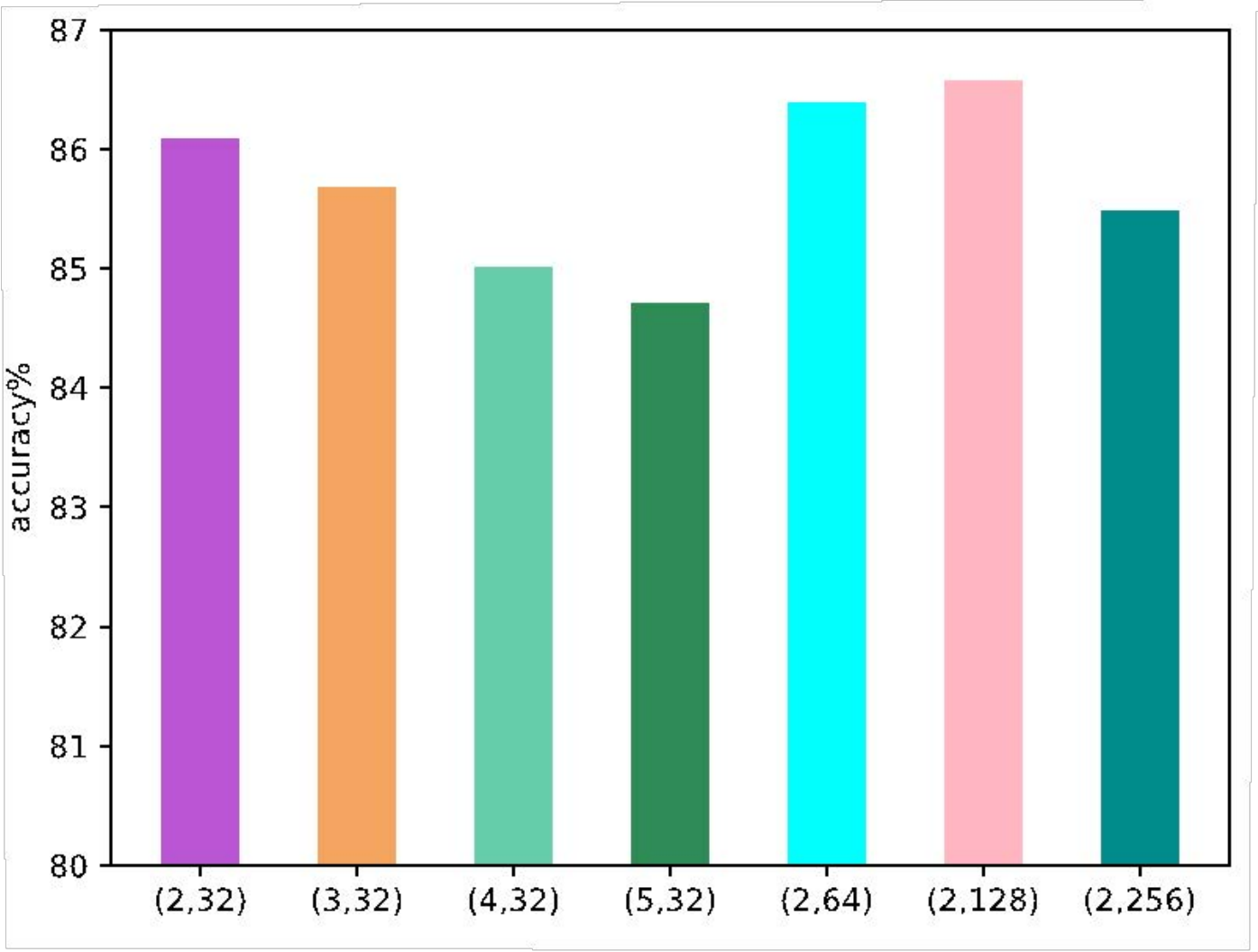}
    \end{minipage}
}
\caption{Variations in $L$ and ${D_{X'}}$ on different datasets}
\end{figure}

\subsection{Comparing Different Algorithms on Different Datasets}

We extensively compare the proposed technique with existing state-of-the-art methods for online learning. These include Relaxed Online Maximum Margin Algorithm[ROMMA] and its variation aggressive ROMMA algorithms(AROMMAS)\cite{45li2002relaxed}, Adaptive Regularization of Weights(AROW)\cite{46crammer2013adaptive}, confidence weighted algorithms (CW)\cite{47crammer2012confidence}, Soft Confidence Weight Learning algorithm (SCW1 and SCW2)\cite{48wang2016soft}, Online Gradient Descent algorithms(OGD)\cite{49ying2008online}, Passive-Aggressive  learning algorithms for Multiclass(PAM) and its variations PAM1 and PAM2 \cite{50crammer2006online}, Perceptron Algorithms with Max-score update(PerceptronM), Perceptron Algorithms with similarity-score update(PerceptronS), Perceptron Algorithms with uniform update(PerceptronU)\cite{51hoi2014libol}, Online Deep Learning: Learning Deep Neural Networks on the Fly(ONN)\cite{52sahoo2018online}.

Fig. 6 compares different methods on different datasets, the results indicate the superior performance of the proposed method. Compared with the second best method on the rotatedmnist dataset, our approach achieves a relative gain of 3.5\% and 4.77\% respectively. For the case of N-Balo dataset, our performance is only 2.35\% and 2.25\% below the SCW1 approach. In addition, the predictive results of different algorithms on these datasets are also listed in Table 2, and in Table 2, we provide more comparisons on the performance on different datasets, such as precision, F-1, and haming loss. For each evaluation index, our algorithm has achieved good results. The best performances are indicated in the bold.

\begin{figure}[htbp]
\centering
\subfigure[forestcovtype]
{
    \begin{minipage}[b]{.45\linewidth}
        \centering
        \includegraphics[scale=0.23]{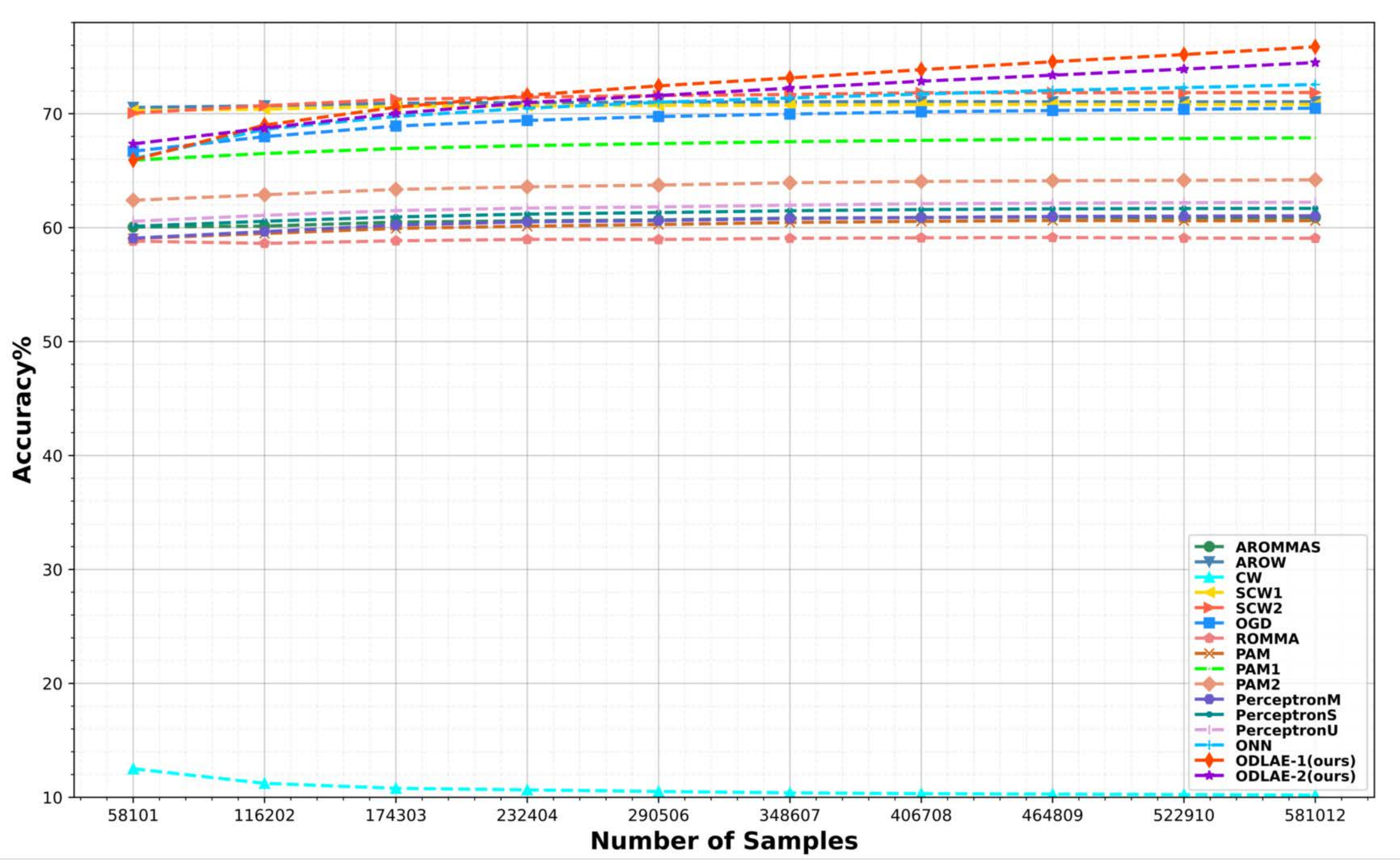}
    \end{minipage}
}
\subfigure[gesture]
{
 	\begin{minipage}[b]{.45\linewidth}
        \centering
        \includegraphics[scale=0.23]{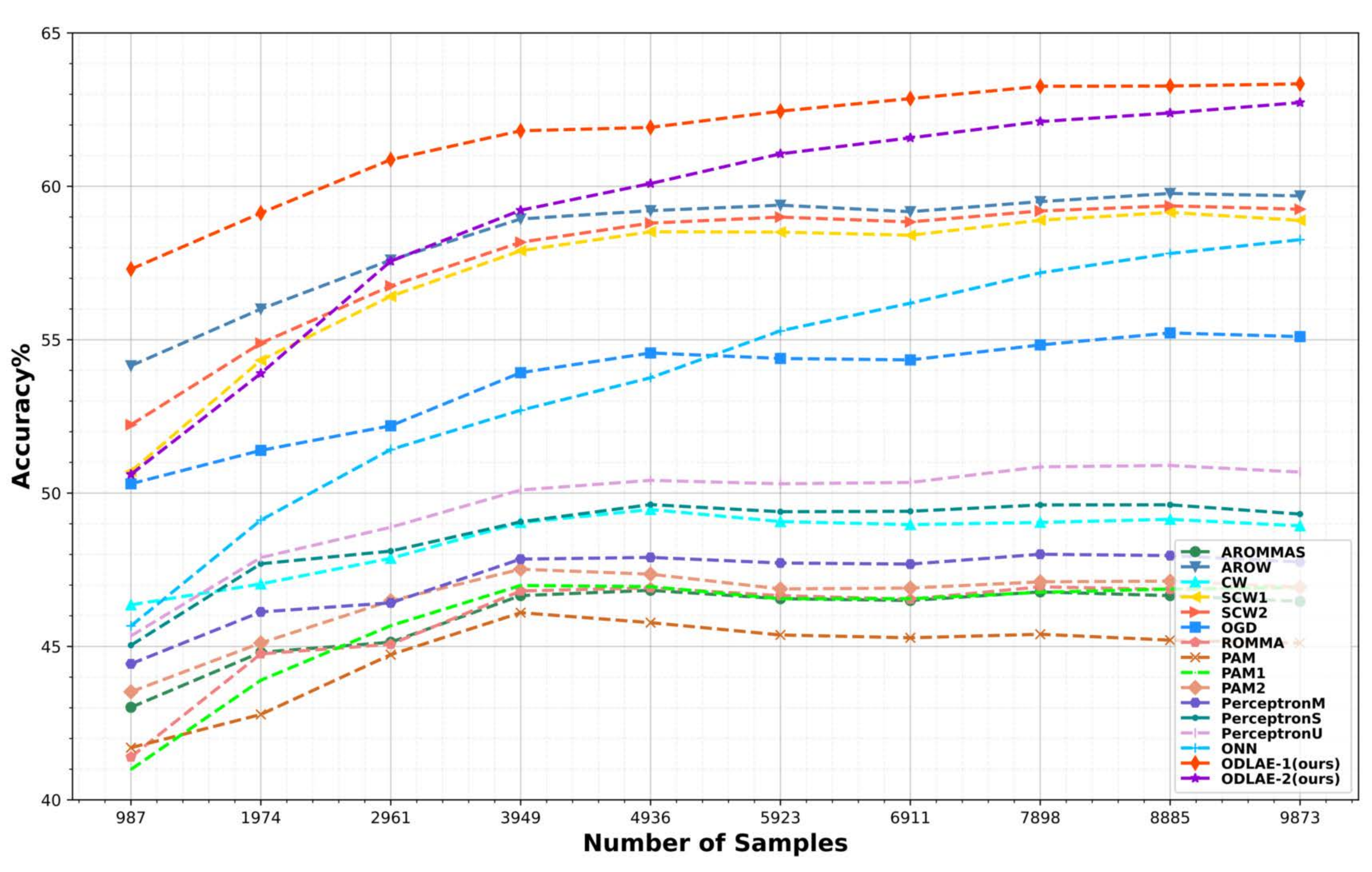}
    \end{minipage}
}
\subfigure[mnist]
{
 	\begin{minipage}[b]{.45\linewidth}
        \centering
        \includegraphics[scale=0.23]{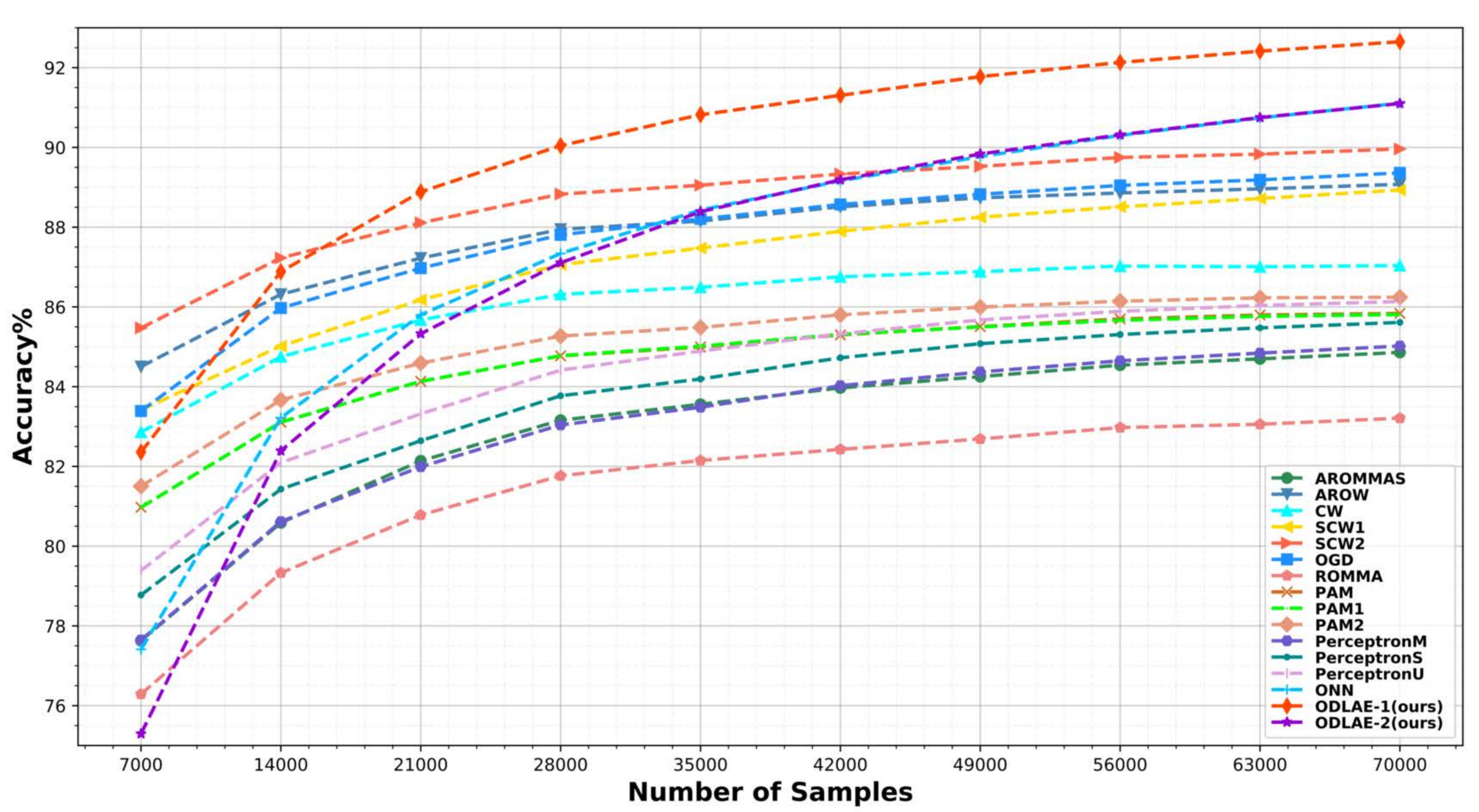}
    \end{minipage}
}
\subfigure[rotatedmnist]
{
 	\begin{minipage}[b]{.45\linewidth}
        \centering
        \includegraphics[scale=0.23]{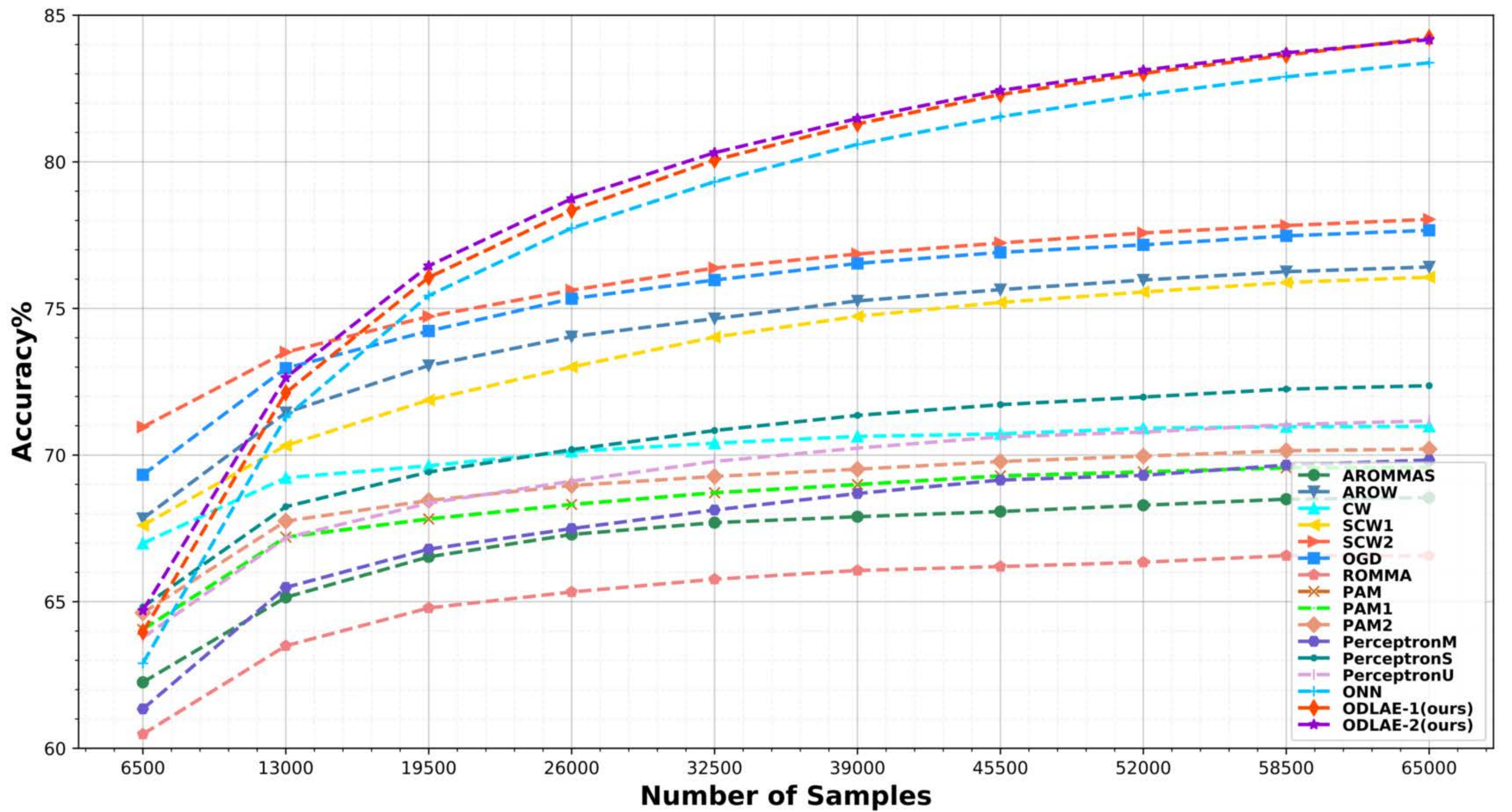}
    \end{minipage}
}
\subfigure[permutedmnist]
{
 	\begin{minipage}[b]{.45\linewidth}
        \centering
        \includegraphics[scale=0.23]{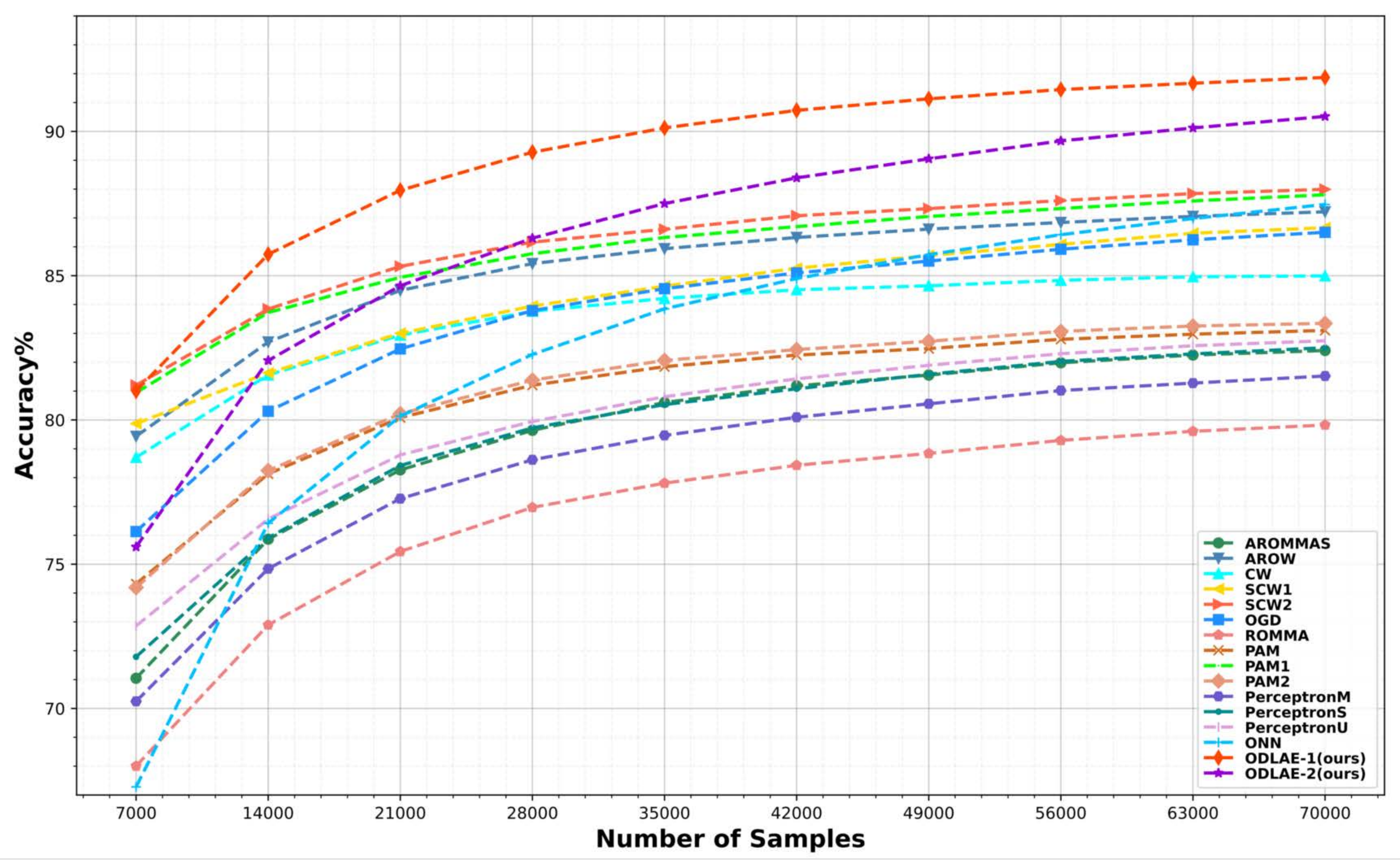}
    \end{minipage}
}
\subfigure[rfid]
{
 	\begin{minipage}[b]{.45\linewidth}
        \centering
        \includegraphics[scale=0.23]{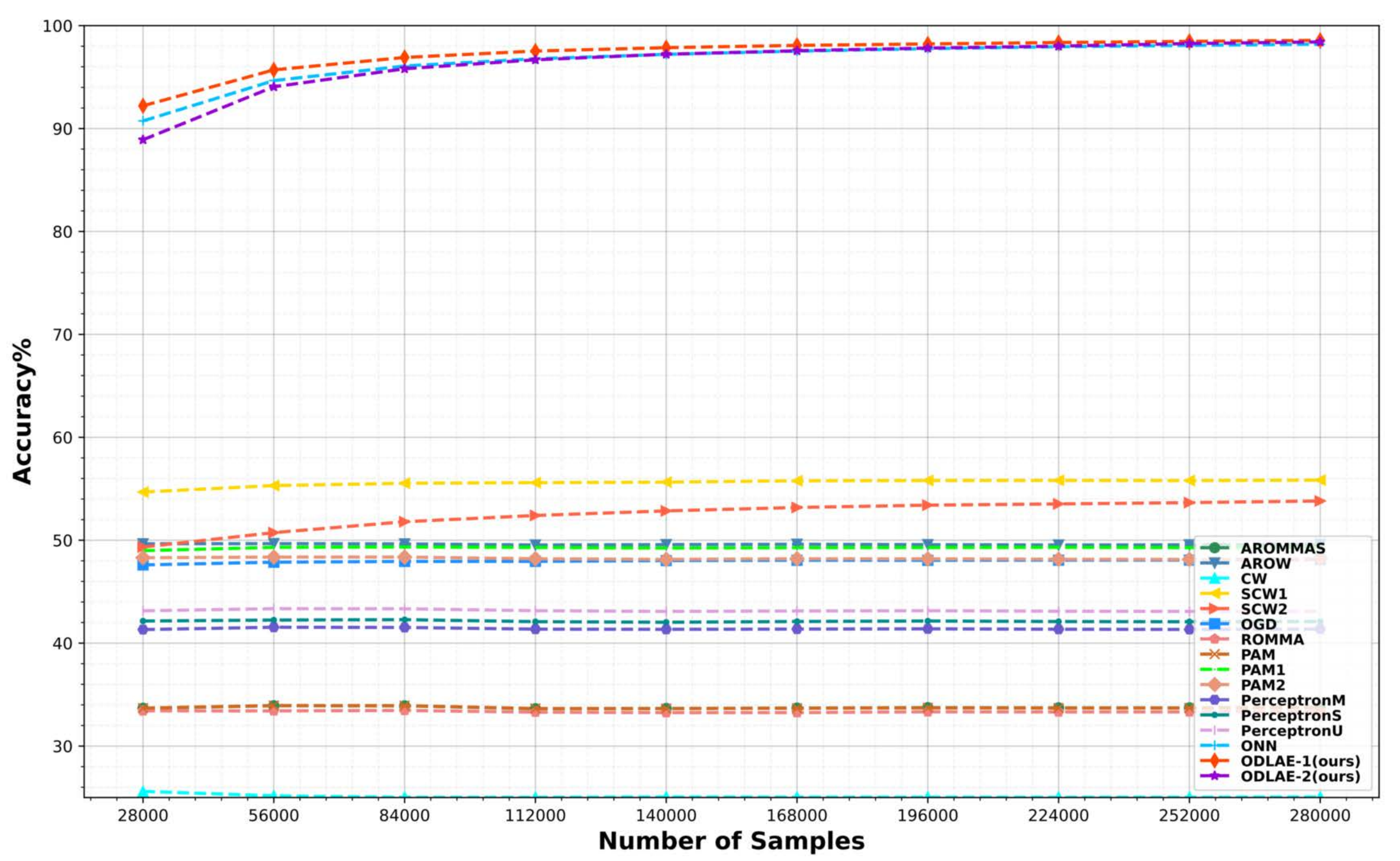}
    \end{minipage}
}
\subfigure[N-Balo]
{
 	\begin{minipage}[b]{.45\linewidth}
        \centering
        \includegraphics[scale=0.23]{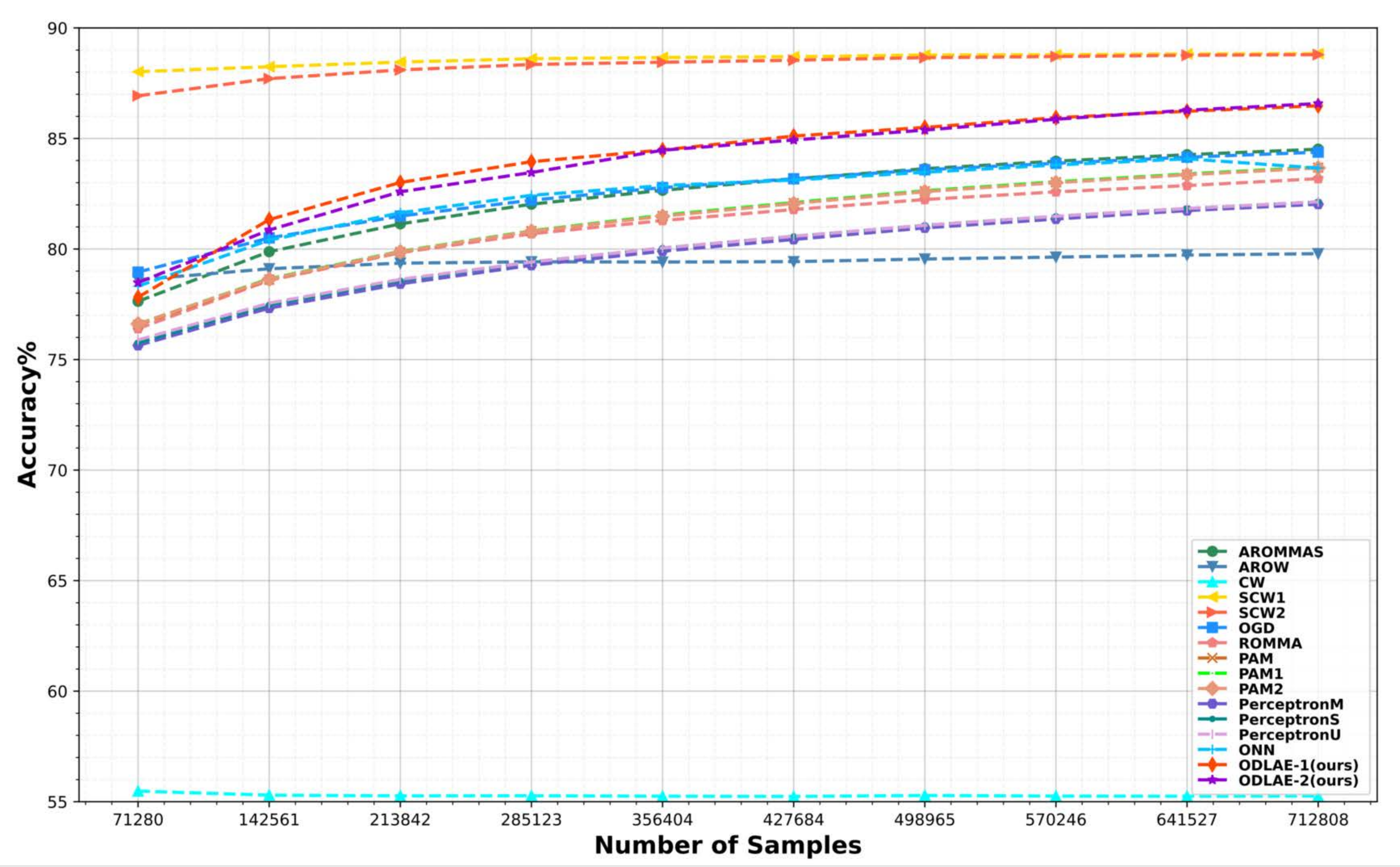}
    \end{minipage}
}
\caption{The accuracies of different algorithms on different datasets}
\end{figure}

\begin{table}[!htbp]
	\centering
	\caption{Comparisons on the different datasets with different evaluation indexes}
	\begin{tabular}{c|ccccccc}
		\hline
		\multirow{2}{*}{Method} & \multicolumn{7}{c}{Accuracy}\\ \cline{2-8}
	 &Forestcovtype	&gesture	&mnist	&rotatedmnist	&permutedmnist	&rfid &N-Balo \\ \hline
	 ODLAE-1(ours) &75.87\%	&63.34\%	&94.82\%	&86.88\%	&91.87\%	&98.50\%	&86.48\% \\ \hline
	 ODLAE-2(ours) &74.48\%	&62.73\%	&93.16\%	&88.15\%	&90.52\%	&98.42\%	&86.58\% \\ \hline
	 AROMMAS	&60.89\%	&46.47\%	&84.85\%	&68.54\%	&82.40\%	&33.76\%	&84.51\% \\ \hline
AROW	&71.04\%	&59.68\%	&89.07\%	&76.41\%	&87.21\%	&49.60\% 	&79.78\% \\ \hline
CW	&10.19\%	&48.93\%	&87.03\%	&70.98\%	&84.99\%	&25.05\%	&55.25\% \\ \hline
SCW1	&70.81\%	&58.88\%	&88.93\%	&76.06\%	&86.66\%	&55.83\%	&88.83\% \\ \hline
SCW2	&71.85\%	&59.25\%	&89.96\%	&78.04\%	&87.99\%	&53.81\%	&88.79\% \\ \hline
OGD	&70.47\%	&55.09\%	&89.36\%	&66.56\%	&86.50\%	&48.12\%	&84.38\% \\ \hline
ROMMA	&59.06\%	&46.92\%	&83.21\%	&77.66\%	&79.82\%	&33.35\%	&83.18\% \\ \hline
PAM	&60.60\%	&45.11\%	&85.83\%	&69.57\%	&83.10\%	&33.73\%	&83.67\% \\ \hline
PAM1	&67.88\%	&46.92\%	&85.80\%	&69.57\%	&87.80\%	&49.31\%	&83.70\% \\ \hline
PAM2	&64.19\%	&46.93\%	&86.24\%	&70.20\%	&83.34\%	&48.18\%	&83.66\% \\ \hline
ONN	&72.56\%	&58.26\%	&91.11\%	&83.38\%	&87.47\%	&98.20\%	&83.66\% \\ \hline
PerceptronM	&61.03\%	&47.75\%	&85.01\%	&69.83\%	&81.52\%	&41.36\%	&82.02\% \\ \hline
PerceptronS	&61.68\%	&49.31\%	&85.61\%	&72.36\%	&82.50\%	&42.10\%	&82.08\% \\ \hline
PerceptronU	&62.23\%	&50.68\%	&86.13\%	&71.16\%	&82.74\%	&43.09\%	&82.13\% \\ \hline
	\end{tabular}
\end{table}

\begin{table}[!htbp]
	\centering
	\begin{tabular}{c|ccccccc}
		\hline
		\multirow{2}{*}{Method} & \multicolumn{7}{c}{F-1}
\\ \cline{2-8}
	 &Forestcovtype	&gesture	&mnist	&rotatedmnist	&permutedmnist	&rfid &N-Balo \\ \hline
ODLAE-1(ours)	&0.7507	&0.6117	&0.9482	&0.8684	&0.9185	&0.9857	&0.8473 \\ \hline
ODLAE-2(ours)	&0.7401	&0.6109	&0.9315	&0.8813	&0.9051	&0.9828	&0.8629 \\ \hline
AROMMAS	&0.6092	&0.4647	&0.8484	&0.6853	&0.8237	&0.3376	&0.8450 \\ \hline
AROW	&0.6952	&0.5723	&0.8901	&0.7596	&0.8711	&0.4323	&0.7628 \\ \hline
CW	&0.0918	&0.4876	&0.8702	&0.7088	&0.8496	&0.2084	&0.4607 \\ \hline
SCW1	&0.7063	&0.5750	&0.8890	&0.7590	&0.8663	&0.5562	&0.8573 \\ \hline
SCW2	&0.7089	&0.5791	&0.8994	&0.7790	&0.8796	&0.5235	&0.8588 \\ \hline
OGD	&0.6866	&0.5358	&0.8935	&0.7756	&0.8647	&0.4067	&0.8397 \\ \hline
ROMMA	 &0.5929	&0.4685	&0.8320	&0.6664	&0.7984	&0.3341	&0.8320 \\ \hline
PAM	&0.6060	&0.4490	&0.8580	&0.6945	&0.8305	&0.3374	&0.8365 \\ \hline
PAM1	&0.6699	&0.4651	&0.8577	&0.6945	&0.8776	&0.4548	&0.8367 \\ \hline
PAM2	&0.6378	&0.4652	&0.8621	&0.7006	&0.8329	&0.4524	&0.8363 \\ \hline
ONN	&0.7136	&0.5418	&0.9109	&0.8329	&0.8747	&0.9820	&0.8348 \\ \hline
PerceptronM	&0.6103	&0.4771	&0.8502	&0.6983	&0.8153	&0.4083	&0.8203 \\ \hline
PerceptronS	&0.6156	&0.4912	&0.8561	&0.7235	&0.8251	&0.4151	&0.8210 \\ \hline
PerceptronU	&0.6200	&0.5033	&0.8614	&0.7116	&0.8275	&0.4246	&0.8216 \\ \hline
	\end{tabular}
\end{table}

\begin{table}[!htbp]
	\centering
	\begin{tabular}{c|ccccccc}
		\hline
		\multirow{2}{*}{Method} & \multicolumn{7}{c}{Haming Loss}
\\ \cline{2-8}
	 &Forestcovtype	&gesture	&mnist	&rotatedmnist	&permutedmnist	&rfid &N-Balo \\ \hline
ODLAE-1(ours)	&0.2412	&0.3666	&0.0517	&0.1312	&0.0813	&0.0140	&0.1352 \\ \hline
ODLAE-2(ours)	&0.2516	&0.3726	&0.0684	&0.1185	&0.0947	&0.0170	&0.1341 \\ \hline
AROMMAS	&0.3910	&0.5352	&0.1514	&0.3145	&0.1759	&0.6624	&0.1548 \\ \hline
AROW	&0.2895	&0.4031	&0.1092	&0.2358	&0.1278	&0.5039	&0.2021 \\ \hline
CW	&0.8980	&0.5106	&0.1296	&0.2902	&0.1500	&0.7494	&0.4474 \\ \hline
SCW1	&0.2918	&0.4111	&0.1106	&0.2393	&0.1333	&0.4416	&0.1116 \\ \hline
SCW2	&0.2814	&0.4074	&0.1003	&0.2196	&0.1200	&0.4618	&0.1120 \\ \hline
OGD	&0.2952	&0.4490	&0.1063	&0.3343	&0.1349	&0.5187	&0.1561 \\ \hline
ROMMA	&0.4093	&0.5307	&0.1679	&0.2233	&0.2017	&0.6664	&0.1681 \\ \hline
PAM	&0.3939	&0.5488	&0.1416	&0.3042	&0.1689	&0.6626	&0.1632 \\ \hline
PAM1	&0.3211	&0.5307	&0.1419	&0.3042	&0.1219	&0.5068	&0.1629 \\ \hline
PAM2	&0.3580	&0.5306	&0.1375	&0.2979	&0.1665	&0.5181	&0.1633 \\ \hline
ONN	&0.2743	&0.4173	&0.8884	&0.1661	&0.1225	&0.0179	&0.1633 \\ \hline
PerceptronM	&0.3896	&0.5224	&0.1498	&0.3016	&0.1847	&0.5863	&0.1797 \\ \hline
PerceptronS	&0.3831	&0.5068	&0.1438	&0.2763	&0.1749	&0.5789	&0.1791 \\ \hline
PerceptronU	&0.3776	&0.4931	&0.1386	&0.2883	&0.1725	&0.5690	&0.1786 \\ \hline
	\end{tabular}
\end{table}

\begin{table}[!htbp]
	\centering
	\begin{tabular}{c|ccccccc}
		\hline
		\multirow{2}{*}{Method} & \multicolumn{7}{c}{Precision}
\\ \cline{2-8}
	 &Forestcovtype	&gesture	&mnist	&rotatedmnist	&permutedmnist	&rfid &N-Balo \\ \hline
ODLAE-1(ours)	&0.7554	&0.6116	&0.9481	&0.8683	&0.9184	&0.9856	&0.8601 \\ \hline
ODLAE-2(ours)	&0.7400	&0.6073	&0.9314	&0.8812	&0.9050	&0.9828	&0.8640 \\ \hline
AROMMAS	&0.6095	&0.4649	&0.8483	&0.6852	&0.8235	&0.3386	&0.8450 \\ \hline
AROW	&0.6959	&0.5784	&0.8902	&0.7600	&0.8713	&0.4644	&0.7547 \\ \hline
CW	&0.3757	&0.4941	&0.8701	&0.7081	&0.8494	&0.2750	&0.5507 \\ \hline
SCW1	&0.7048	&0.5759	&0.8889	&0.7580	&0.8661	&0.5632	&0.8490 \\ \hline
SCW2	&0.7068	&0.5851	&0.8994	&0.7783	&0.8794	&0.5336	&0.8800 \\ \hline
OGD	&0.6900	&0.5366	&0.8934	&0.7751	&0.8647	&0.4570	&0.8410 \\ \hline
ROMMA	&0.5954	&0.4717	&0.8321	&0.6673	&0.7986	&0.3352	&0.8323 \\ \hline
PAM	&0.6060	&0.4604	&0.8578	&0.6934	&0.8301	&0.3383	&0.8363 \\ \hline
PAM1	&0.6669	&0.4743	&0.8575	&0.6934	&0.8775	&0.4701	&0.8366 \\ \hline
PAM2	&0.6349	&0.4743	&0.8618	&0.6995	&0.8324	&0.4563	&0.8362 \\ \hline
ONN	&0.7232	&0.5668	&0.9109	&0.8330	&0.8752	&0.9821	&0.8352 \\ \hline
PerceptronM	&0.6103	&0.4828	&0.8504	&0.6987	&0.8156	&0.4164	&0.8204 \\ \hline
PerceptronS	&0.6144	&0.4946	&0.8562	&0.7237	&0.8254	&0.4236	&0.8213 \\ \hline
PerceptronU	&0.6182	&0.5053	&0.8616	&0.7118	&0.8279	&0.4344	&0.8220 \\ \hline
	\end{tabular}
\end{table}
 
\subsection{Comparing Different Algorithm’s Robustness on Different Datasets}

Generally speaking, a good encoding representation should be devoted to capturing useful and stable structures from maybe partially collapsed inputs with unknown distribution. In this section, motivated by this principle, in order to further improve the robustness of the algorithm to learning latent representation of partially corrupted input data. We introduce and motivate a new training principle for  ODLAE based on the idea of Denoising Auto-Encoder(DAE), a more recent variant of the basic autoencoder. 

This is done by first corrupting the initial input $x$ into $\tilde x$ in the form of a stochastic mapping $\tilde x \sim {q_D}(\tilde x\left| x \right.)$.Then the collapsed input $\tilde x$  is then mapped, as the same process as auto-encoder does before, to a hidden representation 

\begin{equation}
{\tilde h_0} = {s_g}({W^{(0)}} \cdot \tilde x + {b^{(0)}})
\end{equation}

\begin{equation}
{\tilde h_{l + 1}} = {s_g}({W^{(l)}} \cdot {\tilde h_l} + {b^{(l)}}), l = 1,2, \ldots ,L - 1
\end{equation}

where ${s_g}$  is a non-liner activation function, ${W^{(0)}} \in \mathbb{R}^{{{D'}_X} \times {D_X}}$, and ${W^{(l)}} \in \mathbb{R}^{{{D'}_X} \times {{D'}_X}}$, $l = 1,2, \ldots ,L - 1$ are the weight matrix, and ${b^{(0)}} \in \mathbb{R}^{{{D'}_X}}$, $l = 1, \ldots ,L - 1$ are the bias vector, from which we could then obtain the reconstructed input $z$:

\begin{equation}
{\tilde h'_{L - 1}} = {s_g}({W'^{(L)}} \cdot {\tilde h_L}(\tilde x) + {b^{(L)}})
\end{equation}

\begin{equation}
{\tilde h'_{l - 1}} = {s_g}({W'^{(l)}} \cdot {\tilde h_L}(\tilde x) + {b'^{(l)}}), l = 0,1, \ldots ,L - 1
\end{equation}

\begin{equation}
z = {W'^{(0)}} \cdot {\tilde h_0}(\tilde x) + {b'^{(0)}}
\end{equation}

where ${W'^{(0)}} \in \mathbb{R}^{{D_Y} \times {{D'}_X}}$ and ${W^{(l)}} \in \mathbb{R}^{{{D'}_X} \times {{D'}_X}}$, $l = 1,2, \ldots ,L$ are the weight matrix, and ${b'^{(0)}} \in \mathbb{R}^{{D_Y}}$ and ${b^{(l)}} \in \mathbb{R}^{{{D'}_X}}$, $l = 1,2, \ldots ,L$ are the bias vector. The DAE training process consists of finding parameters $\Xi  = \{ {W^{(l)}},{W'^{(l)}},{b^{(l)}},{b'^{(l)}}\} $, $l = 0,1, \ldots ,L$ which minimize the reconstruction error. This corresponds to minimizing the following the squared error loss ${L_{re}}(x,z) = \left\| {x - z} \right\|_2^2$. Other different from the ODLAE-1 and ODLAE-2 is that, we only replace the auto-encoder in ODLAE-1 or ODLAE-2 with the denoising auto-encoder, the remaining part is consistent with the ODLAE-1 and ODLAE-2, so for convenience, we dub it as ODLDAE-1and ODLDAE-2 respectively. 

As we can see in Fig.6, we compare the accuracies of different algorithms on different datasets with and without adding noise. The results indicate superior performance of the proposed method in all datasets, including N-Balo dataset, From the Fig.7, we can see that the accuracies of ODLDAE-1and ODLDAE-2 algorithms are slightly inferior to SCW1 algorithm when noise data is not added on denoising autoencoder, but the accuracies of our algorithm are higher than that of SCW1 algorithm after adding noise to data, and the accuracies of ODLDAE-1and ODLDAE-2 algorithms are higher than other baseline algorithms on other datasets. In addition, the accuracies of different algorithm on different dataset with adding noise are listed in the Table3. The best performances are indicated in the bold.

\begin{figure}[htbp]
\centering
\subfigure[forestcovtype]
{
    \begin{minipage}[b]{.4\linewidth}
        \centering
        \includegraphics[scale=0.46]{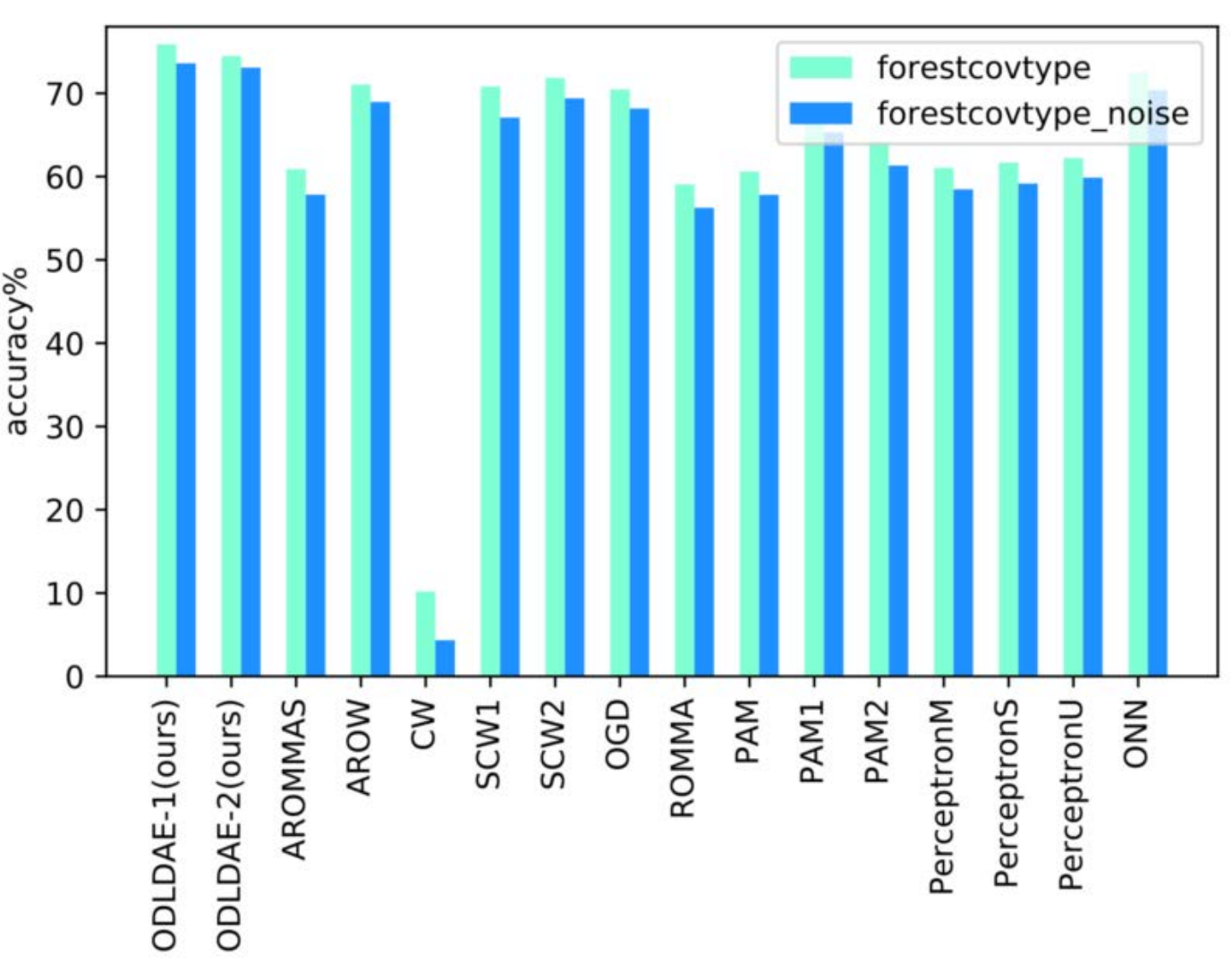}
    \end{minipage}
}
\subfigure[gesture]
{
 	\begin{minipage}[b]{.4\linewidth}
        \centering
        \includegraphics[scale=0.46]{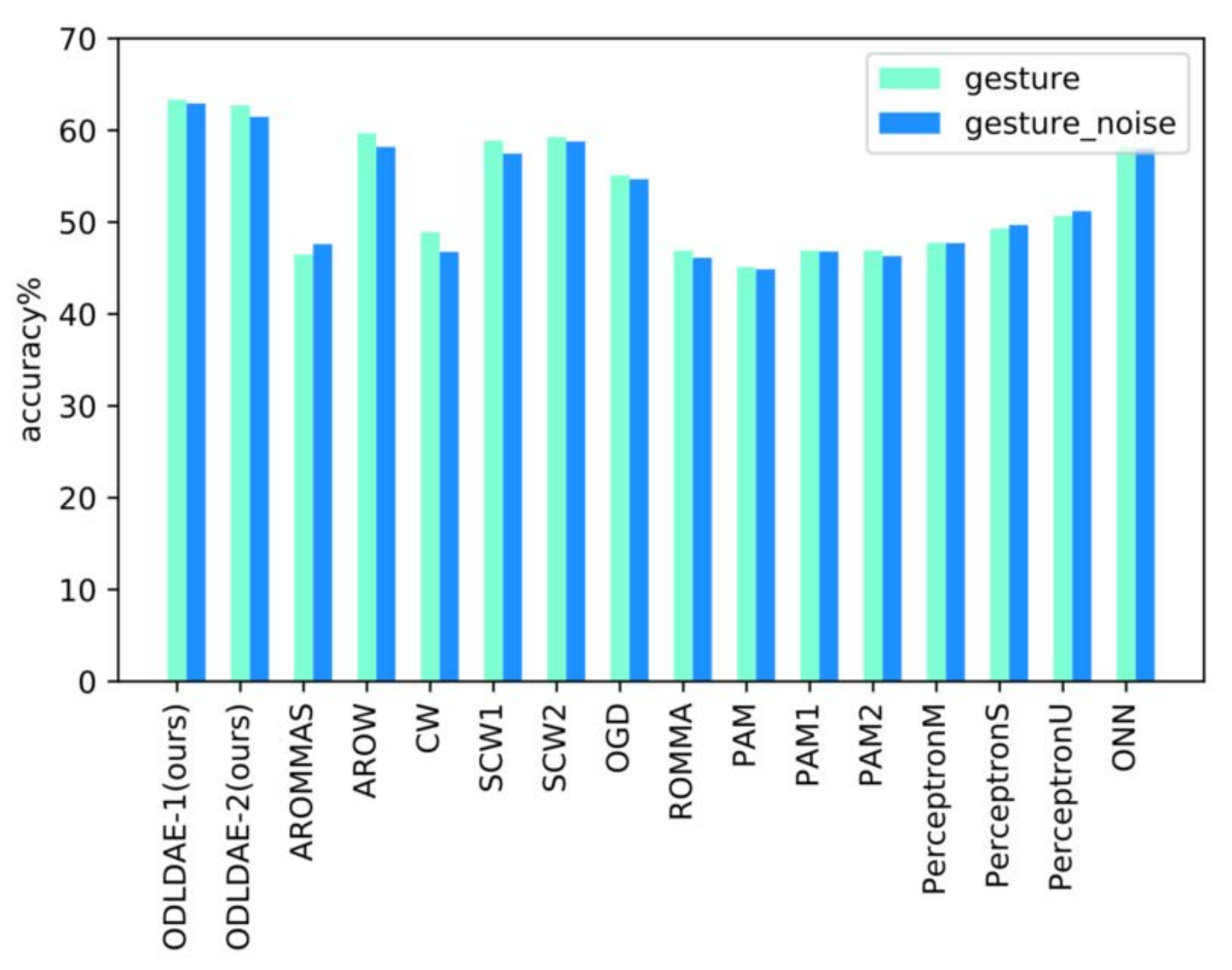}
    \end{minipage}
}
\subfigure[mnist]
{
 	\begin{minipage}[b]{.4\linewidth}
        \centering
        \includegraphics[scale=0.46]{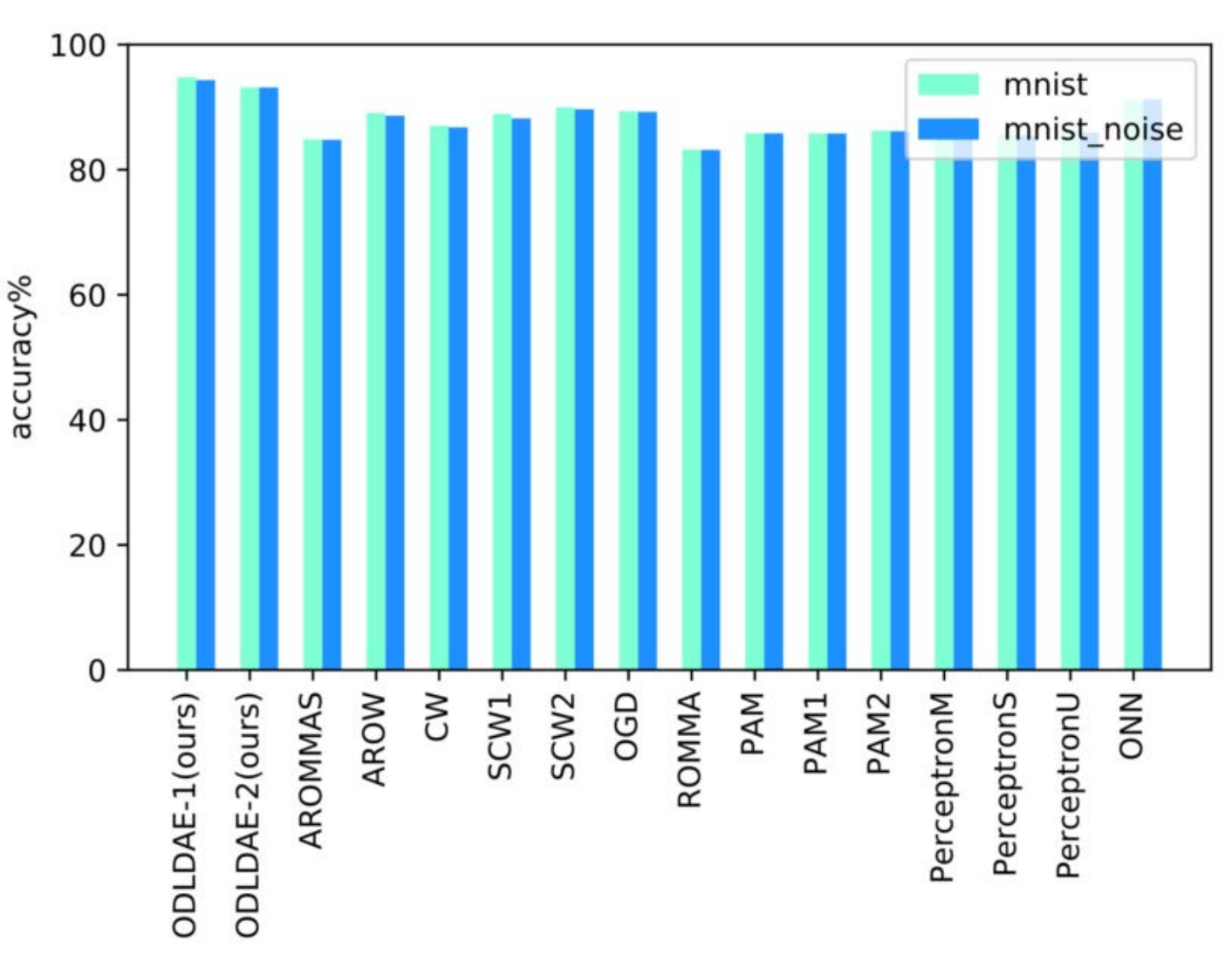}
    \end{minipage}
}
\subfigure[rotatedmnist]
{
 	\begin{minipage}[b]{.4\linewidth}
        \centering
        \includegraphics[scale=0.46]{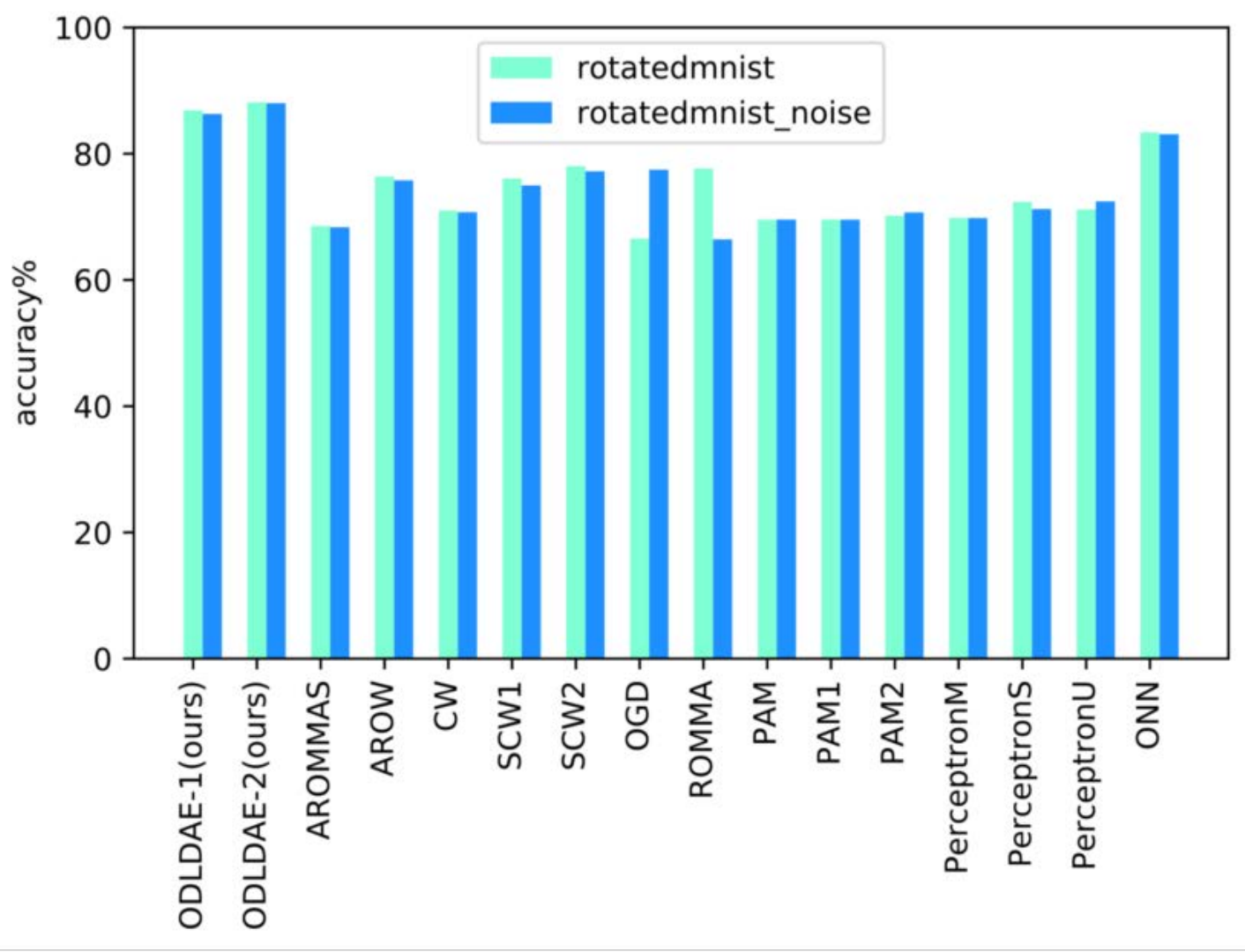}
    \end{minipage}
}
\subfigure[permutedmnist]
{
 	\begin{minipage}[b]{.4\linewidth}
        \centering
        \includegraphics[scale=0.46]{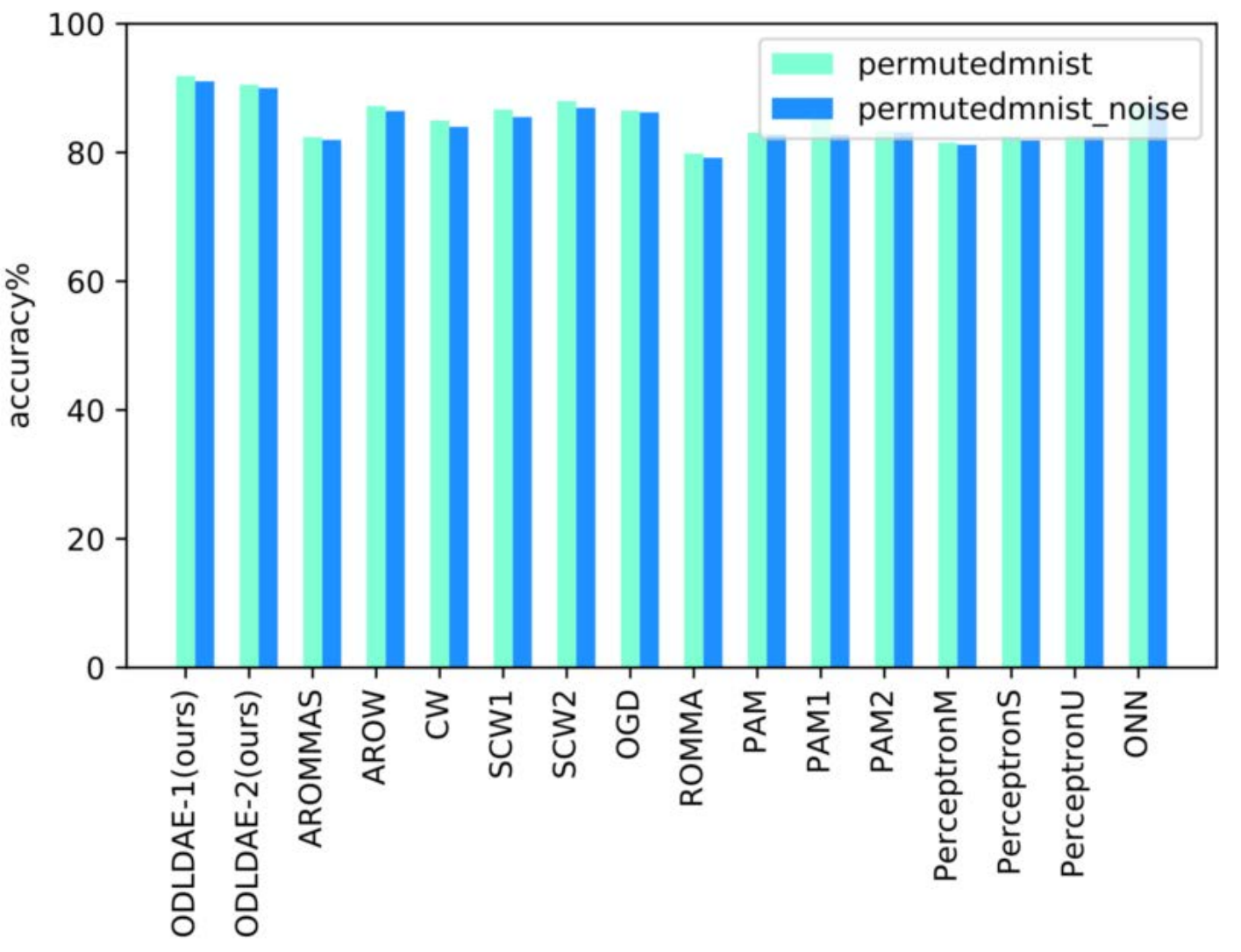}
    \end{minipage}
}
\subfigure[rfid]
{
 	\begin{minipage}[b]{.4\linewidth}
        \centering
        \includegraphics[scale=0.46]{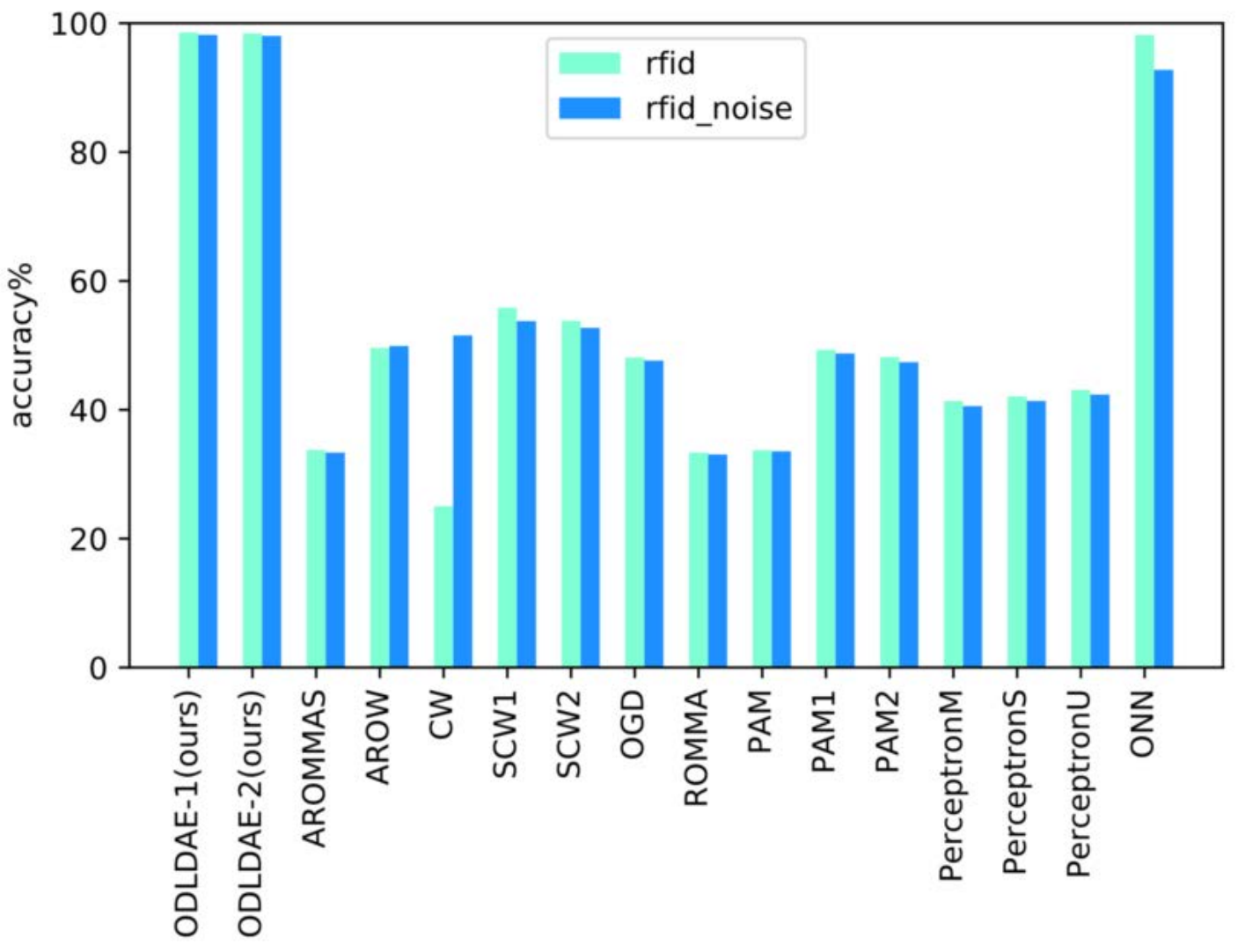}
    \end{minipage}
}
\subfigure[N-Balo]
{
 	\begin{minipage}[b]{.4\linewidth}
        \centering
        \includegraphics[scale=0.46]{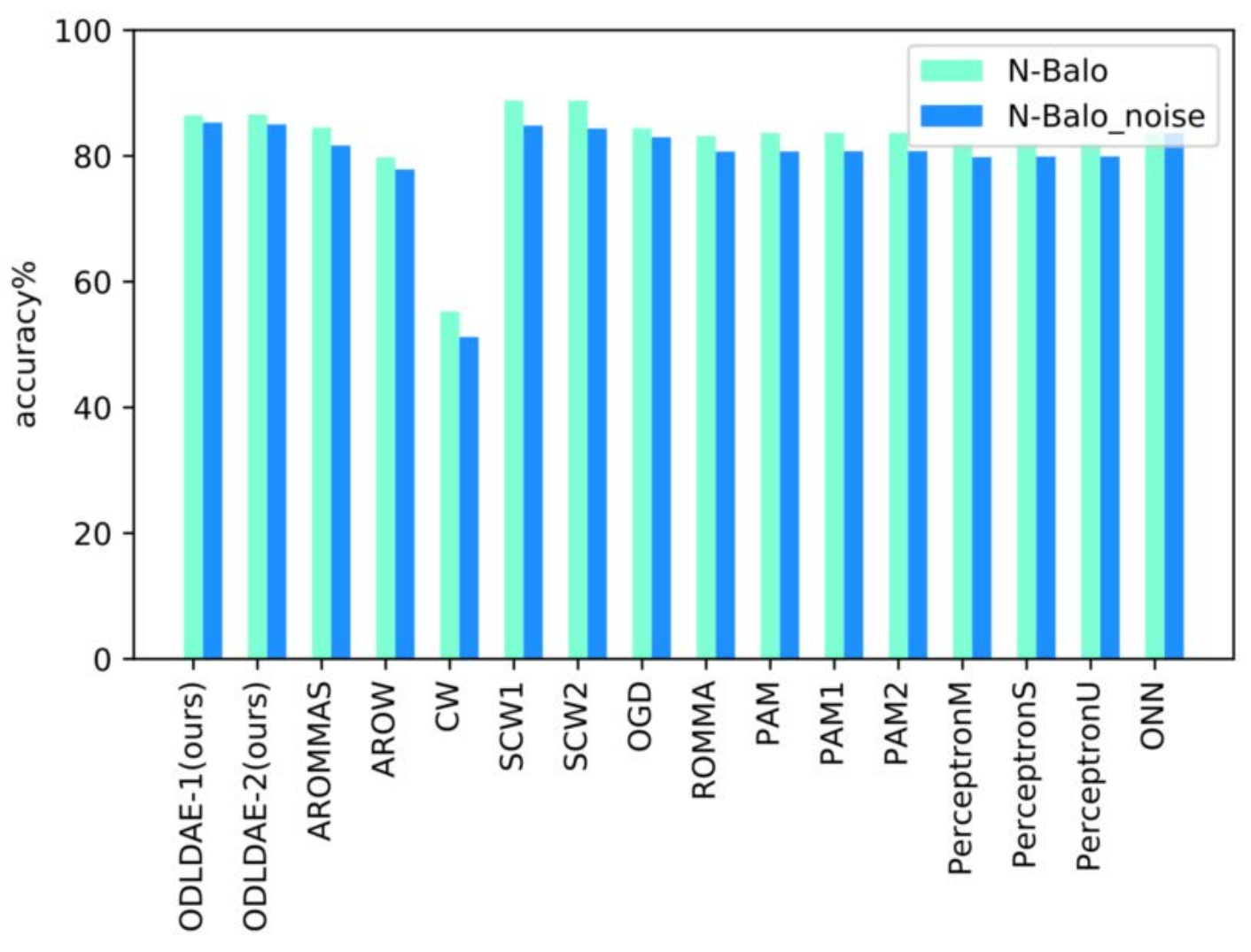}
    \end{minipage}
}
\caption{Comparing the accuracies of different algorithms on different data sets with and without adding noise}
\end{figure}

\begin{table}[!htbp]
	\centering
	\caption{ The predictive results of different algorithm on different dataset with adding noise}
	\begin{tabular}{c|ccccccc}
		\hline
		\multirow{3}{*}{Method} & \multicolumn{7}{c}{Accuracy}\\ \cline{2-8}
	&forestcovtype &gesture &mnist	&rotatedmnist	&permutedmnist	&rfid&N-Balo \\
    &(noise)  &(noise)  &(noise)  &(noise)  &(noise)  &(noise)  &(noise)
\\ \hline
ODLDAE-1(ours)	&73.60\%  &62.92\%	&94.33\%	&86.30\%	&91.05\%	&98.19\%	&85.30\% \\ \hline
ODLDAE-2(ours)	&73.08\%	&61.47\%	&93.15\%	&88.01\%	&90.01\%	&98.03\%	&85.01\% \\ \hline
aROMMS	&57.83\%	&47.61\%	&84.79\%	&68.35\%	&81.96\%	&33.55\%	&81.66\% \\ \hline
AROW	&68.96\%	&58.19\%	&88.62\%	&75.77\%	&86.44\%	&49.91\%	&77.85\% \\ \hline
CW	&4.34\%	&46.76\%	&86.78\%	&70.74\%	&83.98\%	&51.54\%	&51.21\% \\ \hline
SCW1	&67.08\%	&57.46\%	&88.20\%	&74.98\%	&85.51\%	&53.77\%	&84.83\% \\ \hline
SCW2	&69.40\%	&58.79\%	&89.67\%	&77.22\%	&86.93\%	&52.72\%	&84.35\% \\ \hline
OGD	&68.17\%	&54.68\%	&89.25\%	&77.48\%	&86.23\%	&47.64\%	&82.98\% \\ \hline
ROMMA	&56.25\%	&46.13\%	&83.17\%	&66.42\%	&79.18\%	&33.07\%	&80.69\% \\ \hline
PAM	&57.81\%	&44.87\%	&85.82\%	&69.57\%	&82.81\%	&33.56\%	&80.70\% \\ \hline
PAM1	&65.32\%	&46.81\%	&85.79\%	&69.56\%	&82.81\%	&48.75\%	&80.74\% \\ \hline
PAM2	&61.32\%	&46.31\%	&86.14\%	&70.07\%	&83.16\%	&47.40\%	&80.76\% \\ \hline
ONN	&70.36\%	&58.08\%	&91.25\%	&83.09\%	&87.47\%	&92.76\%	&83.65\% \\ \hline
PerceptronM	&58.47\%	&47.72\%	&84.84\%	&69.81\%	&81.19\%	&40.57\%	&79.81\% \\ \hline
PerceptronS	&59.16\%	&49.69\%	&85.48\%	&71.23\%	&81.87\%	&41.37\%	&79.90\% \\ \hline
PerceptronU	&59.87\%	&51.21\%	&85.95\%	&72.44\%	&82.50\%	&42.36\%	&79.91\% \\ \hline
	\end{tabular}
\end{table}

\section{ Conclusion and Future Work}

Based on auto-encoder, we propose a new two-phase online deep learning framework, by encoder, different abstract hierarchical latent representations are acquired, by means of the output-level fusion strategy and feature-level fusion strategy, constructed classifier make the best of the input information with different abstract level. We devise objective function to balance the prediction and reconstruction loss. This combination of loss functions gives us better prediction performance. 

In order to improve the robustness of the algorithm, we incorporate a denoising auto-encoder, which effectively improves the anti-noise performance of the algorithm compared with the normal algorithm. Our experimental results suggest that:(1) online deep learning provides an feasible way to learn effective implicit representation, and the good predictive results were also obtained;(2) the proposed algorithm obtains smaller reconstruction and prediction loss, the overall prediction performance is also better than other state-of-art algorithms;(3) compared with the simple weighted average fusion method, our paper proposes two different fusion strategies, which effectively improve the predictive accuracies of the algorithm;(4) the encouraging experimental results showed that the robustness of the algorithm is further improved in comparison to most of diverse online learning algorithms.

   For future work, we plan to consider sequential characteristics of samples to get better prediction results. At the same time, we can extend our single task online learning to multi-task online deep learning setup, adaptively learn task weight vector and task correlation from multi-task streaming data.

\bibliography{mybibfile}

\end{document}